\documentclass[runningheads]{llncs}
\usepackage{graphicx}
\usepackage{caption, subcaption}
\usepackage{floatrow}
\usepackage{float}
\usepackage{xcolor}
\usepackage{amsfonts, amsmath}
\usepackage{amssymb}
\usepackage{units}
\definecolor{citecolor}{RGB}{83,83,182}
\definecolor{linkcolor}{RGB}{128,0,128}
\usepackage{hyperref}
\hypersetup{
	colorlinks=true,
	citecolor=citecolor,
	linkcolor=linkcolor,
	urlcolor=linkcolor
}

\usepackage[ruled,vlined]{algorithm2e}
\SetKwInOut{Parameters}{Parameters}
\usepackage{natbib}
\setcitestyle{authoryear} 
\usepackage{color}
\newcommand{\overbar}[1]{\mkern 1.5mu\overline{\mkern-1.5mu#1\mkern-1.5mu}\mkern 1.5mu}

\newcommand*{\VEC}[1]  {\ensuremath{\boldsymbol{#1}}}
\newcommand*{\MAT}[1]  {\ensuremath{\boldsymbol{#1}}}

\DeclareMathOperator{\tr}{tr}
\DeclareMathOperator{\D}{D}
\DeclareMathOperator{\symm}{sym}
\DeclareMathOperator{\ddiag}{ddiag}
\DeclareMathOperator{\uf}{uf}

\usepackage[normalem]{ulem}
\usepackage[misc]{ifsym}

\begin{document}
\title{Learning Graphical Factor Models with Riemannian Optimization}

\author{Alexandre Hippert-Ferrer \Letter\inst{1}\orcidID{0000-0002-7740-5415} \and
	Florent Bouchard\inst{2}\orcidID{0000-0003-3003-7317} \and
	Ammar Mian\inst{3}\orcidID{0000-0003-1796-8707} \and Titouan Vayer\inst{4}\orcidID{0000-0002-8115-572X} \and Arnaud Breloy\inst{5}\orcidID{0000-0002-3802-9015}}
\authorrunning{A. Hippert-Ferrer et al.}
%

\institute{Univ Gustave Eiffel, IGN, ENSG, LASTIG, F-77454 Marne-la-Vallée, France \email{alexandre.hippert-ferrer@univ-eiffel.fr} \and 
	Univ Paris-Saclay, CNRS, CentraleSup\'elec, Laboratoire des signaux et systèmes, Gif-sur-Yvette, France 
	\and Univ Savoie Mont Blanc, LISTIC, Annecy, France 
	\and Univ Lyon, Inria, CNRS, ENS de Lyon, UCB Lyon 1,
	LIP UMR 5668, F-69342, Lyon, France 
	\and Univ Paris-Nanterre, LEME, IUT Ville d'Avray, Ville d’Avray, France}

\tocauthor{Alexandre~Hippert-Ferrer, Florent~Bouchard, Ammar~Mian, Titouan~Vayer, Arnaud~Breloy}
\toctitle{Learning Graphical Factor Models with Riemannian Optimization}

\titlerunning{Learning Graphical Factor Models with Riemannian Optimization}
\maketitle              
\setcounter{footnote}{0} 
\begin{abstract}
	Graphical models and factor analysis are well-established tools in multivariate statistics.
	While these models can be both linked to structures exhibited by covariance and precision matrices, they are generally not jointly leveraged within graph learning processes.
	This paper therefore addresses this issue by proposing a flexible algorithmic framework for graph learning under low-rank structural constraints on the covariance matrix.
	The problem is expressed as penalized maximum likelihood estimation of an elliptical distribution (a generalization of Gaussian graphical models to possibly heavy-tailed distributions), where the covariance matrix is optionally constrained to be structured as low-rank plus diagonal (low-rank factor model).
	The resolution of this class of problems is then tackled with Riemannian optimization, where we leverage geometries of positive definite matrices 
	and positive semi-definite matrices of fixed rank that are well suited to elliptical models.
	Numerical experiments on synthetic and real-world data sets illustrate the effectiveness of the proposed approach.
	
	\keywords{Graph learning  \and Low-rank factor models \and Riemannian optimization.}
\end{abstract}

\section{Introduction}

Graphical models allow us to represent specific correlation structures between any two variables (entries) of multivariate observations.
Inferring the topology of this structure directly from the data is referred to as \textit{graph learning}, which has been increasingly leveraged in numerous applications, such as biology \citep{li2006gradient, smith2011network, stegle2015computational}, finance \citep{marti2021review}, or signal processing \citep{shuman2013emerging, kalofolias2016learn}.

Within Gaussian graphical models (GGMs), graph learning boils down to the problem of estimating the precision (inverse covariance) matrix of a Gaussian Markov random field \citep{friedman2008sparse, lake2010discovering}.
In practice, achieving an accurate covariance matrix estimation is often a difficult task due to low sample support. 
Thus, it is common to introduce prior assumptions on the structure of this matrix that guarantee a correct estimation with fewer samples. 
A popular approach is related to low-rank factorizations, which relies on the assumption that the data is driven by an underlying low-dimensional linear model, corrupted by an independent perturbation. The resulting covariance matrix decomposition then involves a core that is a low-rank positive semi-definite matrix. Such model is ubiquitous in statistics, and for example, at the heart of probabilistic principal component analysis \citep{tipping1999probabilistic}, low-rank factor analysis \citep{robertson2007maximum, khamaru2019computation, rubin1982algorithms}, and their many generalizations.

Since GGMs and low-rank factorizations share a common root in structured covariance (or precision) matrix estimation, it appears desirable to leverage both approaches in a unified graph learning formulation.
On one hand, linear dimension reduction approaches rely on particular spectral structures that can be beneficial for graph learning \citep{kumar2020unified}.
On the other hand, it also opens the way to graph-oriented view of sparse principal component analysis \citep{yoshida2010bayesian, meng2014learning}.
Though theoretically appealing, such unification is challenging because it formulates optimization problems with objective functions and constraints that apply both on the covariance matrix and its inverse.
Thus, deriving single-step learning algorithms for these models has only recently been addressed \citep{chandra2021bayesian}.

In this paper, we propose a new family of methods for graph learning with low-rank constraints  on the covariance matrix, hereafter referred to as graphical factor models (GFM).
First, we reformulate graph learning as a problem that encompasses both elliptical distributions and low-rank factor models.
The main interest of generalizing Gaussian graphical models to elliptical ones is to ensure robustness to underlying heavy-tailed distributions \citep{vogel2011elliptical, finegold2014robust, zhang2013multivariate, de2021graphical}. Moreover, additionally considering low-rank factor models allows for an effective dimensionality reduction.
The main novelty of our approach is to tackle the resulting class of constrained and penalized maximum likelihood estimation in a unified way with Riemannian optimization \citep{absil2008optimization, boumal2020introduction}.
To do so, we leverage geometries of of both the positive definite matrices~\citep{bhatia2009positive}, and positive semi-definite matrices of fixed rank~\citep{bonabel2009riemannian, bouchard2021riemannian} that are well suited to the considered models.
The corresponding tools allows us to develop optimization methods that ensure the desired structures for the covariance matrix, thus providing a flexible and numerically efficient framework for learning graphical factor models.

Finally, experiments\footnote{\scriptsize The code is available at: \url{https://github.com/ahippert/graphfactormodel}.} conducted on synthetic and real-world data sets demonstrate the interest of considering both elliptical distributions and factor model structures in a graph learning process.
We observe that the proposed algorithms lead to more interpretable graphs compared to unstructured models.
Notably, the factor model approaches compare well with the current state-of-the-art on Laplacian-constrained graph learning methods that require to set the number of components as additional prior information \citep{kumar2020unified, de2021graphical}. 
The interest of our method is twofold:
$i$) 
it requires less supervision
to unveil meaningful clusters in the conditional correlation structure of the data;
$ii$)
the computational bottleneck is greatly reduced,
as the proposed algorithm iterations only requires the thin-SVD of a low-rank factor, rather than the whole SVD of the Laplacian (or adjacency) matrix.

\section{Background and proposed framework}

\subsection{Gaussian graphical and related models}

Gaussian graphical models assume that each observation is a centered multivariate Gaussian random vector $\VEC{x} = [x_1, \ldots, x_p]^\top$ with covariance matrix $\mathbb{E}[\VEC{x}\VEC{x}^\top] = \MAT{\Sigma}$, denoted $\VEC{x}\sim\mathcal{N}(\MAT{0},\MAT{\Sigma})$.
For the corresponding Gaussian Markov random field, an undirected graph is matched to the variables as follows: each variable corresponds to a vertex, and an edge is present between two vertices if the corresponding random variables are conditionally dependent given the others \citep{dempster1972covariance, lauritzen1996graphical}.
The support of the precision matrix $\MAT{\Theta} = \MAT{\Sigma}^{-1}$ directly accounts for this conditional dependency, since 
\begin{equation}\label{eq:condcorr}
	{\rm corr}\left[ x_q x_\ell | \VEC{x}_{[\![ 1,p ]\!]\backslash \{q,\ell\}} \right]= - \MAT{\Theta}_{q\ell} / \sqrt{\MAT{\Theta}_{qq}\MAT{\Theta}_{\ell\ell}}.
\end{equation}
Hence, a non-zero entry $\MAT{\Theta}_{q\ell}$ implies a conditional dependency between the variables $x_q$ and $x_\ell$, underlined by an edge between vertices $q$ and $\ell$ of the graph.
Within Gaussian graphical models, graph learning is therefore tied to the problem of estimating the precision matrix $\MAT{\Theta}$ from a set of observations $\{ \VEC{x}_i \}_{i=1}^n \in (\mathbb{R}^p)^n$.
In order to exhibit such correlation structure, a standard approach is to resort to regularized maximum likelihood estimation, \emph{i.e.}, solving the problem:
\begin{equation}\label{eq:glasso}
	\underset{\MAT{\Theta}\in\mathcal{S}_p^{++}}{\rm maximize} \quad
	\log \det (\MAT{\Theta})
	- {\rm Tr} \{ \MAT{S} \MAT{\Theta} \}
	- \lambda h(\MAT{\Theta}),
\end{equation}
where $\MAT{S}=\frac{1}{n} \sum_{i=1}^n \VEC{x}_i\VEC{x}_i^\top$ is the sample covariance matrix, $h$ is a regularization penalty, and $\lambda\in\mathbb{R}^{+}$ is a regularization parameter.
The $\ell_1$-norm if often used as penalty in order to promote a sparse structure in $\MAT{\Theta}$, which is at the foundation of the well-known ${\rm GLasso}$ algorithm \citep{friedman2008sparse, lake2010discovering, mazumder2012graphical, fattahi2019graphical}.
Many other convex on non-convex penalties have been considered in this stetting \citep{lam2009sparsistency, shen2012likelihood, benfenati2020proximal}.
Depending on the assumptions, it can also be beneficial to consider penalties that promote certain types of structured sparsity patterns \citep{heinavaara2016inconsistency, tarzanagh2018estimation}.
Another main example of structure within the conditional correlations is total positivity, also known as attractive Gaussian graphical models, that assumes $\MAT{\Theta}_{q\ell}\leq 0,~\forall q \neq \ell$ \citep{fallat2017total, lauritzen2019maximum}.
In attractive Gaussian graphical models, the identifiability of the precision matrix to the graph Laplacian \citep{chung1997spectral}, has also attracted recent interest in graph learning \citep{egilmez2017graph, ying2020nonconvex, kumar2020unified}.

\subsection{Elliptical distributions}

A first shortcoming of graph learning as formulated in \eqref{eq:glasso} is its lack of robustness to outliers, or heavy-tailed distributed samples.
This is a consequence of the underlying Gaussian assumption, that cannot efficiently describe such data.
A possible remedy is to consider a larger family of multivariate distributions.
In this context, elliptical distributions \citep{anderson1990theory, kai1990generalized} offer a good alternative, that are well-known to provide robust estimators of the covariance matrix \citep{maronna1976robust, tyler1987distribution, wald2019globally}.
In our present context, this framework has been successfully used to extend graphical models \citep{vogel2011elliptical, finegold2014robust, zhang2013multivariate, de2021graphical}, as well as low-rank structured covariance matrices models \citep{zhao2006probabilistic, zhou2019robust, bouchard2021riemannian} for corrupted or heavy-tailed data.

A vector is said to follow a centered elliptically symmetric distribution of scatter matrix $\MAT{\Sigma}$ and density generator $g$, denoted $\VEC{x}\sim\mathcal{ES}(\MAT{0},\MAT{\Sigma},g)$, if its density has the form
\begin{equation}
	f  (\VEC{x})
	\propto
	{\det (\MAT{\Sigma})}^{-1/2} g(\VEC{x}_i^\top \MAT{\Sigma}^{-1} \VEC{x}_i ),
\end{equation}
which yields the negative log-likelihood 
\begin{equation}\label{eq:likelihood}
	\mathcal{L} (\MAT{\Sigma}) \propto \frac{1}{n} \sum_{i=1}^n 
	\rho (\VEC{x}_i^\top \MAT{\Sigma}^{-1} \VEC{x}_i )
	+
	\frac{1}{2}
	\log | \MAT{\Sigma} | + {\rm const.}
\end{equation}
for the sample set $\{ \VEC{x}_i \}_{i=1}^n$,
where $\rho(t) = - \log g(t)$. Notice that using $g(t)= \exp(-t/2)$ allow us to recover the Gaussian case.
However, the density generator $g$ enables more flexibility, and notably to encompass many heavy-tailed multivariate distributions. Among popular choices, elliptical distributions include the multivariate $t$-distribution with degree of freedom $\nu>2$, which is obtained with $g(t) = (1+t/\nu)^{-\frac{\nu+p}{2}}$.
For this distribution, the parameter $\nu$ measures the rate of decay of the tails. The Gaussian case also corresponds to the limit case $\nu\rightarrow \infty$.

\subsection{Low-rank factor models}

A second limitation of \eqref{eq:glasso} is that it does not account for other potential structure exhibited by the covariance or precision matrices.
This can problematic when the sample support is low ($n\simeq p$, or $n<p$) as the sample covariance matrix is
not a reliable estimate in these setups \citep{ledoit2004well, smith2005covariance, vershynin2012close}.
In this context, a popular approach consists in imposing low-rank structures on the covariance matrix.
These structures arise from the assumption that the data is driven by an underlying low-dimensional linear model, \emph{i.e.}, $\VEC{x} = \MAT{W}\VEC{s}  + \VEC{\epsilon} $, where $\MAT{W}$ is a rank-$k$ factor loading matrix, and $\VEC{s}\in\mathbb{R}^k$ and $\VEC{\epsilon}\in\mathbb{R}^p$ are independent random variables.
Thus the resulting covariance matrix is of the form $\MAT{\Sigma} \overset{\Delta}{=} \MAT{H} +  \MAT{\Psi}$, where $\MAT{H} = \MAT{W}\mathbb{E}[\VEC{s}\VEC{s}^\top]\MAT{W}^\top $ 
belongs to the set of positive semi-definite matrices of rank $k$, denoted $\mathcal{S}_{p,k}^+
=
\left\{
\MAT{\Sigma}\in \mathcal{S}_p ,
\MAT{\Sigma} \succcurlyeq  \MAT{0},
{\rm rank}(\MAT{\Sigma})=k
\right\}$.
This model is well known in statistics, and for example, at the core of probabilistic principal component analysis, that assumes $\MAT{\Psi}\propto \MAT{I}_p$ \citep{tipping1999probabilistic}.
Also notice that in Laplacian-constrained models, a rank-$k$ precision matrix implies a $(p-k)$-component graph \citep{chung1997spectral}.
Hence it is also interesting to leverage such spectral structures from the graph learning perspective \citep{kumar2020unified}.

In this paper, we will focus on the low-rank factor analysis models \citep{robertson2007maximum, rubin1982algorithms, khamaru2019computation}, that consider the general case $\MAT{\Psi}\in\mathcal{D}_p^{++}$, where $\mathcal{D}_p^{++} = \left\{ \MAT{\Sigma} = {\rm diag}(\VEC{d}), ~\VEC{d}\in \mathbb{R}^p_{+*}
\right\}$ denotes the space of positive definite diagonal matrices.
Given this model, the covariance matrix belongs to the space of rank-$k$ plus diagonal matrices, denoted as
\begin{equation}\label{eq:M_pk}
	\mathcal{M}_{p,k}
	= 
	\left\{
	\MAT{\Sigma}=
	\MAT{H} + \MAT{\Psi}
	,~
	\MAT{H}\in\mathcal{S}_{p,k}^+, \MAT{\Psi}\in\mathcal{D}_p^{++}
	\right\}.
\end{equation}
Notice that this parameterization reduces the dimension of the estimation problem reduces from $p(p+1)/2$ to $p(k+1) - k(k-3)/2$, which is why it is often used in regimes with few samples, or high dimensional settings.

\subsection{Learning elliptical graphical factor models}

We cast the problem of graph learning for elliptical graphical factor models as
\begin{equation}\label{eq:global_problem}
	\begin{array}{c l}
		\underset{\MAT{\Sigma}\in \mathcal{S}_p^{++}}{\rm minimize}
		&
		\mathcal{L}(\MAT{\Sigma}) + \lambda h (\MAT{\Sigma})
		\\
		{\rm subject~to} 
		&
		\MAT{\Sigma}\in \mathcal{M}_{p,k},
	\end{array}
\end{equation}
where $\mathcal{L}$ is the negative-log likelihood in \eqref{eq:likelihood}, and $\mathcal{M}_{p,k}$ is the space of rank-$k$ plus diagonal positive definite matrices in \eqref{eq:M_pk}.
The penalty $h$ is a smooth function that promotes a sparse structure on the graph, and $\lambda\in\mathbb{R}_+$ is a regularization parameter.
Although this formulation take into account many options, we focus on the usual element-wise penalty applied to the precision matrix $\MAT{\Theta}=\MAT{\Sigma}^{-1}$, defined as:
\begin{equation}
	h(\MAT{\Sigma}) = \sum_{q\neq \ell} \phi ( [\MAT{\Sigma}^{-1}]_{q\ell} )
\end{equation}
It is important to notice that the considered optimization framework will require $h$ to be smooth. This is why we presently use a surrogate of the $\ell_1$-norm defined as
\begin{equation}
	\phi(t) = \varepsilon\log(\cosh(t/\varepsilon)),
\end{equation}
with $\varepsilon>0$ ($\lim_{\varepsilon\to0} \phi(t)$ yields the $\ell_1$-norm).
In practice, we use $\varepsilon=1e^{-12}$. However, note that the output of the algorithm is not critically sensitive to this parameter: the obtained results were similar for $\varepsilon$ ranging from $1e^{-6}$ to the numerical tolerance.

In conclusion, the problem in \eqref{eq:global_problem} accounts for a low-rank structure in a graph learning formulation, which is also expected to be more robust to data following an underlying heavy-tailed distributions.
We then introduce four main cases, and their corresponding acronyms:

\noindent
\textsf{GGM}/\textsf{GGFM}: Gaussian graphical factor models (\textsf{GGFM}) are obtained with the Gaussian log-likelihood, \emph{i.e.}, setting $g(t)=\exp(-t/2)$ in \eqref{eq:likelihood}. The standard Gaussian graphical models (\textsf{GGM}) as in \eqref{eq:glasso} are recovered when dropping the constraint $\MAT{\Sigma}\in \mathcal{M}_{p,k}$.

\noindent
\textsf{EGM}/\textsf{EGFM}: elliptical graphical factor models (\textsf{EGFM}) are obtained for the more general case where $\mathcal{L}$ defines a negative log-likelihood of an elliptical distribution.
Dropping the constraint $\MAT{\Sigma}\in \mathcal{M}_{p,k}$ also yields a relaxed problem that we refer to as elliptical graphical models (\textsf{EGM}).

\section{Learning graphs with Riemmanian optimization}

\begin{algorithm}[t]
	\KwIn{
		Data $\{ \VEC{x}_i \}_{i=1}^n \in (\mathbb{R}^p)^n$}
	\Parameters{
		\textsf{GGM/GGFM/EGM/EGFM} $\rightarrow$ density generator $g$, rank $k$\\
		
		Regularization parameter $\lambda$ and penalty shape $\phi$\\
		
		Tolerance threshold {\rm tol}
	}
	\KwOut{Learned graph adjacency $\MAT{A}$ (Boolean)
		\vspace{0.2cm}
	}
	
	\If{k=p}{
		\textsf{GGM/EGM} initialization $\MAT{\Sigma}_{0} = \frac{1}{n} \sum_{i=1}^n \VEC{x}_i\VEC{x}_i^\top$
		
		\For{$t=0$ \textbf{to convergence}}{
			Compute the Riemannian gradient with 
			\eqref{eq:riem_grad_ggm1} and \eqref{eq:riem_grad_ggm2}
			\\
			Computes steps $\alpha_t$ and $\beta_t$ with \citep{hestenes1952methods}
			\\
			Update $\MAT{\Sigma}_{t+1}$ with
			\eqref{eq:riem_grad_spd}
			(transport in \eqref{eq:transport_spd},
			retraction in \eqref{eq:retraction_spd})
		}
	}
	\Else
	{
		\textsf{GGFM/EGFM} initialization
		
		($\MAT{V}_0 = \text{$k$ leading eigenvectors of $\frac{1}{n} \sum_{i=1}^n \VEC{x}_i\VEC{x}_i^\top$} $, $\MAT{\Lambda}_0 = \MAT{I}_k$, $\MAT{\Psi}_0 =\MAT{I}_p$)
		
		\For{$t=0$ \textbf{to convergence}}{
			Compute the Riemannian gradient with 
			\eqref{eq:riem_grad_mpk}
			\\
			Computes steps $\alpha_t$ and $\beta_t$ with \citep{hestenes1952methods}
			\\
			Update $(\MAT{V}_{t+1},\MAT{\Lambda}_{t+1},\MAT{\Psi}_{t+1})$ with
			\eqref{eq:riem_grad_spd} transposed to $\mathcal{M}_{p,k}$ (i.e., with transport in \eqref{eq:riem_grad_mpk}, retraction in \eqref{eq:retraction_mpk})
		}
		
		$\MAT{\Sigma}_{\rm end} = \MAT{V}_{\rm end} \MAT{\Lambda}_{\rm end} \MAT{V}_{\rm end}^{\top} +  \MAT{\Psi}_{\rm end} $ 
	}
	Compute 
	$\MAT{\Theta} = \MAT{\Sigma}_{\rm end}^{-1}
	$
	and conditional correlations 
	matrix $\tilde{\MAT{\Theta}}$
	from \eqref{eq:condcorr}
	
	Activate edges $(i,j)$ in adjacency  $\MAT{A} $
	for each $\tilde{\MAT{\Theta}}_{ij} \geq {\rm tol}$

	\caption{Graph learning with elliptical graphical (factor) models}
	\label{algo:MainAlgo}
\end{algorithm}

The optimization problem in \eqref{eq:global_problem} is non-convex and difficult to address.
Indeed, it involves objective functions and constraints that apply on both the covariance matrix and its inverse.
One approach to handle these multiple and complex constraints is to resort to variable splitting and alternating direction method of multipliers \citep{kovnatsky2016madmm} which has, for example, been considered for Laplacian-constrained models in \citep{zhao2019optimization, de2021graphical}.
In this work, we propose a new and more direct approach by harnessing Riemannian optimization \citep{absil2008optimization, boumal2020introduction}.
Besides being computationally efficient, this solution also 
has the advantage of not being an approximation of the original problem, nor requiring extra tuning parameters.

For \textsf{GGM} and \textsf{EGM}, the resolution of \eqref{eq:global_problem} requires to consider optimization on $\mathcal{S}_{p}^{++}$.
This will be presented in Section \ref{sec:optim_sp},
which will serve as both a short introduction to optimization on smooth Riemannian manifold, and as a building block for solving \eqref{eq:global_problem} for \textsf{GGFM} and \textsf{EGFM}. For these models, we will derive Riemannian optimization algorithms on $\MAT{\Sigma}\in\mathcal{M}_{p,k}$ in Section \ref{sec:optim_man}, which will be done by leveraging a suitable geometry for this space.
The corresponding algorithms are summarized in \autoref{algo:MainAlgo}.

\subsection{Learning GGM/EGM: optimization on $\mathcal{S}_{p}^{++}$}
\label{sec:optim_sp}

When relaxing the constraint in \eqref{eq:global_problem}, the problem still requires to be solved on $\mathcal{S}^{++}_p$.
Since $\mathcal{S}^{++}_p$ is open in $\mathcal{S}_p$ the tangent space $T_{\MAT{\Sigma}}\mathcal{S}^{++}_p$ can simply be identified as $\mathcal{S}_p$, for any $\MAT{\Sigma}\in\mathcal{S}^{++}_p$.
To be able to perform optimization, the first step is to define a Riemannian metric, \emph{i.e.}, an inner product on this tangent space.
In the case of $\mathcal{S}^{++}_p$, the most natural choice is the affine-invariant metric~\citep{skovgaard1984riemannian,bhatia2009positive}, that corresponds to the Fisher information metric of the Gaussian distribution and features very interesting properties from a geometrical point of view.
It is defined for all $\MAT{\Sigma}\in\mathcal{S}^{++}_p$, $\MAT{\xi}, \, \MAT{\eta}\in\mathcal{S}_p$ as
\begin{equation}
	\langle\MAT{\xi},\MAT{\eta}\rangle^{\mathcal{S}^{++}_p}_{\MAT{\Sigma}} = \tr(\MAT{\Sigma}^{-1}\MAT{\xi}\MAT{\Sigma}^{-1}\MAT{\eta}).
\end{equation}
The Riemannian gradient~\citep{absil2008optimization} at $\MAT{\Sigma}\in\mathcal{S}^{++}_p$ of an objective function $f:\mathcal{S}^{++}_p\to\mathbb{R}$ is the only tangent vector such that for all $\MAT{\xi}\in\mathcal{S}_p$,
\begin{equation}
	\langle\nabla^{\mathcal{S}^{++}_p}f(\MAT{\Sigma}),\MAT{\xi}\rangle^{\mathcal{S}^{++}_p}_{\MAT{\Sigma}} = \D f(\MAT{\Sigma})[\MAT{\xi}],
	\label{eq:grad_def}
\end{equation}
where $\D f(\MAT{\Sigma})[\MAT{\xi}]$ denotes the directional derivative of $f$ at $\MAT{\Sigma}$ in the direction $\MAT{\xi}$.
The Riemannian gradient $\nabla^{\mathcal{S}^{++}_p}f(\MAT{\Sigma})$ can also be obtained from the Euclidean gradient $\nabla^{\mathcal{E}}f(\MAT{\Sigma})$ through the formula
\begin{equation}
	\nabla^{\mathcal{S}^{++}_p}f(\MAT{\Sigma}) = \MAT{\Sigma} \symm(\nabla^{\mathcal{E}}f(\MAT{\Sigma})) \MAT{\Sigma},
	\label{eq:egrad2rgrad_SPD}
\end{equation}
where $\symm(\cdot)$ returns the symmetrical part of its argument.
The Riemannian gradient is sufficient in order to define a descent direction of $f$, \emph{i.e.}, a tangent vector inducing a decrease of $f$, hence to define a proper Riemannian steepest descent algorithm.
However, if one wants to resort to a more sophisticated optimization algorithm such as conjugate gradient or BFGS, vector transport ~\citep{absil2008optimization} is required to transport a tangent vector from one tangent space to another.
In the case of $\mathcal{S}^{++}_p$, we can employ the most natural one, \emph{i.e.}, the one corresponding to the parallel transport associated with the affine-invariant metric~\citep{jeuris2012survey}.
The transport of the tangent vector $\MAT{\xi}$ of $\MAT{\Sigma}$ onto the tangent space of $\MAT{\overbar{\Sigma}}$ is given by
\begin{equation}  \label{eq:transport_spd}
	\mathcal{T}^{\mathcal{S}^{++}_p}_{\MAT{\Sigma}\to\MAT{\overbar{\Sigma}}}(\MAT{\xi}) = (\MAT{\overbar{\Sigma}}\MAT{\Sigma}^{-1})^{\nicefrac12} \MAT{\xi} (\MAT{\Sigma}^{-1}\MAT{\overbar{\Sigma}})^{\nicefrac12}.
\end{equation}
Now that we have all the tools needed to obtain a proper descent direction of some objective function $f$, it remains to be able to get from the tangent space back onto the manifold $\mathcal{S}^{++}_p$. This is achieved by means of a retraction map.
The best solution on $\mathcal{S}^{++}_p$ in order to ensure numerical stability while taking into account the chosen geometry is
\begin{equation} \label{eq:retraction_spd}
	R^{\mathcal{S}^{++}_p}_{\MAT{\Sigma}}(\MAT{\xi}) = \MAT{\Sigma} + \MAT{\xi} + \frac12\MAT{\xi}\MAT{\Sigma}^{-1}\MAT{\xi}.
\end{equation}
This retraction corresponds to a second order approximation of the geodesics of $\mathcal{S}^{++}_p$~\citep{jeuris2012survey}, which generalize the concept of straight lines for a manifold.
We now have all the necessary elements to perform the optimization of $f$ on $\mathcal{S}^{++}_p$.
For instance, the sequence of iterates $\{\MAT{\Sigma}_t\}$ and descent directions $\{\MAT{\xi}_t\}$ generated by a Riemannian conjugate gradient algorithm is
\begin{equation} \label{eq:riem_grad_spd}
	\begin{array}{l}
		\MAT{\Sigma}_{t+1} = R^{\mathcal{S}^{++}_p}_{\MAT{\Sigma}_t}(\MAT{\xi}_t)
		\\[3pt]
		\MAT{\xi}_t = \alpha_t(-\nabla^{\mathcal{S}^{++}_p}f(\MAT{\Sigma}_t) + \beta_t\mathcal{T}^{\mathcal{S}^{++}_p}_{\MAT{\Sigma}_{t-1}\to\MAT{\Sigma}_t}(\MAT{\xi}_{t-1})),
	\end{array}
\end{equation}
where $\alpha_t$ is a stepsize that can be computed through a linesearch~\citep{absil2008optimization} and $\beta_t$ can be computed using the rule in~\citep{hestenes1952methods}.

From there, to obtain a specific algorithm that solves~\eqref{eq:global_problem} on $\mathcal{S}^{++}_p$, it only remains to provide the Riemannian gradients of the negative log-likelihood $\mathcal{L}$ and penalty $h$. 
The Riemannian gradient of $\mathcal{L}$ at $\MAT{\Sigma}\in\mathcal{S}^{++}_p$ is
\begin{equation} \label{eq:riem_grad_ggm1}
	\nabla^{\mathcal{S}^{++}_p}\mathcal{L}(\MAT{\Sigma}) = \frac12\MAT{\Sigma} - \frac1{2n}\textstyle\sum_i u(\VEC{x}_i^{\top}\MAT{\Sigma}^{-1}\VEC{x}_i) \VEC{x}_i\VEC{x}_i^{\top},
\end{equation}
where $u(t)=-2g'(t)/g(t)$.
For the Gaussian distribution, we have $u(t)=1$.
For the $t$-distribution with degree of freedom $\nu$, we have $u(t)=(\nu+p)/(\nu+t)$.
Concerning $h$, the $q\ell$ element of its Riemannian gradient at $\MAT{\Sigma}\in\mathcal{S}^{++}_p$~is
\begin{equation} \label{eq:riem_grad_ggm2}
	[\nabla^{\mathcal{S}^{++}_p} h(\MAT{\Sigma})]_{q\ell} = 
	\begin{cases}
		0 & \text{if} \, q=\ell \\
		\phi'([\MAT{\Sigma}^{-1}]_{q\ell}) & \text{if} \, q\neq\ell,
	\end{cases}
\end{equation}
where $\phi'(t) = \tanh(t/\varepsilon)$.
In the next section, the Euclidean gradients of $\mathcal{L}$ and $h$ in $\mathcal{S}^{++}_p$ are used.
They are easily obtained from~\eqref{eq:egrad2rgrad_SPD} and by noticing that the Euclidean gradient is already symmetric.

\subsection{Learning GGFM/EGFM: optimization on $\mathcal{M}_{p,k}$}
\label{sec:optim_man}

When considering the factor model, we aim at finding a structured matrix $\MAT{\Sigma}$ in $\mathcal{M}_{p,k}$, which is a smooth submanifold of $\mathcal{S}^{++}_p$.
The Riemannian geometry of $\mathcal{M}_{p,k}$ is not straightforward.
In fact, defining an adequate geometry for low-rank matrices is a rather difficult task and, while there were many attempts, see \emph{e.g.},~\citep{bonabel2009riemannian, meyer2011regression, vandereycken2012riemannian, massart2018quotient, neuman2021restricted, bouchard2021riemannian}, no perfect solution has been found yet.
Here, we choose the parametrization considered in~\citep{bonabel2009riemannian, meyer2011regression, bouchard2021riemannian}.
From a geometrical perspective, the novelty here is to adopt the particular structure of $\mathcal{M}_{p,k}$, \emph{i.e.}, by combining a low-rank positive semi-definite matrix and a diagonal positive definite matrix.
Each of the two manifolds of this product are well-known and the proofs of the following can easily be deduced from the proofs in~\citep{bouchard2021riemannian}.
All $\MAT{\Sigma}\in\mathcal{M}_{p,k}$ can be written~as
\begin{equation}
	\MAT{\Sigma} = \MAT{V}\MAT{\Lambda}\MAT{V}^{\top} + \MAT{\Psi},
\end{equation}
where $\MAT{V}\in\textup{St}_{p,k}=\{\MAT{V}\in\mathbb{R}^{p\times k}: \, \MAT{V}^{\top}\MAT{V}=\MAT{I}_k\}$ (Stiefel manifold of $p\times k$ orthogonal matrices), $\MAT{\Lambda}\in\mathcal{S}^{++}_k$ and $\MAT{\Psi}\in\mathcal{D}^{++}_p$.
Let $\mathcal{N}_{p,k}=\textup{St}_{p,k}\times\mathcal{S}^{++}_k\times\mathcal{D}^{++}_p$ and
\begin{equation}
	\begin{array}{rrcl}
		\varphi : & \mathcal{N}_{p,k} & \to & \mathcal{S}^{++}_p  \\
		& (\MAT{V}, \MAT{\Lambda}, \MAT{\Psi}) & \mapsto & \MAT{V}\MAT{\Lambda}\MAT{V}^{\top} + \MAT{\Psi}.
	\end{array}
\end{equation}
It follows that $\mathcal{M}_{p,k} = \varphi(\mathcal{N}_{p,k})$.
Therefore, to solve~\eqref{eq:global_problem} on $\mathcal{M}_{p,k}$, one can exploit $\varphi$ and solve it on $\mathcal{N}_{p,k}$.
However, $\mathcal{N}_{p,k}$ contains invariance classes with respect to $\varphi$, that is
for any $\MAT{O}\in\mathcal{O}_k$ (orthogonal group with $k\times k$ matrices),
\begin{equation}
	\varphi(\MAT{V}\MAT{O},\MAT{O}^{\top}\MAT{\Lambda}\MAT{O},\MAT{\Psi}) = \varphi(\MAT{V}, \MAT{\Lambda}, \MAT{\Psi}).
\end{equation}
As a consequence, the space that best corresponds to $\mathcal{M}_{p,k}$ is the quotient manifold $\mathcal{N}_{p,k}/\mathcal{O}_k$ induced by equivalence classes on $\mathcal{N}_{p,k}$
\begin{equation}
	\pi(\MAT{V}, \MAT{\Lambda}, \MAT{\Psi}) = \{ (\MAT{V}, \MAT{\Lambda}, \MAT{\Psi}) * \MAT{O} : \, \MAT{O}\in\mathcal{O}_k \},
	\label{eq:equivalence_class}
\end{equation}
where $(\MAT{V}, \MAT{\Lambda}, \MAT{\Psi}) * \MAT{O} = (\MAT{V}\MAT{O},\MAT{O}^{\top}\MAT{\Lambda}\MAT{O},\MAT{\Psi})$.
To efficiently solve an optimization problem on $\mathcal{M}_{p,k}$ with the chosen parametrization, we thus need to consider the quotient $\mathcal{N}_{p,k}/\mathcal{O}_k$.
In practice, rather than dealing with the quite abstract manifold $\mathcal{N}_{p,k}/\mathcal{O}_k$ directly, we will manipulate objects in $\mathcal{N}_{p,k}$, implying that the different points of an equivalence class~\eqref{eq:equivalence_class} are, in fact, one and only point.

Our first task is to provide the tangent space of $\theta=(\MAT{V},\MAT{\Lambda},\MAT{\Psi})\in\mathcal{N}_{p,k}$.
It is obtained by aggregating tangent spaces $T_{\MAT{V}}\textup{St}_{p,k}$, $T_{\MAT{\Lambda}}\mathcal{S}^{++}_{k}$ and $T_{\MAT{\Psi}}\mathcal{D}^{++}_{p}$, \emph{i.e.},
\begin{equation}
	T_{\theta}\mathcal{N}_{p,k} =
	\{ (\MAT{\xi}_{\MAT{V}},\MAT{\xi}_{\MAT{\Lambda}},\MAT{\xi}_{\MAT{\Psi}})\in\mathbb{R}^{p\times k}\times\mathcal{S}_k\times\mathcal{D}_p:
	\MAT{V}^{\top}\MAT{\xi}_{\MAT{V}} + \MAT{\xi}_{\MAT{V}}^{\top}\MAT{V} = \MAT{0} \}.
\end{equation}
Then, we define a Riemannian metric on $\mathcal{N}_{p,k}$.
It needs to be invariant along equivalence classes, \emph{i.e.}, for all $\theta\in\mathcal{N}_{p,k}$, $\xi$, $\eta\in T_{\theta}\mathcal{N}_{p,k}$ and $\MAT{O}\in\mathcal{O}_k$
\begin{equation}
	\langle\xi,\eta\rangle^{\mathcal{N}_{p,k}}_{\theta} = 
	\langle \xi * \MAT{O} , \eta * \MAT{O} \rangle^{\mathcal{N}_{p,k}}_{\theta * \MAT{O}}.
\end{equation}
In this work, the Riemannian metric $\mathcal{N}_{p,k}$ is chosen as the sum of metrics on $\textup{St}_{p,k}$, $\mathcal{S}^{++}_k$ and $\mathcal{D}^{++}_p$.
The metric on $\textup{St}_{p,k}$ is the so-called canonical metric~\citep{edelman1998geometry}, which yields the simplest geometry of $\textup{St}_{p,k}$.
The metrics on $\mathcal{S}^{++}_k$ and $\mathcal{D}^{++}_p$ are the affine-invariant ones.
Hence, our chosen metric on $\mathcal{N}_{p,k}$ is
\begin{equation}
	\langle\xi,\eta\rangle^{\mathcal{N}_{p,k}}_{\theta} =
	\tr(\MAT{\xi}_{\MAT{V}}^{\top}(\MAT{I}_p - \frac12\MAT{V}\MAT{V}^{\top})\MAT{\eta}_{\MAT{V}})
	+ \tr(\MAT{\Lambda}^{-1}\MAT{\xi}_{\MAT{\Lambda}}\MAT{\Lambda}^{-1}\MAT{\eta}_{\MAT{\Lambda}})
	+ \tr(\MAT{\Psi}^{-2}\MAT{\xi}_{\MAT{\Psi}}\MAT{\eta}_{\MAT{\Psi}}).
	\label{eq:metric_factor}
\end{equation}
At $\theta\in\mathcal{N}_{p,k}$, $T_{\theta}\mathcal{N}_{p,k}$ contains tangent vectors inducing a move along the equivalence class $\pi(\theta)$.
In our setting, these directions are to be eliminated.
They are contained in the so-called vertical space $\mathcal{V}_{\theta}$, which is the tangent space $T_{\theta}\pi(\theta)$ to the equivalence class.
In our case, it is
\begin{equation}
	\mathcal{V}_{\theta}=\{ (\MAT{V}\MAT{\Omega}, \MAT{\Lambda}\MAT{\Omega} - \MAT{\Omega}\MAT{\Lambda}, \MAT{0}) : \, \MAT{\Omega}\in\mathcal{S}^{\perp}_k \},
\end{equation}
where $\mathcal{S}^{\perp}_k$ denotes the vector space of skew-symmetric matrices.
Now that the unwanted vectors have been identified, we can deduce the ones of interest: they are contained in the orthogonal complement of $\mathcal{V}_{\theta}$ according to metric~\eqref{eq:metric_factor}.
This space is called the horizontal space $\mathcal{H}_{\theta}$ and is equal to
\begin{equation}
	\mathcal{H}_{\theta} = \{ \xi\in T_{\theta}\mathcal{N}_{p,k} : \, \MAT{V}^{\top}\MAT{\xi}_{\MAT{V}} = 2(\MAT{\Lambda}^{-1}\MAT{\xi}_{\MAT{\Lambda}} - \MAT{\xi}_{\MAT{\Lambda}}\MAT{\Lambda}^{-1}) \}.
	\label{eq:horizontal_space}
\end{equation}
The Riemannian gradient of an objective function $\overbar{f}$ on $\mathcal{N}_{p,k}$ is defined through the chosen metric as for $\mathcal{S}^{++}_p$ in~\eqref{eq:grad_def}.
Again, it is possible to obtain the Riemannian gradient $\nabla^{\mathcal{N}_{p,k}} \overbar{f}(\theta)$ on $\mathcal{N}_{p,k}$ from the Euclidean one $\nabla^{\mathcal{E}} \overbar{f}(\theta)$.
This is achieved by
\begin{equation}
	\nabla^{\mathcal{N}_{p,k}} \overbar{f}(\theta) =
	(
	\MAT{G}_{\MAT{V}} - \MAT{V}\MAT{G}_{\MAT{V}}^{\top}\MAT{V},
	\MAT{\Lambda}\MAT{G}_{\MAT{\Lambda}}\MAT{\Lambda},
	\MAT{\Psi}^2\ddiag(\MAT{G}_{\MAT{\Psi}})
	),
\end{equation}
where $\nabla^{\mathcal{E}} \overbar{f}(\theta)=(\MAT{G}_{\MAT{V}},\MAT{G}_{\MAT{\Lambda}},\MAT{G}_{\MAT{\Psi}})$ and $\ddiag(\cdot)$ cancels the off-diagonal elements of its argument.
Furthermore, since the objective functions we are interested in are invariant along equivalence classes~\eqref{eq:equivalence_class}, their Riemannian gradient is naturally on the horizontal space $\mathcal{H}_{\theta}$ and we can use it directly.
In this work, our focus is objective functions $\overbar{f}:\mathcal{N}_{p,k}\to\mathbb{R}$ such that $\overbar{f}=f\circ\varphi$, where $f:\mathcal{S}^{++}_p\to\mathbb{R}$; see~\eqref{eq:global_problem}.
Hence, it is of great interest to be able to get the gradient of $\overbar{f}$ directly from the one of $f$.
The Euclidean gradient $\nabla^{\mathcal{E}} \overbar{f}(\theta)$ as a function of the Euclidean gradient $\nabla^{\mathcal{E}} f(\varphi(\theta))$ is 
\begin{equation} \label{eq:riem_grad_mpk}
	\nabla^{\mathcal{E}} \overbar{f}(\theta) = (2\nabla^{\mathcal{E}} f(\varphi(\theta))\MAT{V}\MAT{\Lambda}, \MAT{V}^{\top}\nabla^{\mathcal{E}} f(\varphi(\theta))\MAT{V}, \ddiag(\nabla^{\mathcal{E}} f(\varphi(\theta)))).
\end{equation}
While it is not necessary for the gradient, the vector transport requires to be able to project elements from the ambient space $\mathbb{R}^{p\times k}\times\mathbb{R}^{k\times k}\times\mathbb{R}^{p\times p}$ onto the horizontal space. 
The first step is to provide the orthogonal projection from the ambient space onto $T_{\theta}\mathcal{N}_{p,k}$.
For all $\theta\in\mathcal{N}_{p,k}$ and $\xi\in\mathbb{R}^{p\times k}\times\mathbb{R}^{k\times k}\times\mathbb{R}^{p\times p}$
\begin{equation}
	P^{\mathcal{N}_{p,k}}_{\theta}(\xi) = 
	(
	\MAT{\xi}_{\MAT{V}} - \MAT{V}\symm(\MAT{V}^{\top}\MAT{\xi}_{\MAT{V}}),
	\symm(\MAT{\xi}_{\MAT{\Lambda}}),
	\ddiag(\MAT{\xi}_{\MAT{\Psi}})
	).
\end{equation}
From there we can obtain the orthogonal projection from $T_{\theta}\mathcal{N}_{p,k}$ onto $\mathcal{H}_{\theta}$.
Given $\xi\in T_{\theta}\mathcal{N}_{p,k}$, it is
\begin{equation}
	\mathcal{P}^{\mathcal{N}_{p,k}}_{\theta}(\xi) =
	(
	\MAT{\xi}_{\MAT{V}} - \MAT{V}\MAT{\Omega},
	\MAT{\xi}_{\MAT{\Lambda}} + \MAT{\Omega}\MAT{\Lambda} - \MAT{\Lambda}\MAT{\Omega},
	\MAT{\xi}_{\MAT{\Psi}}
	),
\end{equation}
where $\MAT{\Omega}\in\mathcal{S}^{\perp}_k$ is the unique solution to
\begin{equation}
	2(\MAT{\Lambda}^{-1}\MAT{\Omega}\MAT{\Lambda} + \MAT{\Lambda}\MAT{\Omega}\MAT{\Lambda}^{-1}) - 3\MAT{\Omega} = 
	\MAT{V}^{\top}\MAT{\xi}_{\MAT{V}} + 2(\MAT{\xi}_{\MAT{{\Lambda}}}\MAT{\Lambda}^{-1} + \MAT{\Lambda}^{-1}\MAT{\xi}_{\MAT{{\Lambda}}}).
\end{equation}
We can now define an adequate vector transport operator.
Given $\theta$, $\overbar{\theta}\in\mathcal{N}_{p,k}$ and $\xi\in\mathcal{H}_{\theta}$, it is simply
\begin{equation} \label{eq:transport_mpk}
	\mathcal{T}^{\mathcal{N}_{p,k}}_{\theta\to\overbar{\theta}}(\xi) = \mathcal{P}^{\mathcal{N}_{p,k}}_{\overbar{\theta}}(P^{\mathcal{N}_{p,k}}_{\overbar{\theta}}(\xi)).
\end{equation}
The last object that we need is a retraction on $\mathcal{N}_{p,k}$.
To address the invariance requirement, the retraction needs to be invariant along equivalence classes, \emph{i.e.}, for all $\theta\in\mathcal{N}_{p,k}$, $\xi\in\mathcal{H}_{\theta}$ and $\MAT{O}\in\mathcal{O}_k$
\begin{equation}
	R^{\mathcal{N}_{p,k}}_{\theta*\MAT{O}}(\xi*\MAT{O}) = R^{\mathcal{N}_{p,k}}_{\theta}(\xi).
\end{equation}
In this paper, we choose a second order approximation of geodesics on $\mathcal{N}_{p,k}$ given by
\begin{equation} \label{eq:retraction_mpk}
	R^{\mathcal{N}_{p,k}}_{\theta}(\xi) =
	(
	\uf(\MAT{V}+\MAT{\xi}_{\MAT{V}}),	
	\MAT{\Lambda}+\MAT{\xi}_{\MAT{\Lambda}}+\frac12\MAT{\xi}_{\MAT{\Lambda}}\Lambda^{-1}\MAT{\xi}_{\MAT{\Lambda}},
	\MAT{\Psi} + \MAT{\xi}_{\MAT{\Psi}} + \frac12\MAT{\xi}_{\MAT{\Psi}}^2\MAT{\Psi}^{-1}
	),
\end{equation}
where $\uf(\cdot)$ returns the orthogonal factor of the polar decomposition.

We finally have all the tools needed to solve problem~\eqref{eq:global_problem} on $\mathcal{N}_{p,k}$ and thus to ensure the structure of $\mathcal{M}_{p,k}$.
With all the objects given in this section, many Riemannian optimization algorithms can be employed to achieve the minimization~\eqref{eq:global_problem}, \emph{e.g.}, steepest descent, conjugate gradient or BFGS.
In this work, we use in practice a conjugate gradient algorithm.

\subsection{Algorithms properties}
In terms of convergence, the proposed algorithms inherit the standard properties from the Riemannian optimization framework \citep{absil2008optimization, boumal2020introduction}: each iterate satisfies the constraints and ensures a decrease of the objective function until a critical point is reached.
Due to the inversion of $\MAT{\Sigma}$ at each step, the computational complexity of the Riemannian conjugate gradient for \textsf{GGM}/\textsf{EGM} is $\mathcal{O}(T p^3)$, where $T$ is the number of iterations.
This complexity has the same dependence in $p$ than GLasso \citep{friedman2008sparse} and more recent structured graph learning algorithms \citep{kumar2020unified}.
On the other hand, by relying on the structure of the factor models, the Riemannian conjugate gradient for \textsf{GGFM}/\textsf{EGFM} has a $\mathcal{O}(T (pk^2 + k^3))$ computational complexity\footnote{
	Note that \textsf{GGFM/EGFM} require the computation of the sample covariance matrix (SCM), which is in $\mathcal{O}(np^2)$.
	\textsf{EGM/EGFM} require to re-compute a weighted SCM at each step (i.e., $\mathcal{O}(T np^2)$).
	However, these operations consist only in matrix multiplications that can be parallelized, thus are not actual computational bottlenecks.
}. 
As we usually set $k\ll p$, \textsf{EGM/EGFM} are therefore more suited to large dimensions.

While the question of statistical properties will not be explored in this paper, it is also worth mentioning that \eqref{eq:global_problem} appears as a regularized maximum likelihood estimation problem for elliptical models.
Consequently, the algorithms presented in \autoref{algo:MainAlgo} aim at solving a class of maximum a posteriori where a prior related to $\phi$ is set independently on the elements $\MAT{\Theta}_{ij}$.
Thus, one can expect good statistical performance if the data falls within the class of elliptical distribution, and especially the algorithms to be robust to a mismatch related to the choice of $g$ \citep{maronna1976robust, dravskovic2018new}.

\section{Experiments}

\subsection{Validations on synthetic data}

Due to space limitations, extensive experiments on synthetic data (ROC curves for edge detection, sensitivity to tuning parameters) are described in the supplementary material available in \autoref{sec:exp} at the end of this paper.
The  short conclusions of these experiments are the following:
$(i)$ the proposed methods compare favorably to existing methods in terms of ROC curves for many underlying random graphs models (Erdős-Rényi, Barabási–Albert, Watts–Strogatz, and random geometric graph), especially when the sample support is limited;
$(ii)$ For the aforementioned graphs models, the covariance matrix does not always exhibit a low-rank structure, so factor models do not necessarily improve the ROC performance.
However it is experienced to be extremely useful when applied to real-world data, as illustrated below (improved clustering capabilities);
$(iii)$ As all sparsity-promoting methods, the proposed algorithms require a proper set of the regularization parameter $\lambda$.
However, the results are not critically sensitive to a change of this value when its order of magnitude is correctly set. The code for these experiments is available at: \url{https://github.com/ahippert/graphfactormodel}.

\subsection{Real-world data sets}
\label{subsec:real_data}

In this section, the performances of the proposed algorithms (\textsf{GGM}/\textsf{GGFM} and \textsf{EGM}/\textsf{EGFM}) are evaluated on three real-world data sets consisting of $(i)$ animal species; $(ii)$ GNSS time-series; $(iii)$
concepts data. 
The purpose of the following experiments is to verify whether the factor model, which accounts for potential structures in the covariance matrix, provides an improved version of the learned graph topology. The latter is evaluated using both graph modularity $m$, for which high values measure high separability of a graph into sub-components \citep{newman2006modularity}, and visual inspection\footnote{Note that there is no available ground truth for these data sets. Hence, to facilitate visualization, each graph node is then clustered into colors using a community detection algorithm based on label propagation \citep{cordasco2010community}.}.\\

\noindent
\textit{Benchmarks and parameter setting:}
The proposed algorithms are compared with state-of-the-art approaches for
$(i)$ \textit{connected} graphs: \textsf{GLasso} \citep{friedman2008sparse}, which uses $\ell_1$-norm as a sparse-promoting penalty, and \textsf{NGL} \citep{ying2020nonconvex}, which use a Gaussian Laplacian-constrained model with a concave penalty regularization;    
$(ii)$ \textit{multi-component} graphs: \textsf{SGL} \citep{kumar2020unified}, which constraints the number of graph components in an \textsf{NGL}-type formulation, and \textsf{StGL} \citep{de2021graphical}, which generalizes the above to $t$-distributed data.
For a fair comparison, 
the parameters of all tested algorithms (including competing methods) are tuned to display the best results.
Notably, the proposed algorithms are set using \autoref{algo:MainAlgo} with 
a tolerance threshold of ${\rm tol} = 10^{-2}$.
For \textsf{EGM}/\textsf{EGFM}, the chosen density generator relates to a $t$-distribution with degrees of freedom $\nu=5$.
The rank of factor models is chosen manually. Still, the results were similar for other ranks around the set value.\\

%

\begin{figure*}[!t]
	\centering
	\begin{subfigure}{.2\linewidth}
		\centering
		\includegraphics[scale=.075]{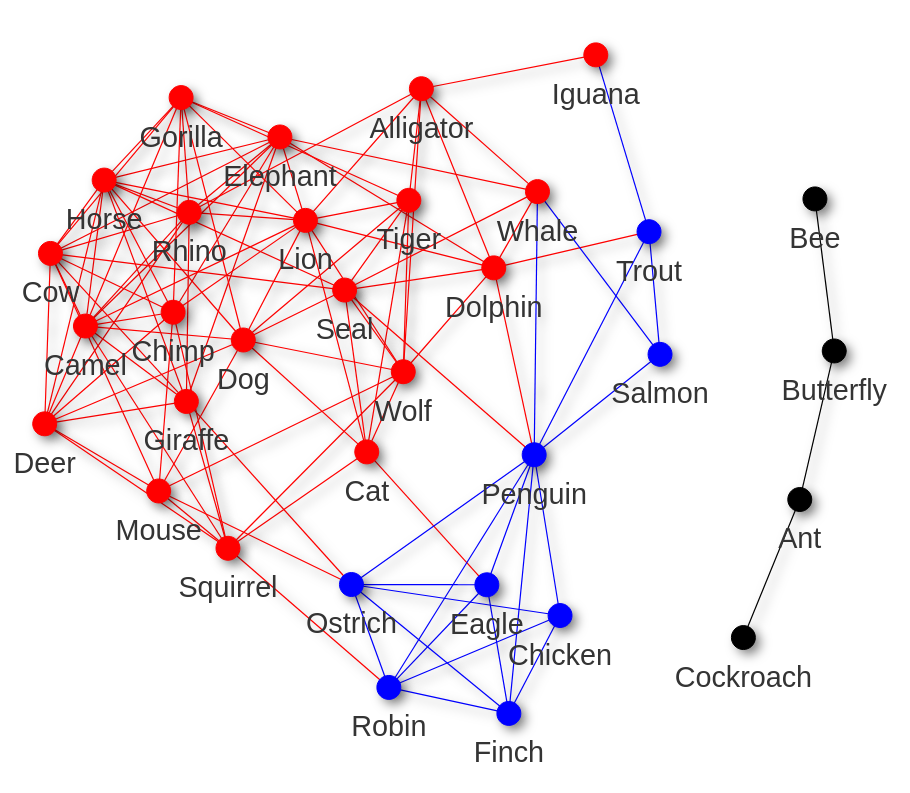}
		\caption{\tiny \textsf{GLasso} ($m=0.56$)}\label{fig:image1}
	\end{subfigure}
	\hfill
	\begin{subfigure}{.2\linewidth}
		\centering
		\includegraphics[scale=.075]{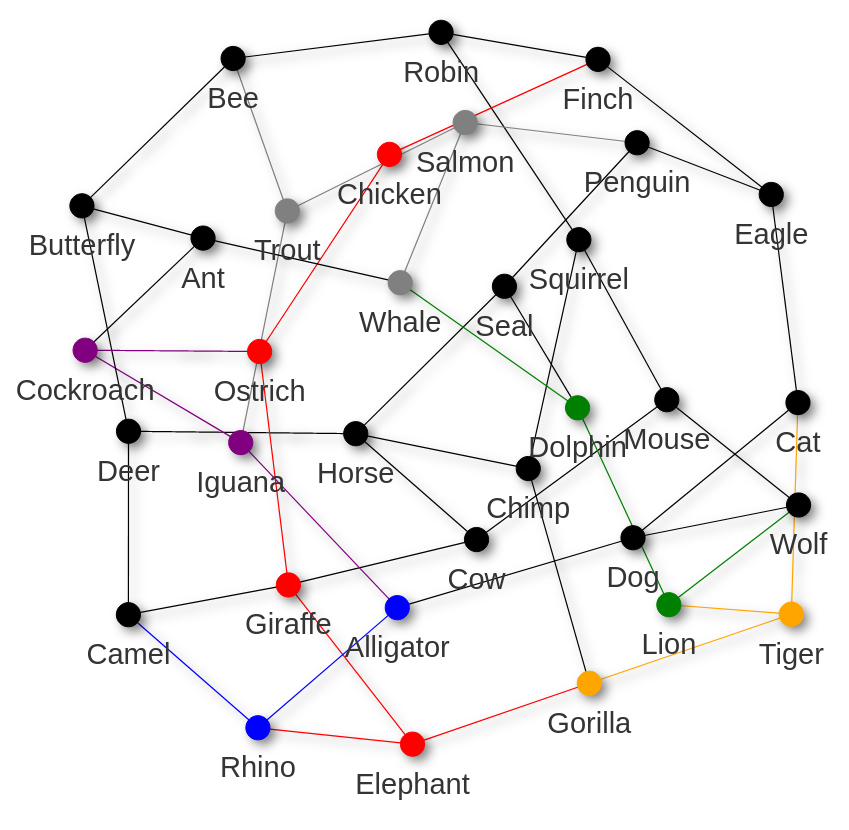}
		\caption{\tiny \textsf{NGL} ($m=0.38$)}\label{fig:image12}
	\end{subfigure}
	\hfill
	\begin{subfigure}{.25\linewidth}
		\centering
		\includegraphics[scale=.075]{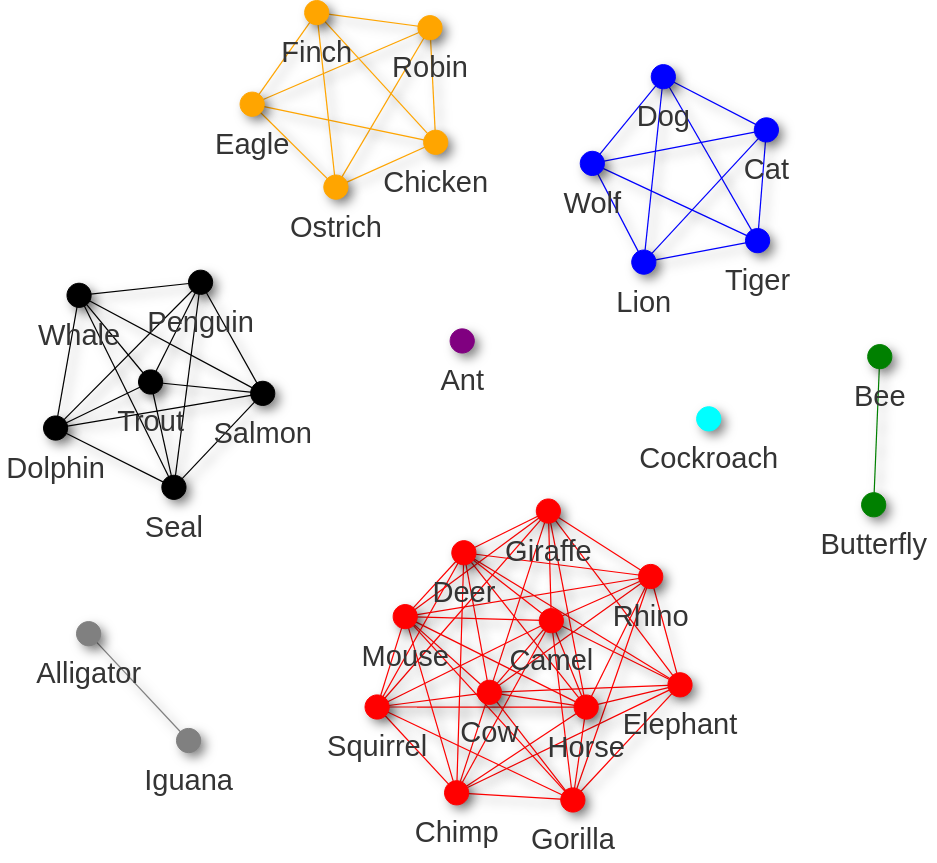}
		\caption{\tiny \textsf{SGL} ($m=0.74$)}\label{fig:image13}
	\end{subfigure}
	\begin{subfigure}{.25\linewidth}
		\centering
		\includegraphics[scale=.075]{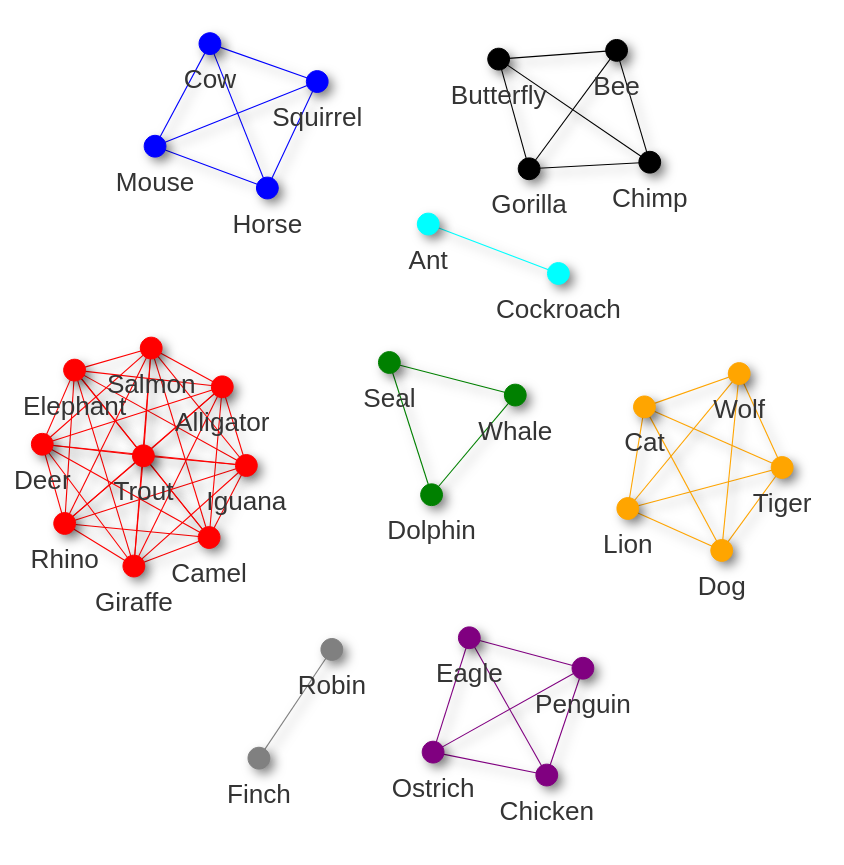}
		\caption{\tiny \textsf{StGL} ($m=0.84$)}\label{fig:image3}
	\end{subfigure}
	\begin{subfigure}{.24\linewidth}
		\centering
		\includegraphics[scale=.075]{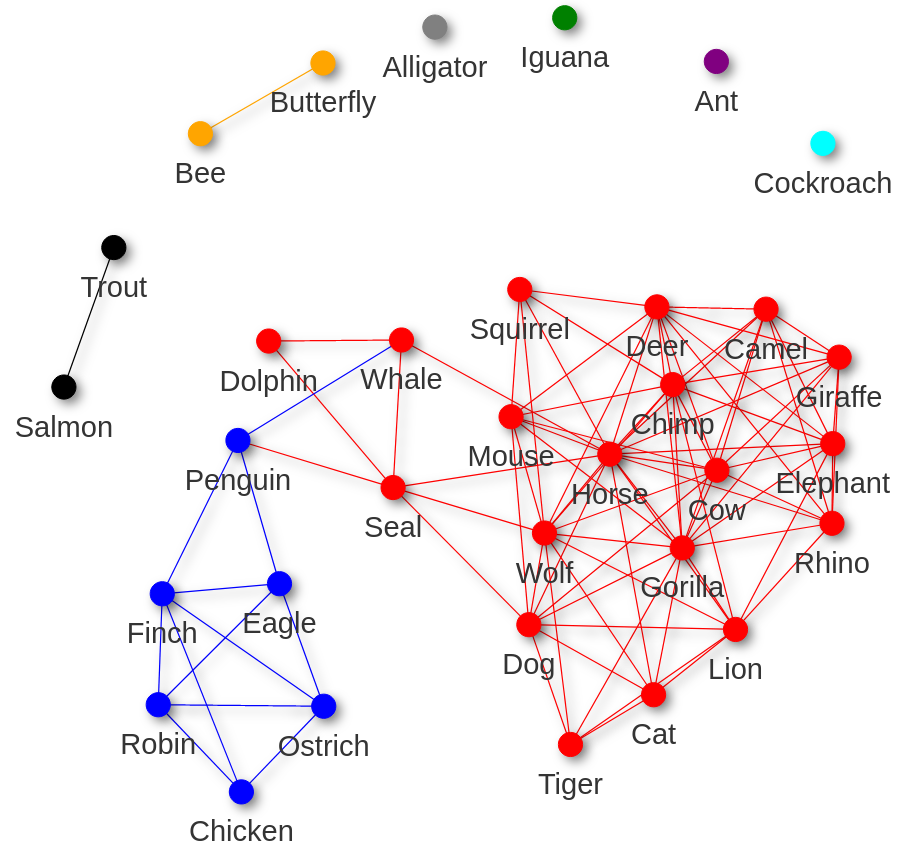}
		\caption{\tiny \textsf{GGM} ($m=0.54$)}\label{fig:image3}
	\end{subfigure}
	\hfill
	\begin{subfigure}{.24\linewidth}
		\centering
		\includegraphics[scale=.075]{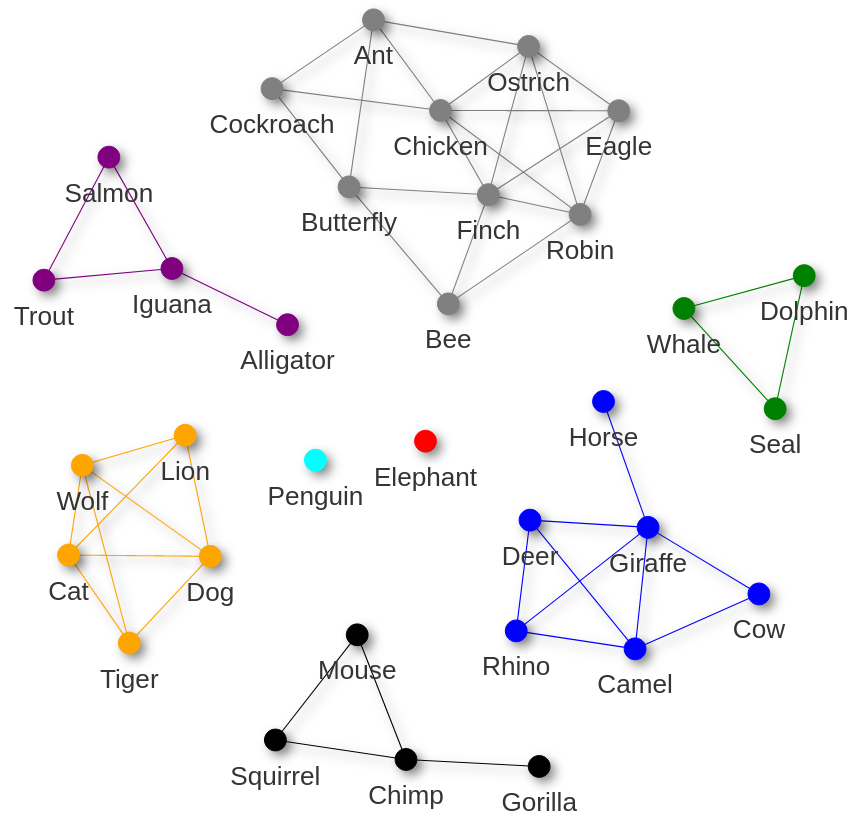}
		\caption{\tiny \textsf{GGFM} ($m=0.79$)}\label{fig:image3}
	\end{subfigure}
	\hfill
	\begin{subfigure}{.24\linewidth}
		\centering
		\includegraphics[scale=.075]{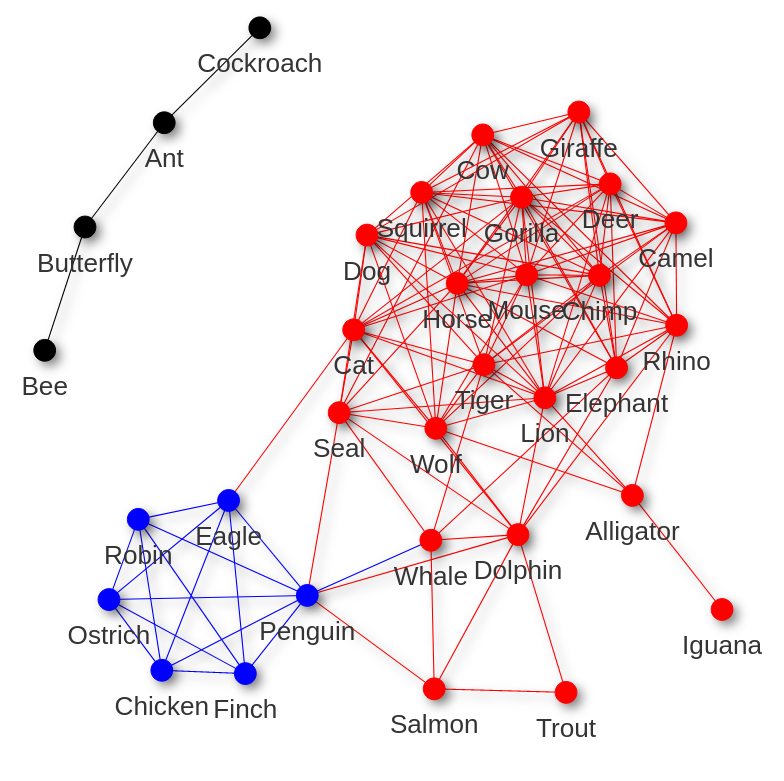}
		\caption{\tiny \textsf{EGM} ($m=0.44$)}\label{fig:image3}
	\end{subfigure}
	\hfill
	\begin{subfigure}{.24\linewidth}
		\centering
		\includegraphics[scale=.075]{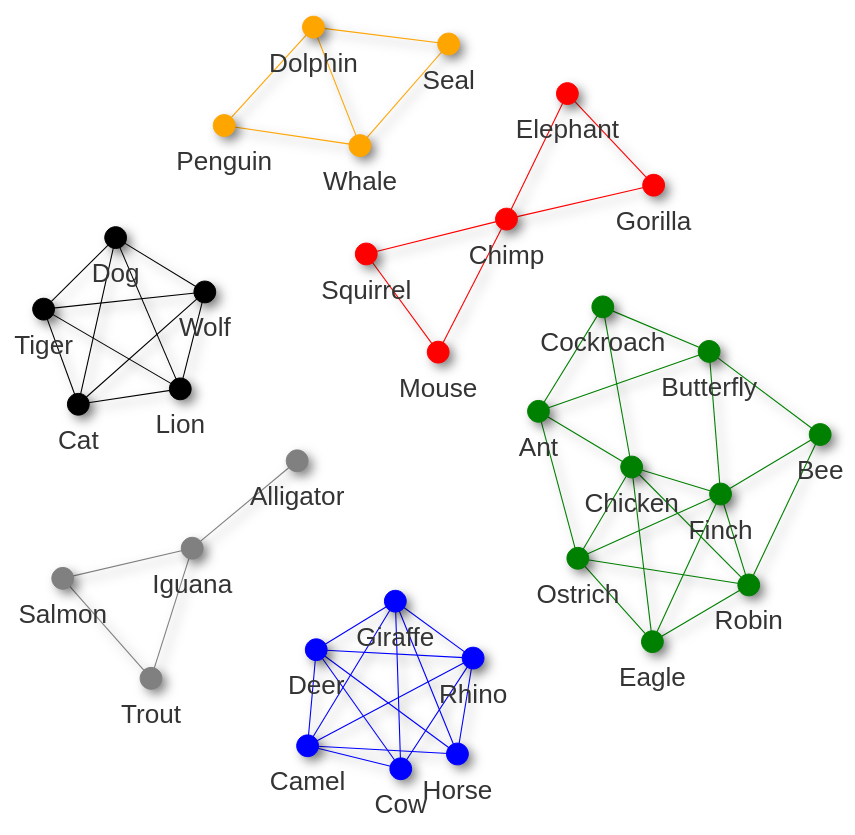}
		\caption{\tiny \textsf{EGFM} ($m=0.8$)}\label{fig:image3}
	\end{subfigure}
	\RawCaption{\caption{Learned graphs from the \textit{animals} data set  with various algorithms and their associated modularity $m$. For \textsf{GGFM}/\textsf{EGFM}, the rank is 10. For \textsf{SGL} and \textsf{StGL}, the number of components is fixed to 8.}
		\label{fig:animal_results}}
\end{figure*}

\noindent 
\textbf{\textit{Animals} data}: 
In the \textit{animals} data set \citep{osherson1991animal, lake2010discovering} each node represents an animal, and is associated with binary features (categorical non-Gaussian data) consisting of answers to questions such as ``has teeth?'', ``is poisonous?'', \emph{etc}. In total, there are $p=33$ unique animals and $n=102$ questions. \autoref{fig:animal_results} displays the learned graphs of the \textit{animals} data set. Following the recommendations of \citep{kumar2020unified, egilmez2017graph}, the \textsf{SGL} algorithm is applied with $\beta = 0.5$ and $\alpha=0$, \textsf{GLasso} with $\alpha=0.05$ and the input for \textsf{SGL} and \textsf{NGL} is set to $\frac{1}{n}\sum_{i}\VEC{x}_i^\top\VEC{x}_i + \frac{1}{3}\MAT{I}_p$. While \textsf{EGM} yields results similar to \textsf{GLasso}, \textsf{GGM} provides a clearer structure with (\textit{trout}, \textit{salmon}) and (\textit{bee}, \textit{butterfly}) clustered together. Interestingly, with no assumptions regarding the number of components, \textsf{GGFM} and \textsf{EGFM} reach similar structure and modularity compared to \textsf{SGL} and \textsf{StGL}, that require to set \textit{a priori} the number of components in the graph.\\

\begin{figure*}[!t]
	\begin{subfigure}{.325\textwidth}
		\includegraphics[trim=204 430 178 172, clip, width=\textwidth]{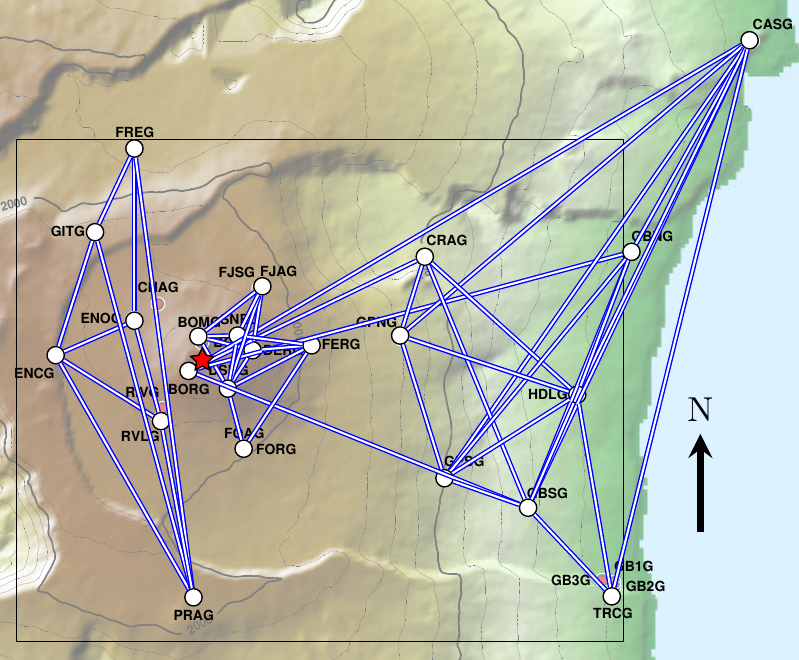}
		\caption{\textsf{StGL} ($m=0.66$)}
	\end{subfigure}
	\begin{subfigure}{.325\textwidth}
		\includegraphics[trim=204 430 178 172, clip, width=\textwidth]{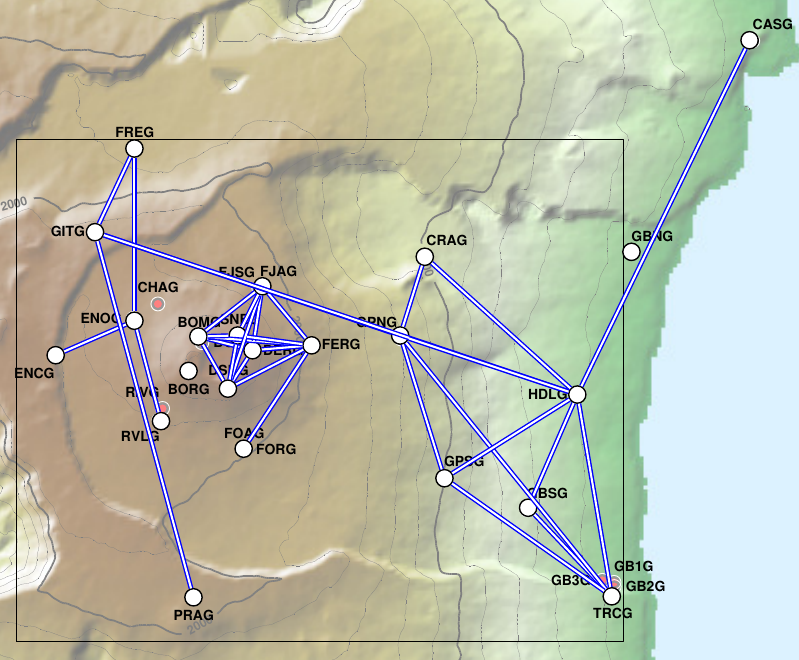}
		\caption{\textsf{GGM} ($m=0.51$)}
	\end{subfigure}
	\begin{subfigure}{.325\textwidth}
		\includegraphics[trim=204 430 178 172, clip, width=\textwidth]{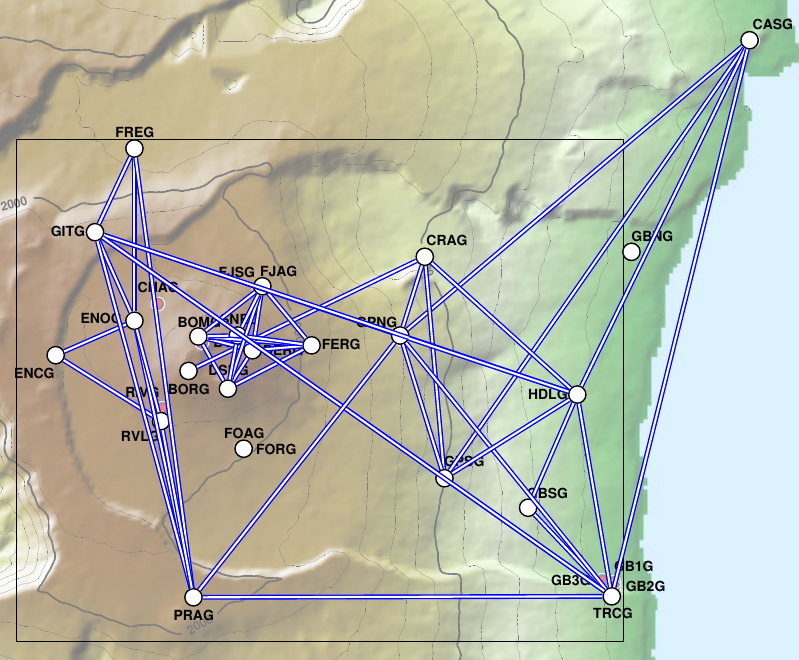}
		\caption{\textsf{EGM} ($m=0.59$)}
	\end{subfigure}
	\bigskip
	\begin{subfigure}{.325\textwidth}
		\includegraphics[trim=204 430 178 172, clip, width=\textwidth]{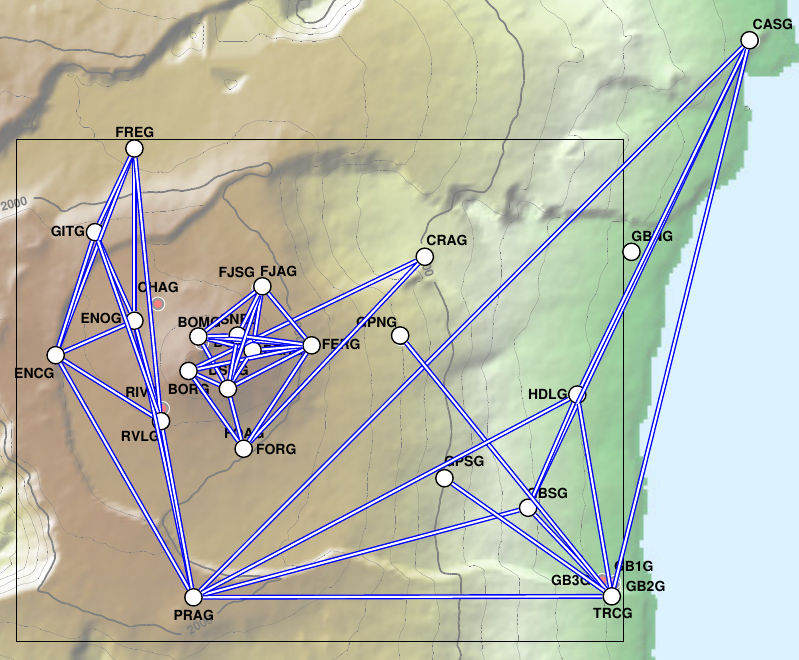}
		\caption{\textsf{GGFM} ($m=0.61$)}
	\end{subfigure}
	\begin{subfigure}{.325\textwidth}
		\includegraphics[trim=204 430 178 172, clip, width=\textwidth]{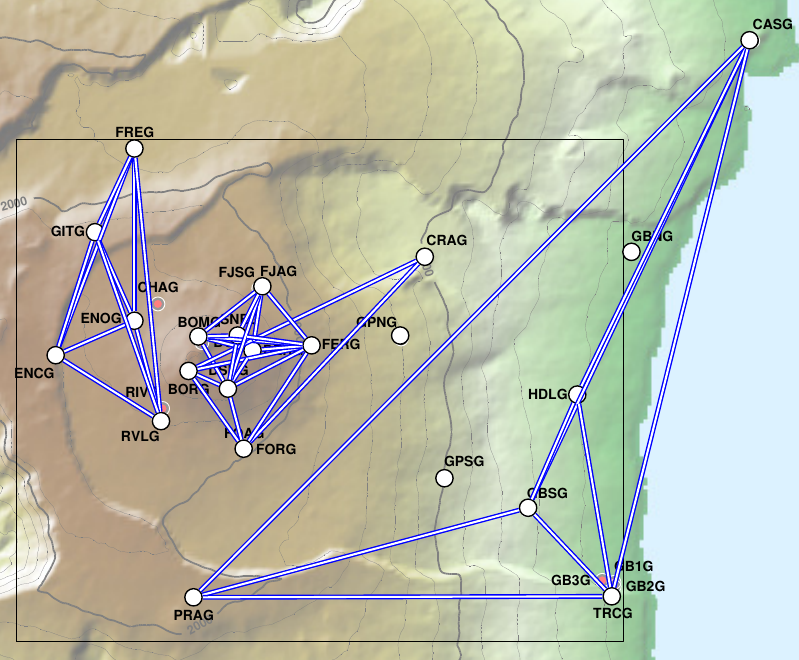}
		\caption{\textsf{EGFM} ($m=0.6$)}
	\end{subfigure}
	\caption{Learned graphs from the GNSS data with the \textsf{StGL} algorithm, our proposed approaches and their associated modularity $m$. For \textsf{StGL}, the number of components is fixed to 3. For \textsf{GGFM} and \textsf{EGFM}, the rank is fixed to 4. For the $t$-distribution, the degrees of freedom are set to $\nu=5$. The red star in (a) points to the summit crater. \textcopyright WEBOBS/IPGP.}
	\label{fig:results_gnss}
\end{figure*}

\noindent
\textbf{GNSS data}: We present an application on Earth surface displacement data collected by a network of Global Navigation Satellite System (GNSS) receivers from the volcanological observatory of Piton de la Fournaise (OVPF-IPGP), located at La Réunion Island. The presented network is composed of $p=22$ receivers recording daily vertical displacements from January 2014 to March 2017 \citep{smittarello2019magma}, with a total of $n=1106$ observations.
During this time, vertical displacements induced by volcano eruptions have been recorded by the receivers network, sometimes leading to abrupt motion changes over time. Depending on their spatial position, some receivers might move in a particular direction (upwards or downwards), thus indicating (thin) spatial correlations.
Results of the learned graphs are presented in \autoref{fig:results_gnss}. A general observation is that all graphs are mainly clustered into three groups: two located West (receivers `GITG', `FREG', etc.) and East (`BORG', `FJAG', etc.) of the summit crater, and one extending from lower altitudes to the seashore (`CASG', `TRCG', etc.). We call these groups \textit{west}, \textit{east} and \textit{low} components, respectively.
As described in \citep{peltier2017assessing}, the four 2015 eruptions (February, May, July and August) are characterized by asymmetric displacement patterns with respect to the North-South eruptive fissures which extend slightly westward from the summit crater. Interestingly, this corresponds to the separation between \textit{west} and \textit{east} graph components, which is best evidenced by factor model-based algorithms, especially \textsf{EGFM}. This result can be explained by the fact that \textsf{GGFM/EGFM} are more robust to abrupt changes in the data as it occurs in heavy-tailed data distributions. The \textit{low} graph component corresponds to receivers with small or no displacement. Note that the `PRAG' receiver is also included in this group, probably because it did not record a significant motion during this period. Finally, \textsf{GGM}, \textsf{EGM} and \textsf{StGL} lead to similar results but with spurious connections from the west side of the crater to seashore receivers (\textit{e.g.}, from `BORG' to `CASG' for \textsf{StGL}, and `PRAG' connected to some \textit{west} and/or \textit{east} receivers).\\

\begin{figure*}[!t]
	\includegraphics[width=\textwidth]{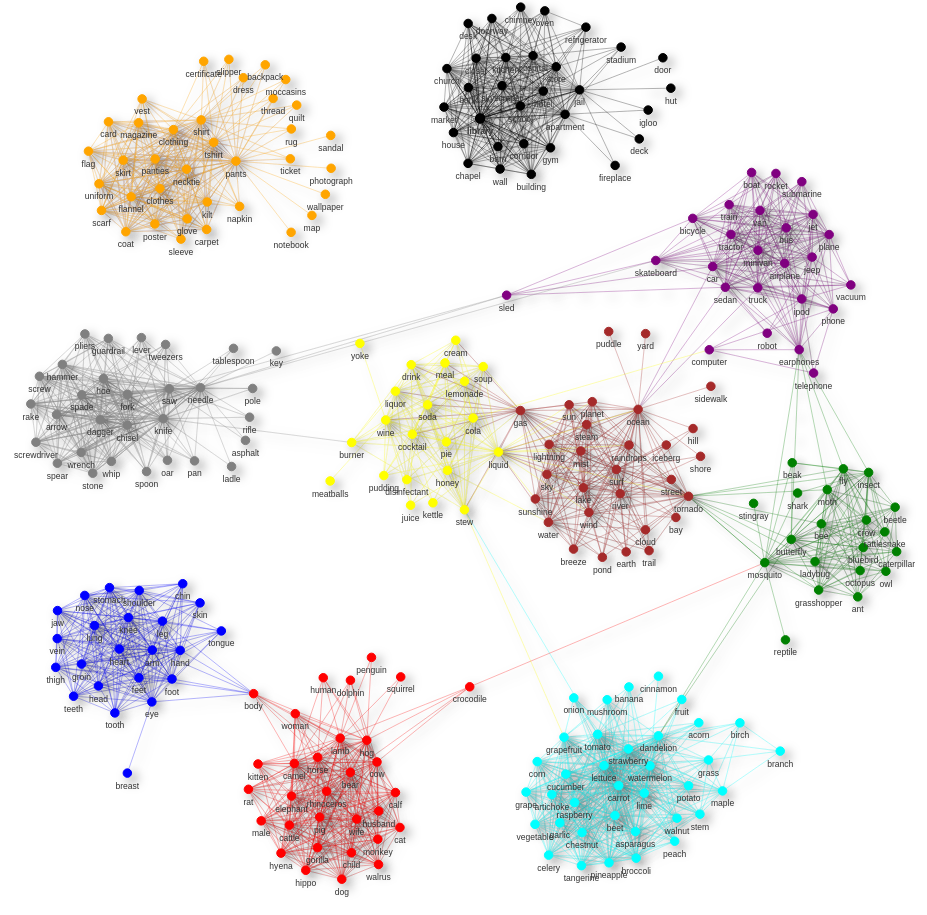}
	\caption{Learned graph with the \textsf{GGFM} algorithm ($k=10$ and $\lambda=3.5$). Note that black and orange graph components are disconnected from the rest. Isolated nodes have been removed by hand for better visualization;}
	\label{fig:concept_overview}
\end{figure*}

\begin{figure}[!t]
	\raisebox{-\height}{\includegraphics[width=\textwidth]{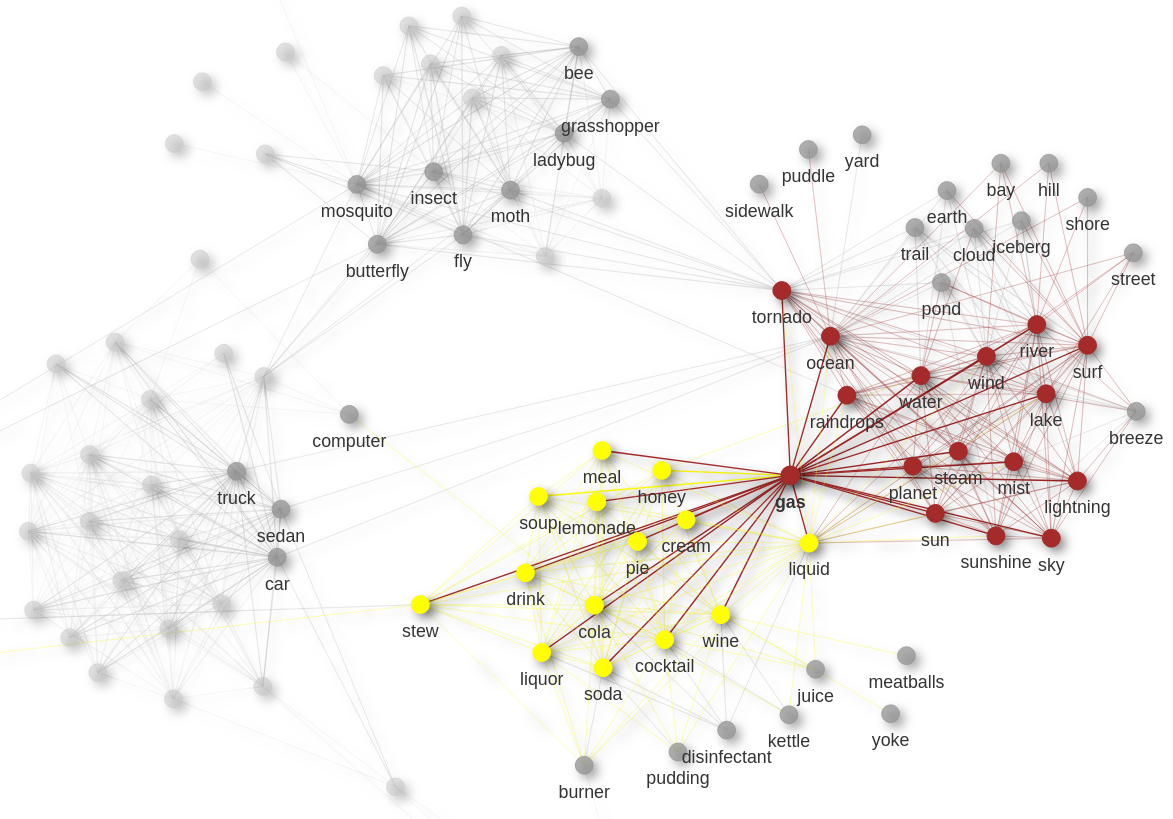}}
	\vspace{.6ex}
	\raisebox{-\height}{\includegraphics[width=\textwidth]{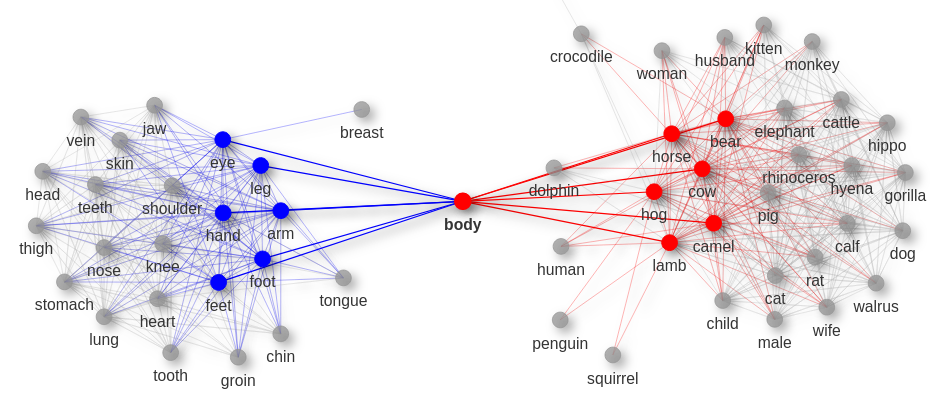}}
	\caption{First (colored) and second (greyed with label) neighbors of nodes `body' and `gas'. These concepts connect two graph components each: body parts and whole bodies for `body'; drinks and natural elements for `gas'.}
	\label{fig:concept_zoom}
\end{figure}

\noindent
\textbf{Concepts data}: The \textit{concepts} data set \citep{lake2010discovering} collected by Intel Labs, includes $p=1000$ nodes\footnote{
	For a clearer visualization, only $500$ nodes are processed, and isolated nodes are manually removed from the displayed graph.} and $n=218$ semantic features. As stated in \citep{cai2022mtp2}, ``each node denotes a concept such as `house', `coat', and `whale', and each semantic feature is a question such as `Can it fly?', `Is it alive?', and `Can you use it?' The answers are on a five-point scale from `definitely no' to `definitely yes', conducted on Amazon Mechanical Turk''. 
This dataset will only be used to assess the suitability of our factor model approaches, which are expected to be more efficient for such high-dimensional setting.
Since \textsf{GGFM} and \textsf{EGFM} lead to very similar results, only the former is shown.
In comparison, \textsf{GGM} and \textsf{EGM} did not provide as clean interpretable results, which motivates the use of factor models in practice. 
Because the ``true'' number of components is unclear in this data, methods that require prior setting of this variable were also found hard to exploit.
The graph learned by the \textsf{GGFM} algorithm is presented in \autoref{fig:concept_overview}.
This graph is mostly composed of connected sub-components, each falling into interpretable categories of similar concepts (``tools'' in grey, ``body parts'' in blue, etc.). 
A closer look on the graph in \autoref{fig:concept_zoom} shows 
interesting nodes that can be interpreted as links between the clusters of concepts (\textit{e.g.}, the node `body' linking the cluster of ``body parts'' to the one of ``whole bodies''). 
Another interpretation is that concepts with ambivalent meaning also act as nodes between these clusters (`gas' can be associated to ``drinks'' as well as to ``natural elements'').

\section{Conclusion}

In summary, we have proposed a family of graph learning algorithms in a unified formulation involving elliptical distributions, low-rank factor models, and Riemannian optimization. 
Our experiments demonstrated that the proposed approaches can evidence interpretable correlation structures in exploratory data analysis for various applications.

\subsubsection{Acknowledgements} This work was supported by the MASSILIA project
(ANR-21-CE23-0038-01) of the French National Research Agency (ANR). We thank V. Pinel for providing us the GNSS data set and D. Smittarello for insightful comments on the GNSS results. We also thank J. Ying for sharing the \textit{concepts} data set.\\

\newpage\phantom{blabla}
\newpage\phantom{blabla}

\begin{center}
\Large \textbf{Supplementary materials to: ``Learning Graphical Factor Models with Riemannian Optimization''}
\end{center}
\author{}
\authorrunning{A. Hippert-Ferrer et al.}
\institute{}
\tocauthor{Alexandre~Hippert-Ferrer, Florent~Bouchard, Ammar~Mian, Titouan~Vayer, Arnaud~Breloy}
\toctitle{Learning Graphical Factor Models with Riemannian Optimization}
\titlerunning{Learning Graphical Factor Models with Riemannian Optimization}

\appendix
\section{Outline}
\label{sec:outline}

\underline{Section \ref{sec:ethical_state}} explains potential ethical issues regarding the collection of the ground truth of the \textit{concepts} data set and underlines the limits of our graph predictions. \\

\noindent
\underline{Section \ref{sec:exp}} presents validation experiments on synthetic data:
\begin{itemize}
	\item[$\bullet$] Subsection \ref{subsec:setting} describes the experimental setup, as well as the graph and data generation parameters.
	\item[$\bullet$] Subsection \ref{subsec:result} displays performance comparisons between the proposed algorithms and other state of the art graph learning methods.
	\item[$\bullet$] Subsection \ref{subsec:robustness} presents a study of robustness regarding the selection of the parameters of the proposed methods (namely, the rank $k$ in factor models, and the regularization parameter $\lambda$ for the sparsity promoting penalty).
	
\end{itemize}
The code for reproducing these experiments is made available in the following repository:
\begin{center}
	\url{https://github.com/ahippert/graphfactormodel}
\end{center}

\newpage 
\section{Ethical statement}
\label{sec:ethical_state}

\subsection{Data set collection}

A part of our numerical experiments are conducted on data sets representing related ``entities'' of the real world (\textit{animals} and \textit{concepts} data sets). We (scientists of the data/machine learning/signal processing communities) are conscious that both data sets are limited, non-exhaustive and partial. For example, the \textit{concepts} data set has been labelled by humans using a survey conducted on the Amazon Mechanical Turk (AMT) platform. As recently documented in \cite{crawford2021}, these data sets might incorporate sociological and cultural biases, embedding a particular view of the world they intend to fit. Furthermore, AMT, amongst other micro-working platforms, employs precarious \textit{crowdworkers} around the world at low pay rates \citep{gray2019} for annotating, labelling and correcting data that help to train and test AI models. This, somehow, maintains the illusion that AI systems are autonomous and intelligent by hiding real workers \citep{aytes2012} and the labor-intensive process they require \citep{tubaro2019}.

\subsection{Limits of the graph prediction}
Our proposed algorithms have been designed to learn graph connections between related entities. As discussed above, the \textit{concepts} data set might incorporate various forms of biases because each semantic feature are questions which require a subjective answer. As no ground truth is available for the considered data sets, we advise users to carefully interpret the predicted graph structures. Indeed, spurious graph links might be learned, which would highlight the aforementioned biases. Any predicted graph is no immutable truth and should be questioned in a critical manner. We strongly encourage the development of strategies to mitigate spurious correlations, as reliance on expert advice (\textit{i.e.}, a volcanologist for the GNSS data set) or the design of more open, transparent and representative training data.

\section{Numerical experiments on synthetic graphs}
\label{sec:exp}

In the following, we present extensive validation experiments on synthetic graphs to evaluate the performance of the proposed algorithms, \textit{i.e.}, \textsf{GGM}, \textsf{EGM}, \textsf{GGFM} and \textsf{EGFM}.

\subsection{Experimental settings}

\label{subsec:setting}

\begin{figure}[!t]
	\centering
	\includegraphics[width=0.8\textwidth]{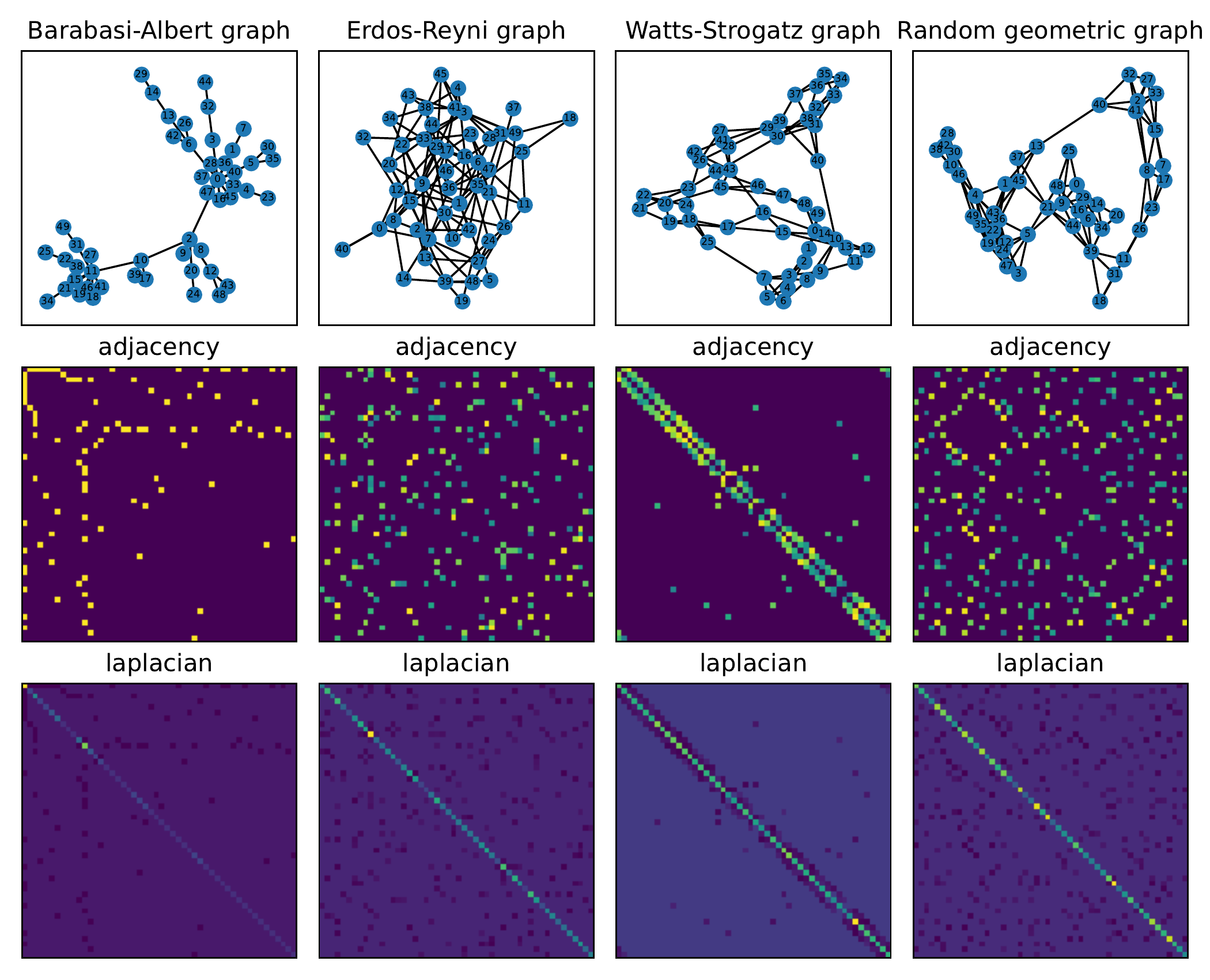}
	\caption{Considered graph structures (here with $p=50$ nodes) and associated adjacency and laplacian (precision) matrices.}
	\label{fig:graph_structures}
\end{figure}

\noindent
\textbf{Graph and sample generation}:
For a fixed dimension $p$, we consider four standard random graphs models that are described in \autoref{tab:graph_models}.
Given a sampled support of the graph, weights of the edges then are sampled from $U(2,5)$ (as in \citep{ying2020nonconvex}), which yields a symmetric weighted adjacency matrix $\MAT{A}$ for each random graph. \autoref{fig:graph_structures} shows examples of sampled graph structures and their adjacency matrices for each of the four models.
The Laplacian matrix of the graph is then used to construct the precision matrix $\MAT{\Theta}$, using the relation $\MAT{\Theta} =
\MAT{L} + \kappa \MAT{I}$, where $\MAT{L}$ is the Laplacian matrix $\MAT{L} = \MAT{D} - \MAT{A}$ (where $\MAT{D}$ is the degree matrix of $\MAT{A}$), and where $\kappa=1e^{-1}$
is used so that $\MAT{\Theta}$ is non-singular. 
For the data generation, a total of $n$ samples $\{\VEC{x}_i \in \mathbb{R}^p\}_{i=1}^n$ are drawn from an elliptical graphical model, parameterized by its covariance matrix $\MAT{\Sigma} = \MAT{\Theta}^{-1}$ and density generator $g$, \textit{i.e.}, $\VEC{x}_i\sim\mathcal{ES}(\MAT{0},\MAT{\Theta}^{-1},g)$. 
For the density generator $g$, we use a Student's $t$-distribution with degrees of freedom $\nu$.\\

\begin{table}[!t]
	\centering
	\begin{tabular}{c|c|c|c|c}
		\textbf{Graph model} & Edge weights & Probability & Neighbors & Radius \\
		\hline
		Barabási-Albert & $U(2,5)$ & -- & -- & -- \\
		Erdős–Rényi & $U(2,5)$ & 0.1 & -- & -- \\
		Watts-Strogatz & $U(2,5)$ & 0.1 & 5 & -- \\
		Random geometric & $U(2,5)$ & -- & -- & 0.2
	\end{tabular}
	\caption{Considered graph models in numerical experiments.}
	\label{tab:graph_models}
\end{table}

\noindent
\textbf{Performance evaluation}:
For a given parameter setup, we sample the graph-data pair $(\MAT{A}, \MAT{X})$ as described above.
Graph learning algorithms are then applied to the input data $\MAT{X}$.
These algorithms can provide different types of estimate structures with various inherent normalization, which are not always comparable.
For a meaningful comparison, we consider evaluating the receiver operating characteristic (ROC) curves obtained from the estimated adjacency matrix $\MAT{A}$. 
The ROC curves displays the true positive rate (tpr) as function of the false positive rate (fpr): in our case, tpr denotes the capacity of the algorithm to recover actual edges of the algorithm, whereas fpr accounts for the false discovery of non-existing edges. For each curve, the area under curve (AUC) is computed. The AUC takes values in $[0,1]$, with $1$ indicating a perfect recovery of the true edges.
These ROC curves are obtained as follows:
First, we compute the conditional correlation coefficients
\begin{equation}\label{eq:condcorr}
	\tilde{\MAT{\Theta}}_{ij} = 
	{\rm corr}\left[ x_q x_\ell | \VEC{x}_{[\![ 1,p ]\!]\backslash \{q,\ell\}} \right]= - \MAT{\Theta}_{q\ell} / \sqrt{\MAT{\Theta}_{qq}\MAT{\Theta}_{\ell\ell}}.
\end{equation}
obtained from any output estimate of the precision matrix $\MAT{\Theta}$ (respectively, $\MAT{\Sigma}^{+}$ if the algorithm provides an estimate of the covariance matrix).
Given these coefficients estimates, a ROC curve is obtained by varying a threshold ${\rm tol}$, i.e., the edge $(i,j)$ is considered active if $\tilde{\MAT{\Theta}}_{ij}  > {\rm tol} $.
The displayed ROC curves are finally obtained from the average of $50$ Monte-Carlo experiments.\\

\noindent
\textbf{Compared methods}:
The proposed algorithms (\textsf{GGM}/\textsf{GGFM}/\textsf{EGM}/\textsf{EGFM}) are compared with state-of-the-art approaches 
to learn unstructured graphs: 
\textsf{GLasso} \citep{friedman2008sparse}, which uses a Gaussian model and $\ell_1$-norm as a sparse-promoting penalty;
\textsf{NGL} and \textsf{SGL} \citep{ying2020nonconvex, kumar2020unified}, which use a Gaussian Laplacian-constrained model with a concave penalty regularization (without prior knowledge on the number of graph components \textsf{NGL} and \textsf{SGL} solve a similar problem with a slightly different implementation);  
\textsf{StGL} \citep{de2021graphical}, which generalizes the above to $t$-distributed data.
For a fair comparison, the regularization parameters of all tested algorithms (including competing methods) are tuned to the best of our ability in order to display the best results. 
Notably, we notice that all these methods are not critically sensitive to a slight change of their regularization parameter $\lambda$  (given a reasonable order of magnitude).
Furthermore, a robustness analysis concerning this point (performance versus change in parameters) is also performed subsequently for the proposed methods.

\subsection{Results: ROC curves comparisons in different setups}
\label{subsec:result}

\begin{figure}
	\centering
	\begin{subfigure}[b]{0.45\textwidth}
		\centering
		\includegraphics[width=\textwidth]{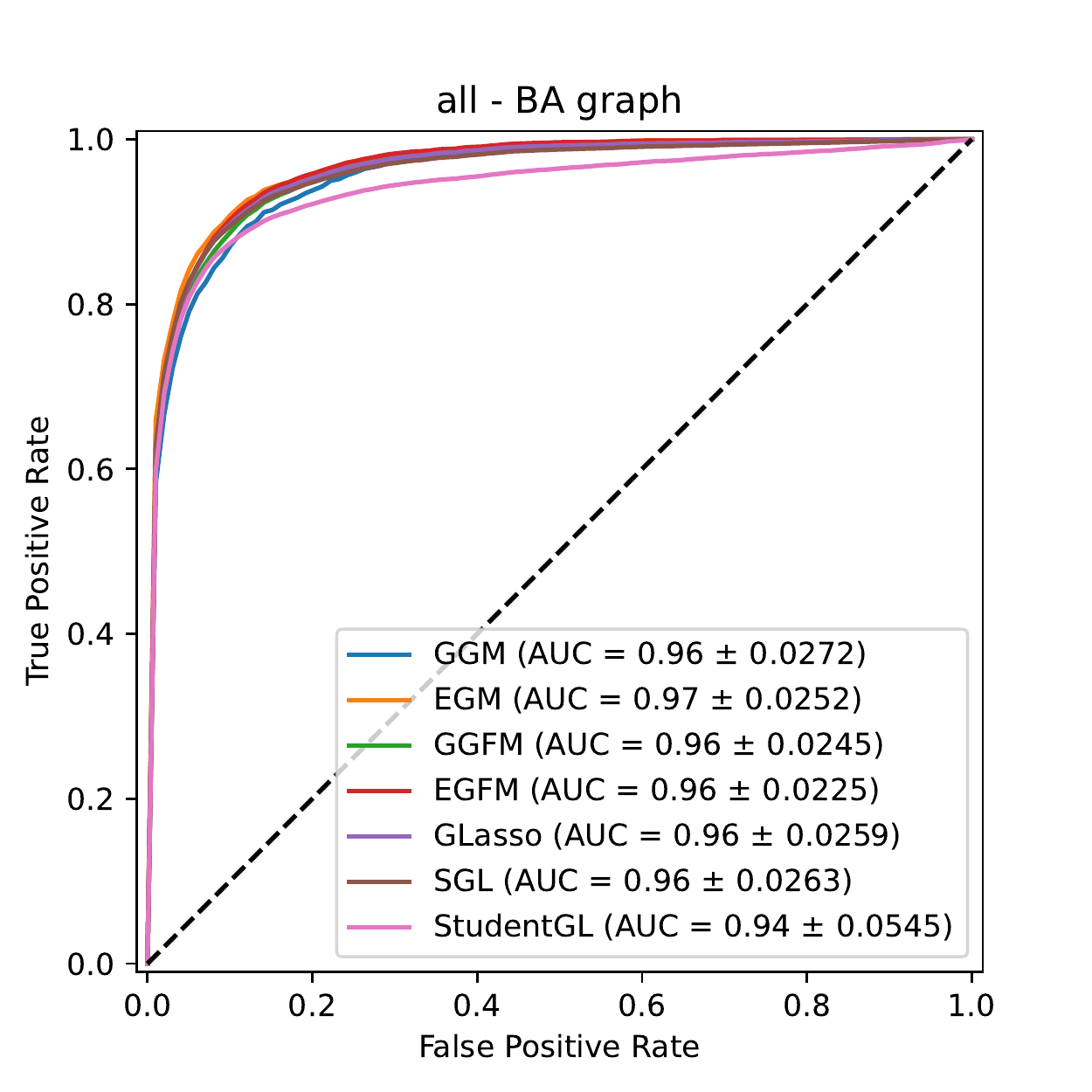}
		\caption{$n=2p$}
	\end{subfigure}
	\hfill
	\centering
	\begin{subfigure}[b]{0.45\textwidth}
		\centering
		\includegraphics[width=\textwidth]{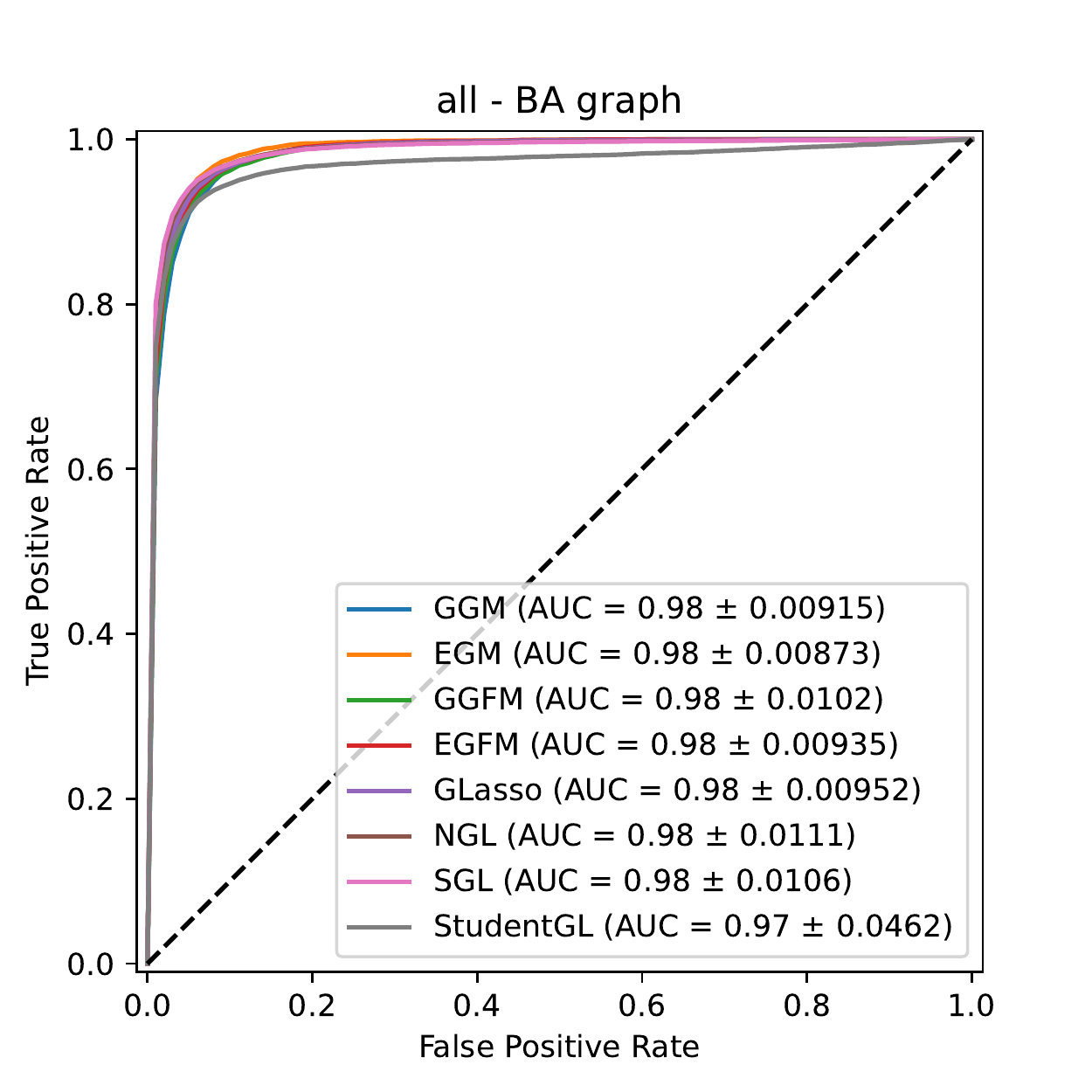}
		\caption{$n=5p$}
	\end{subfigure}
	\caption{Mean ROC curves obtained from estimated adjacency matrices $\hat{\mathbf{A}}$ of a \textbf{Barabási-Albert} model with \textsf{GGM/EGM} ($\lambda=0.05$), \textsf{GGFM/EGFM} ($\lambda=0.01; k=20$), \textsf{GLasso} ($\alpha=0.1$), \textsf{NGL} ($\lambda=0.1$), \textsf{SGL} ($\alpha=0.1$) and \textsf{StudentGL} (1 component) algorithms. Note that \textsf{NGL} is not displayed for $n=2p$ because of numerical divergence.}
	\label{fig:roc_barabasi}
\end{figure}

\begin{figure}
	\centering
	\begin{subfigure}[b]{0.45\textwidth}
		\centering
		\includegraphics[width=\textwidth]{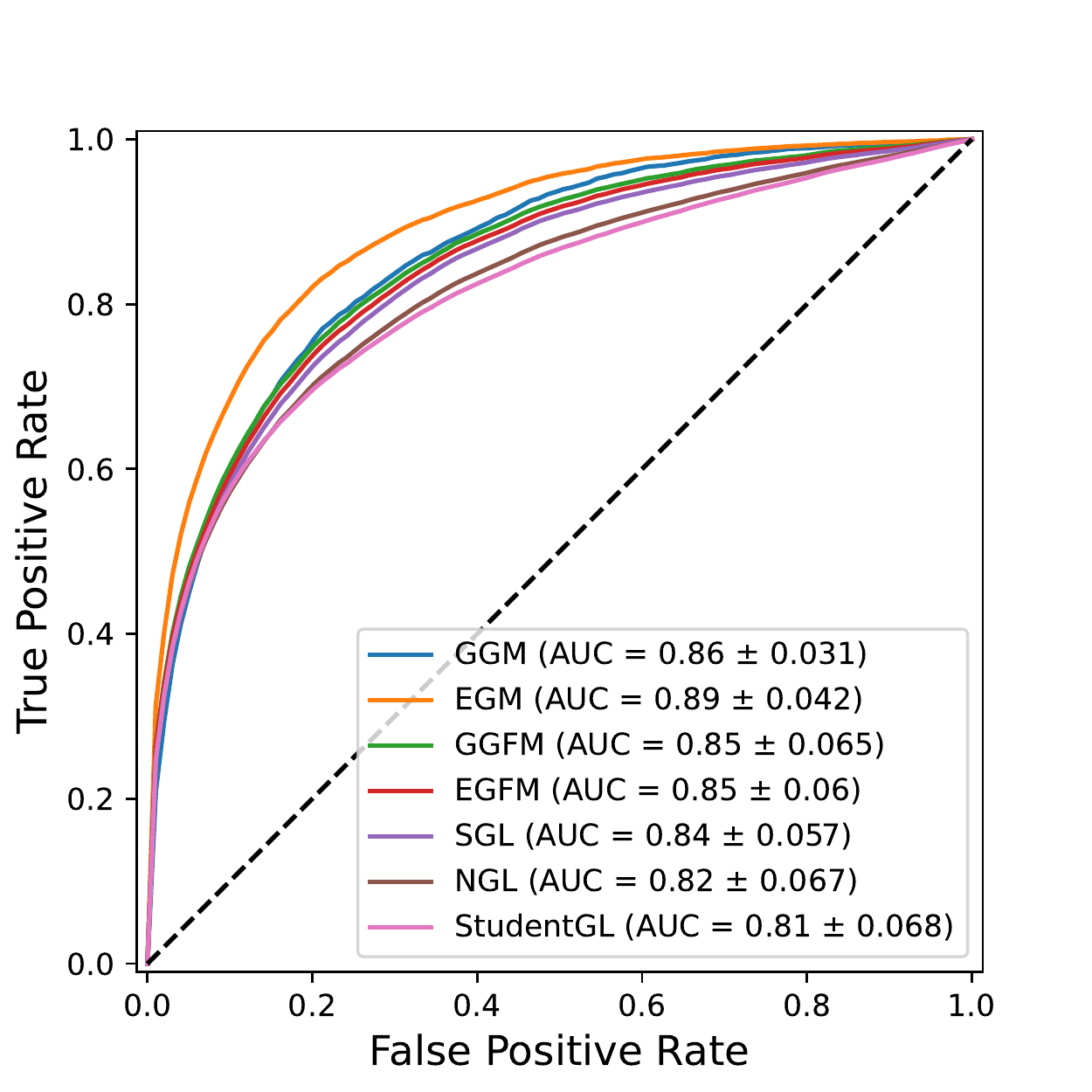}
		\caption{$n=2p$}
	\end{subfigure}
	\hfill
	\centering
	\begin{subfigure}[b]{0.45\textwidth}
		\centering
		\includegraphics[width=\textwidth]{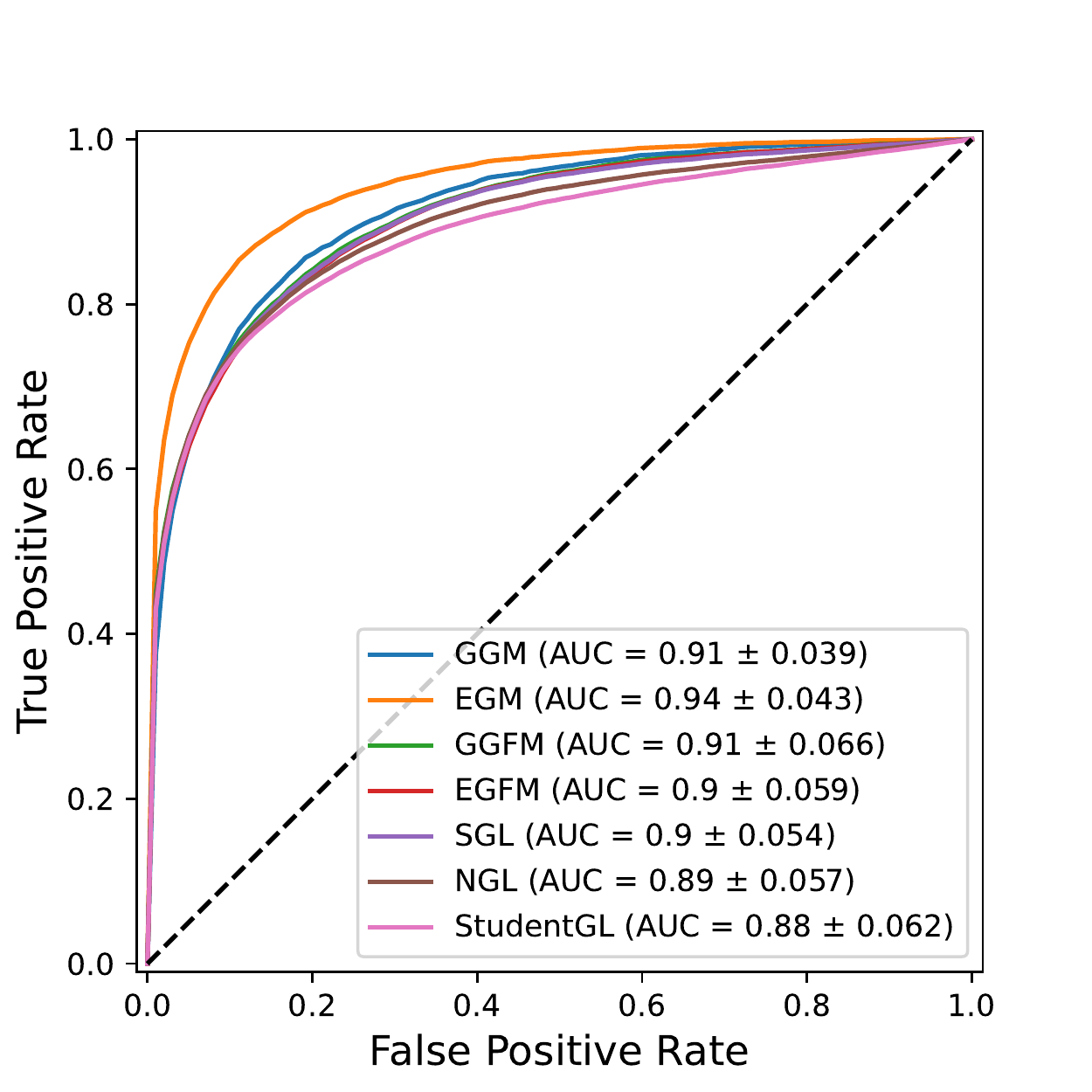}
		\caption{$n=5p$}
	\end{subfigure}
	\caption{Mean ROC curves obtained from estimated adjacency matrices $\hat{\mathbf{A}}$ of a \textbf{Erdős–Rényi} model with \textsf{GGM/EGM} ($\lambda=0.05$), \textsf{GGFM/EGFM} ($\lambda=0.01; k=20$), \textsf{GLasso} ($\alpha=0.1$), \textsf{NGL} ($\lambda=0.1$), \textsf{SGL} ($\alpha=0.1$) and \textsf{StudentGL} (1 component) algorithms. Note that \textsf{GLasso} is not displayed because of numerical divergence.}
	\label{fig:roc_erdos}
\end{figure}

\begin{figure}
	\centering
	\begin{subfigure}[b]{0.45\textwidth}
		\centering
		\includegraphics[width=\textwidth]{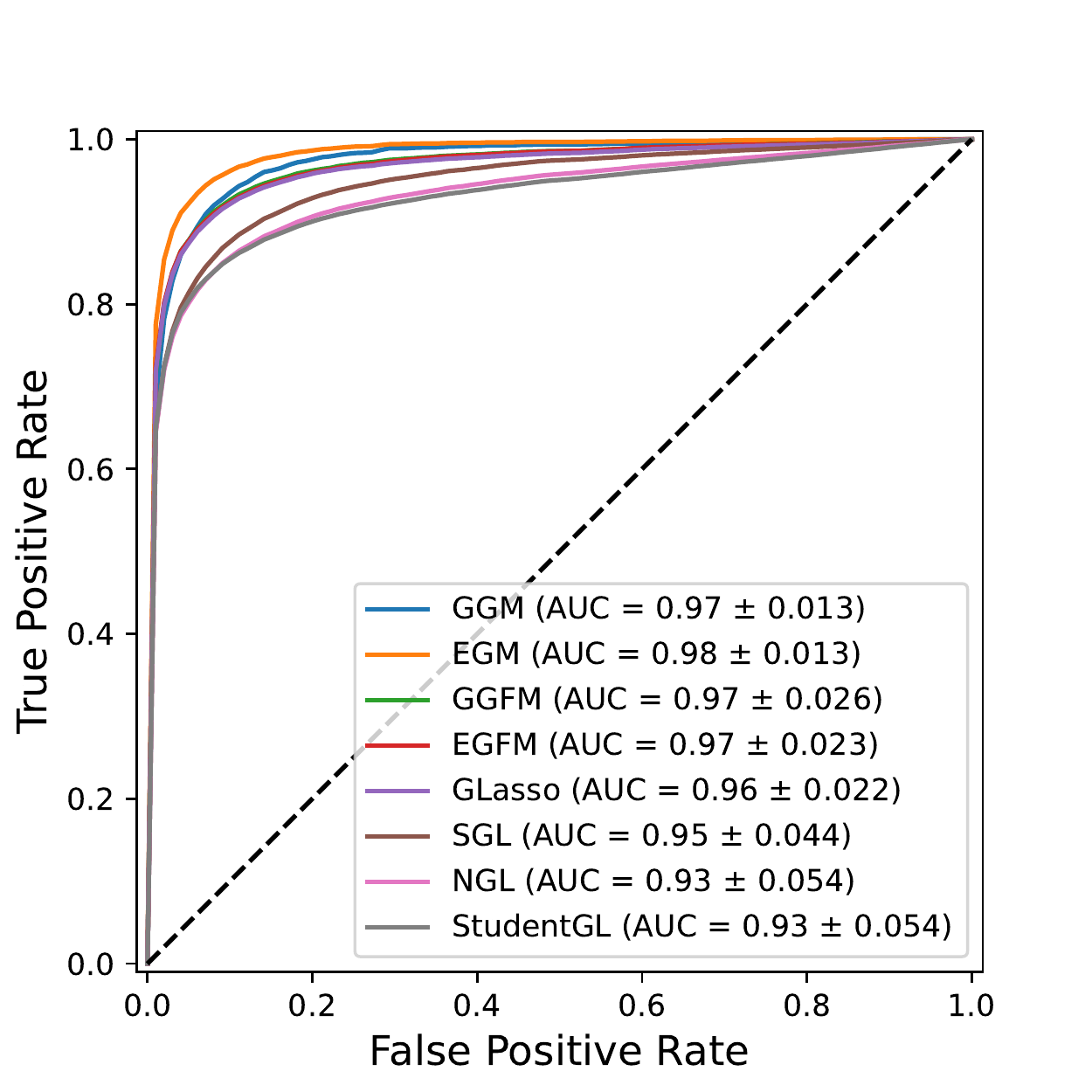}
		\caption{$n=2p$}
	\end{subfigure}
	\hfill
	\centering
	\begin{subfigure}[b]{0.45\textwidth}
		\centering
		\includegraphics[width=\textwidth]{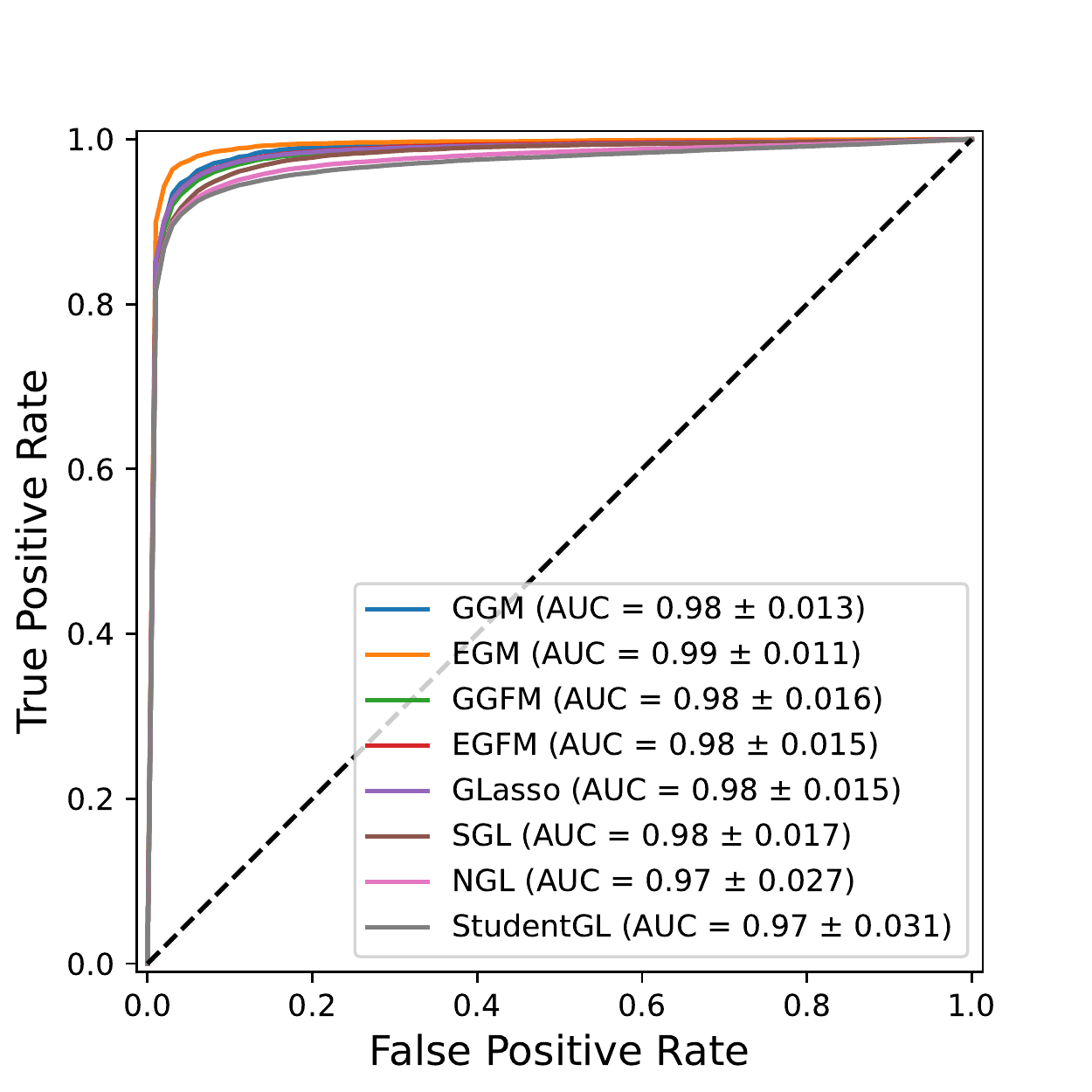}
		\caption{$n=5p$}
	\end{subfigure}
	\caption{Mean ROC curves obtained from estimated adjacency matrices $\hat{\mathbf{A}}$ of a \textbf{Watts Strogatz} model with \textsf{GGM/EGM} ($\lambda=0.1$), \textsf{GGFM/EGFM} ($\lambda=0.01; k=10$), \textsf{GLasso} ($\alpha=0.05$), \textsf{NGL} ($\lambda=0.1$), \textsf{SGL} ($\alpha=0.1$) and \textsf{StudentGL} (1 component) algorithms.}
	\label{fig:roc_watts}
\end{figure}

\begin{figure}
	\centering
	\begin{subfigure}[b]{0.45\textwidth}
		\centering
		\includegraphics[width=\textwidth]{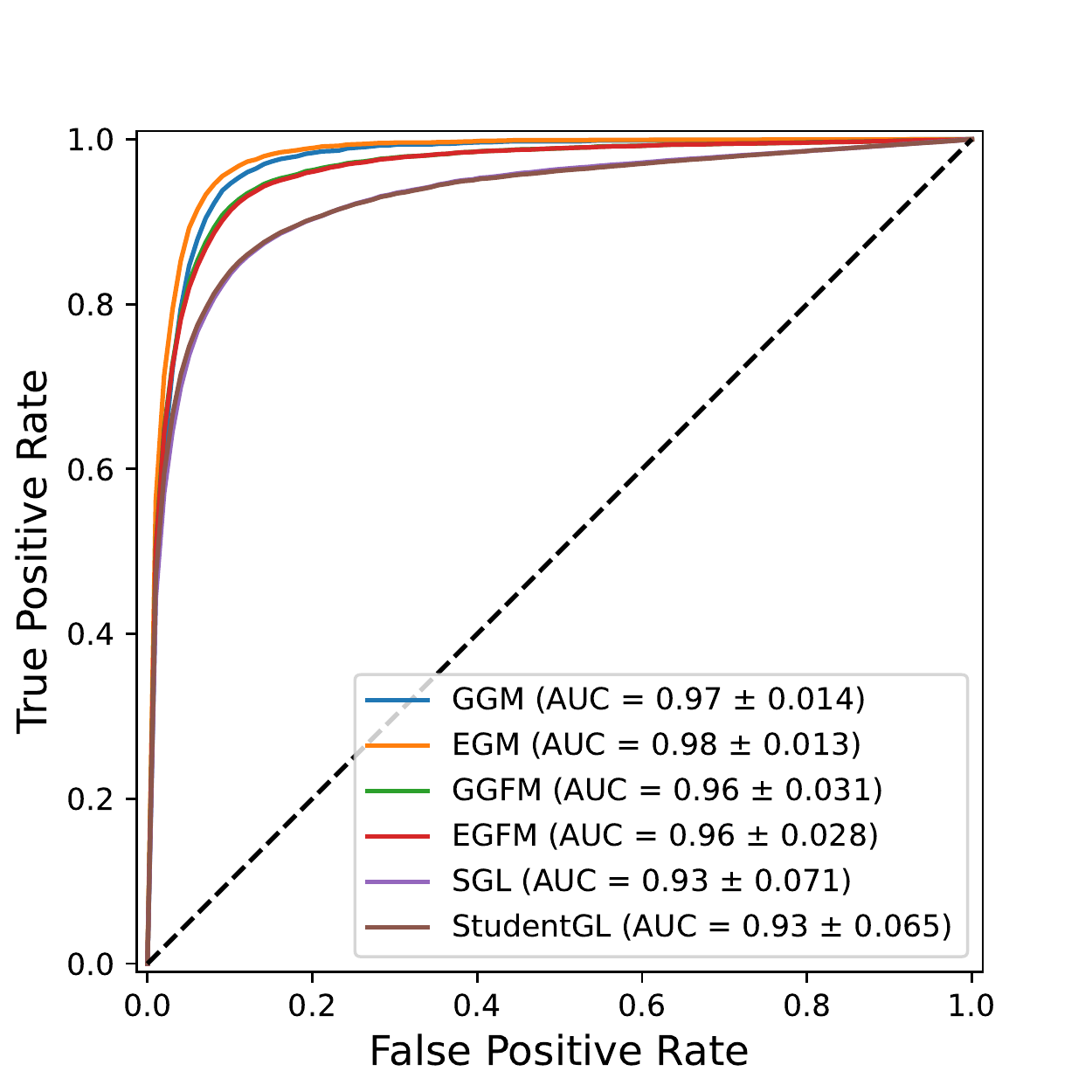}
		\caption{$n=2p$}
	\end{subfigure}
	\hfill
	\centering
	\begin{subfigure}[b]{0.45\textwidth}
		\centering
		\includegraphics[width=\textwidth]{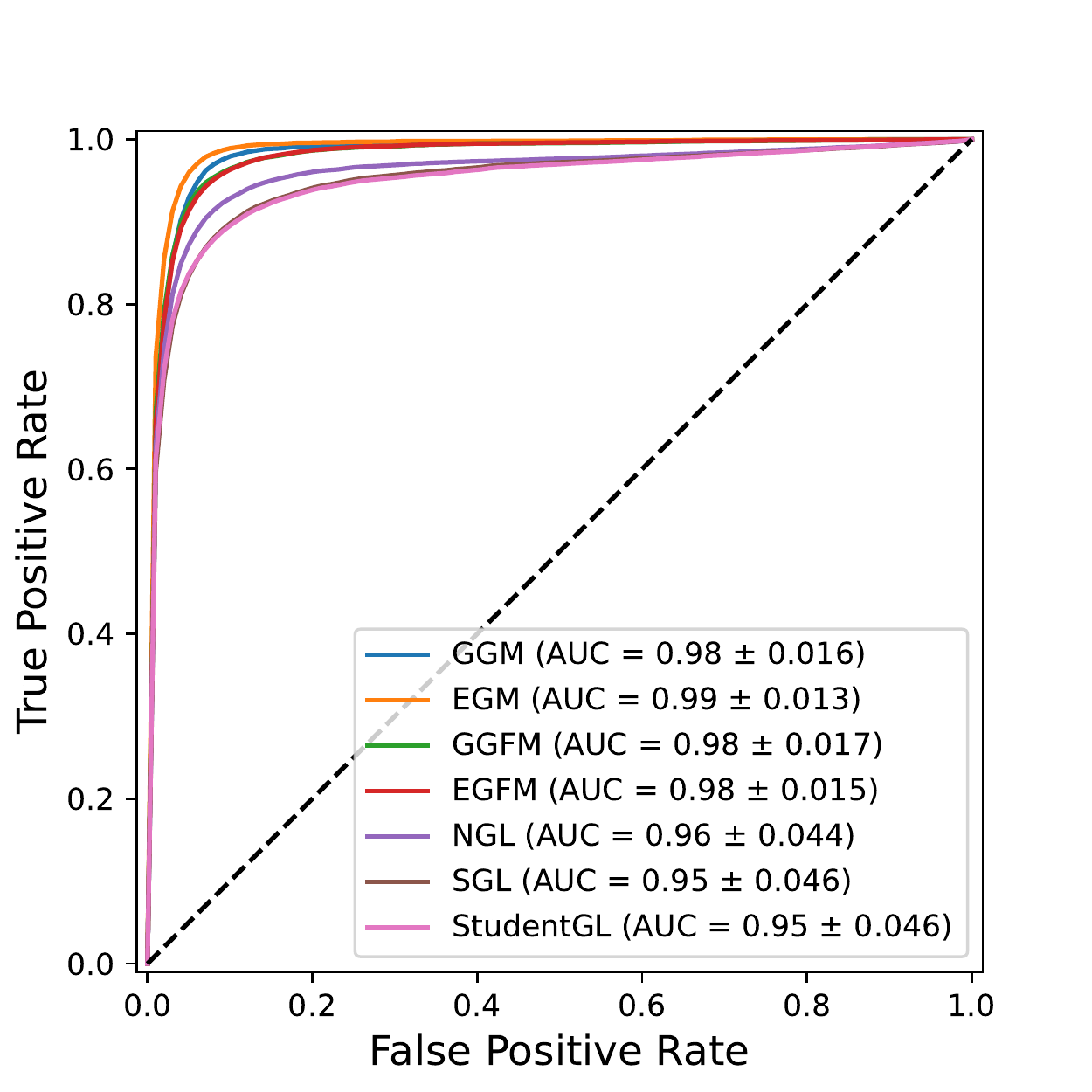}
		\caption{$n=5p$}
	\end{subfigure}
	\caption{Mean ROC curves obtained from estimated adjacency matrices $\hat{\mathbf{A}}$ of a \textbf{Random geometric} model with \textsf{GGM} ($\lambda=0.1$), \textsf{EGM} ($\lambda=0.1$), \textsf{GGFM} ($\lambda=0.01; k=20$), \textsf{EGFM} ($\lambda=0.05; k=20$), \textsf{NGL} ($\lambda=0.02$), \textsf{SGL} ($\alpha=0.1$) and \textsf{StudentGL} (1 component) algorithms. Note that \textsf{GLasso} is not displayed because of numerical divergence, neither is \textsf{NGL} for $n=2p$.}
	\label{fig:roc_geometric}
\end{figure}

\autoref{fig:roc_barabasi}, \ref{fig:roc_erdos}, \ref{fig:roc_watts}, \ref{fig:roc_geometric} show the results with $p=50$ and $\nu=3.5$ (non-Gaussian distributed data) for Barabási-Albert, Erdos-Rényi, Watts-Strogatz and random geometric models, respectively. Two sample support settings are considered, \textit{i.e.}, $n=2p$ and $n=5p$. Overall, the proposed approaches outperform all compared methods (\textit{i.e.}, larger AUC), except for Barabási-Albert where no significant differences are observed. Notable gains are obtained when $n$ is low (i.e., when less samples are available). 
When $n$ increases, the gap in performances between proposed and compared methods gets thinner.
We also notice that \textsf{EGM} outperforms \textsf{GGM}, which was to be expected since the underlying data distribution is not Gaussian.
As the generated graphs do not necessarily yield a low-rank structured covariance matrix, \textsf{EGFM} and \textsf{GGFM} do not outperform their full-rank counterparts in the considered setting.
Still, these algorithms do not perform significantly worse,
which means that one can reasonably reduce the dimension of the model (and benefit from a lower computational complexity).

\subsection{\textsf{GGM}/\textsf{GGFM}/\textsf{EGM}/\textsf{EGFM}: robustness to parameters $\lambda$ and $k$}
\label{subsec:robustness}

\begin{figure}[htb]
	\subfloat{\includegraphics[trim = 1.5mm 3mm 11mm 14mm, clip,width=0.27\textwidth]{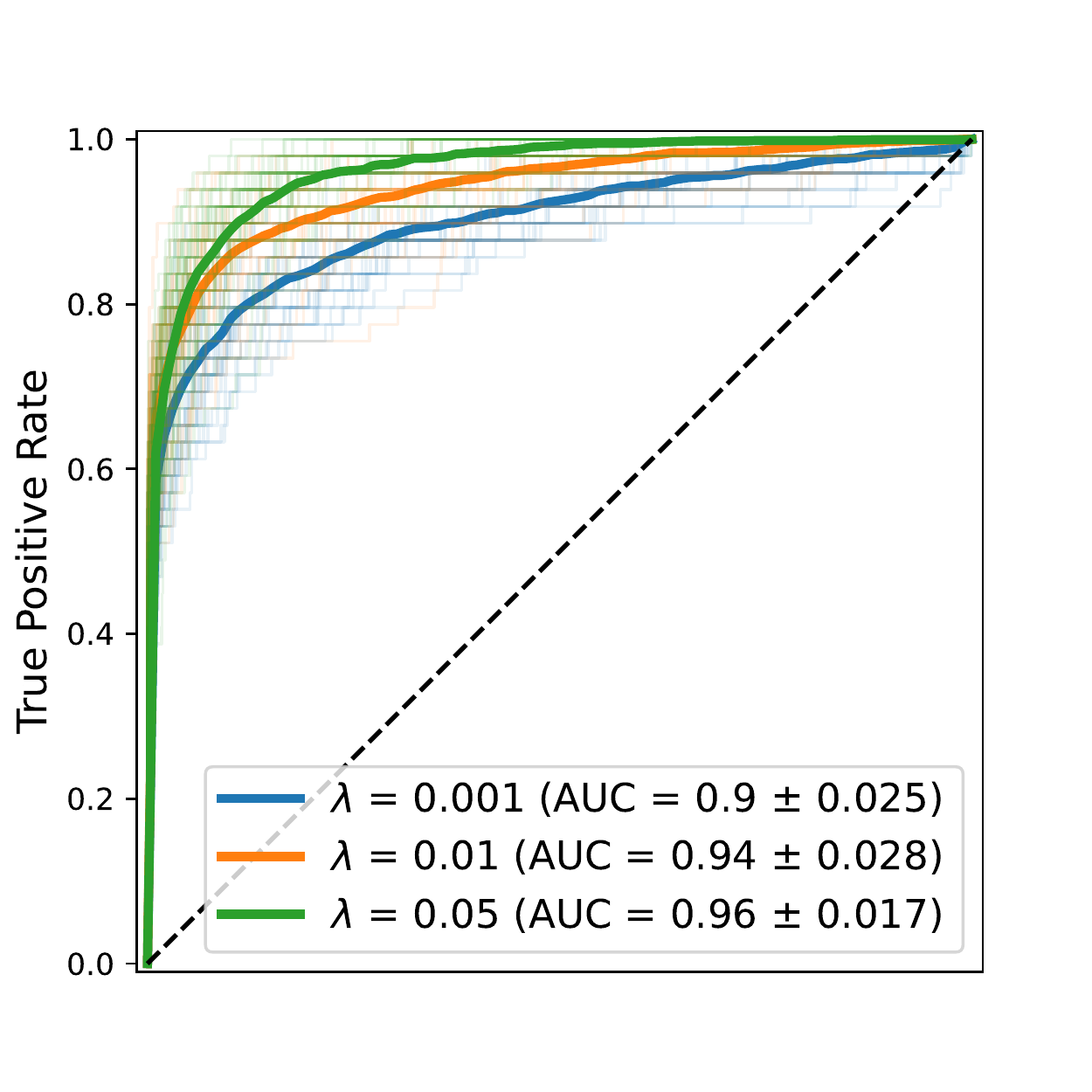}}\hfill
	\subfloat{\includegraphics[trim = 14mm 3mm 11mm 14mm, clip,width=0.24\textwidth]{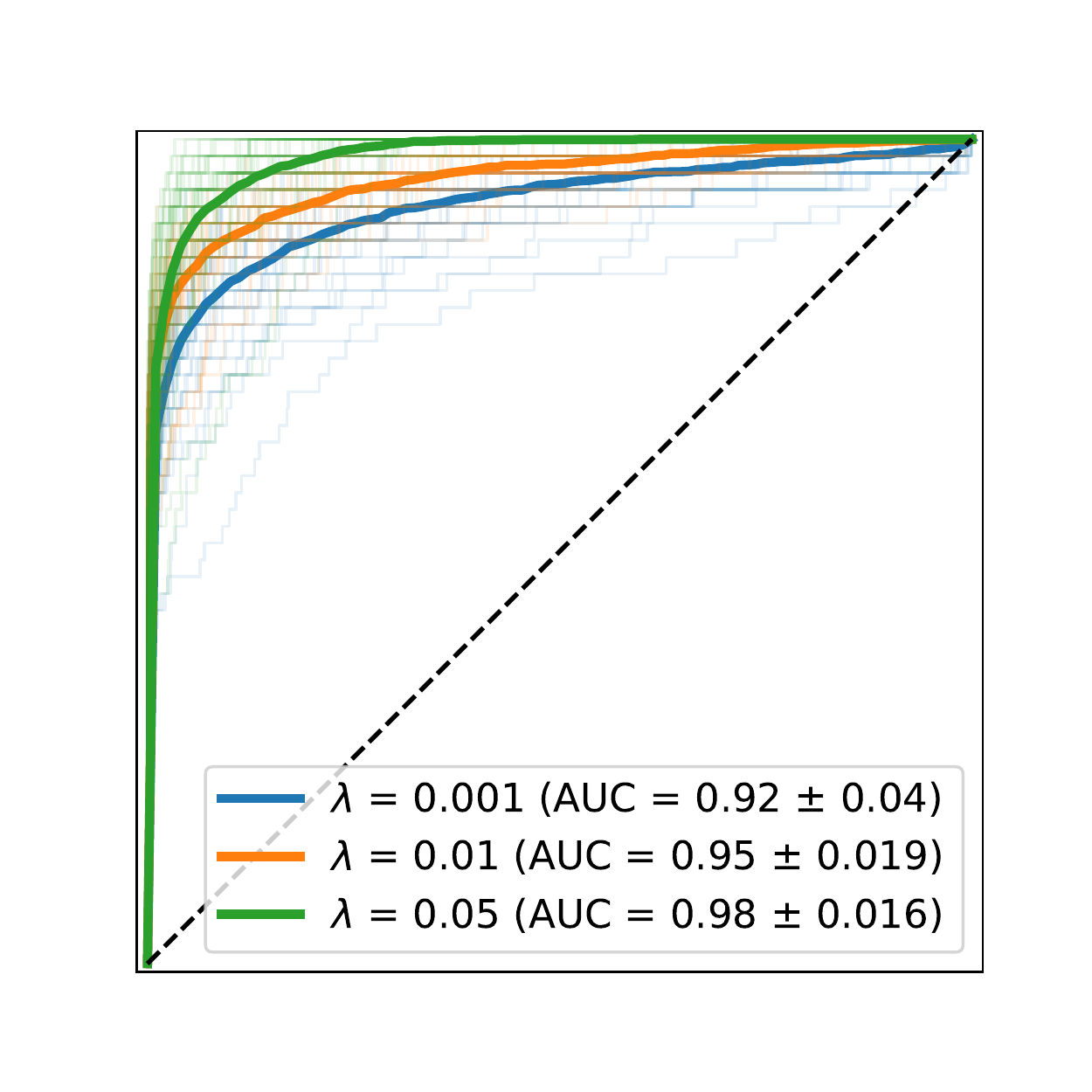}}\hfill  
	\subfloat{\includegraphics[trim = 14mm 3mm 11mm 14mm, clip,width=0.24\textwidth]{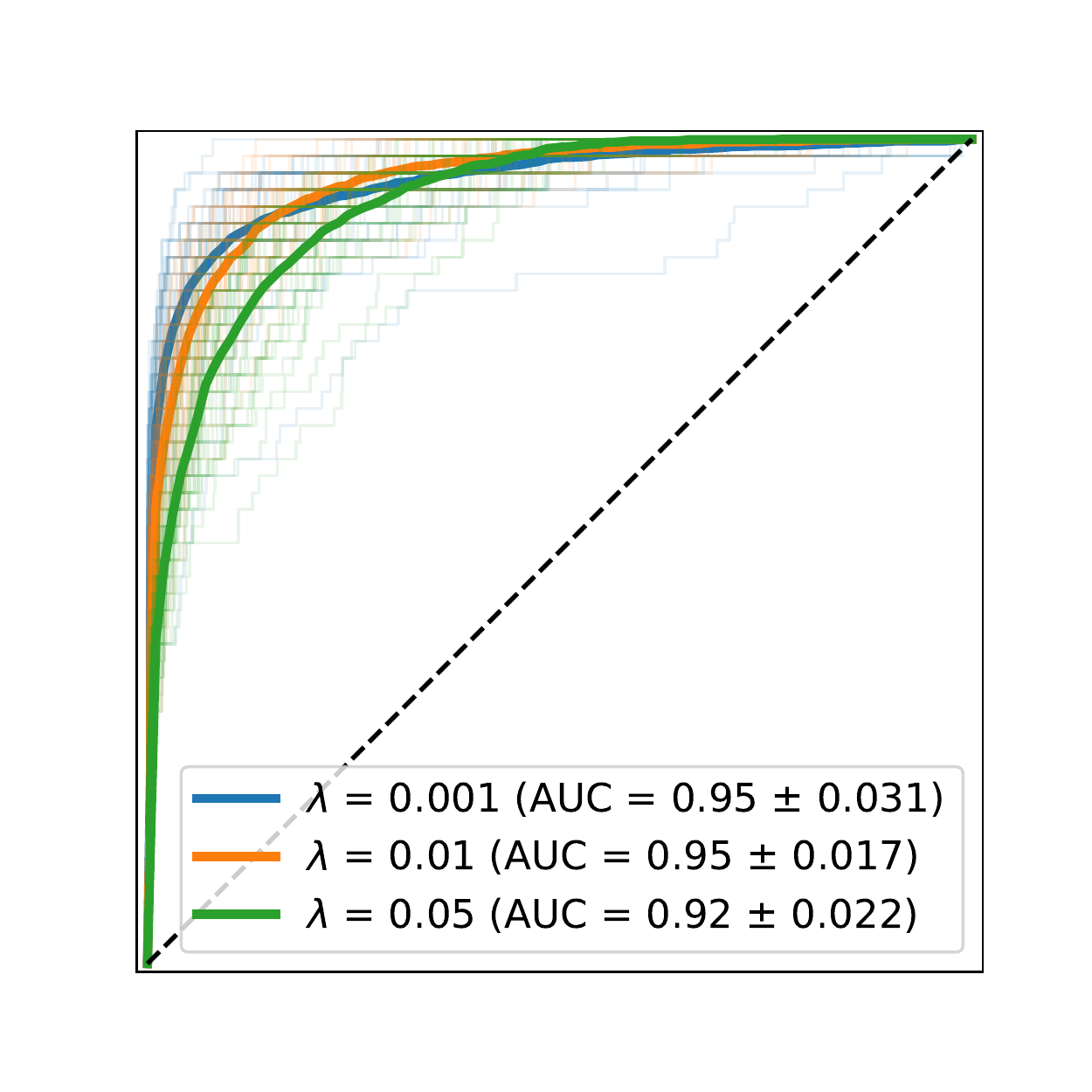}}\hfill
	\subfloat{\includegraphics[trim = 14mm 3mm 11mm 14mm, clip,width=0.24\textwidth]{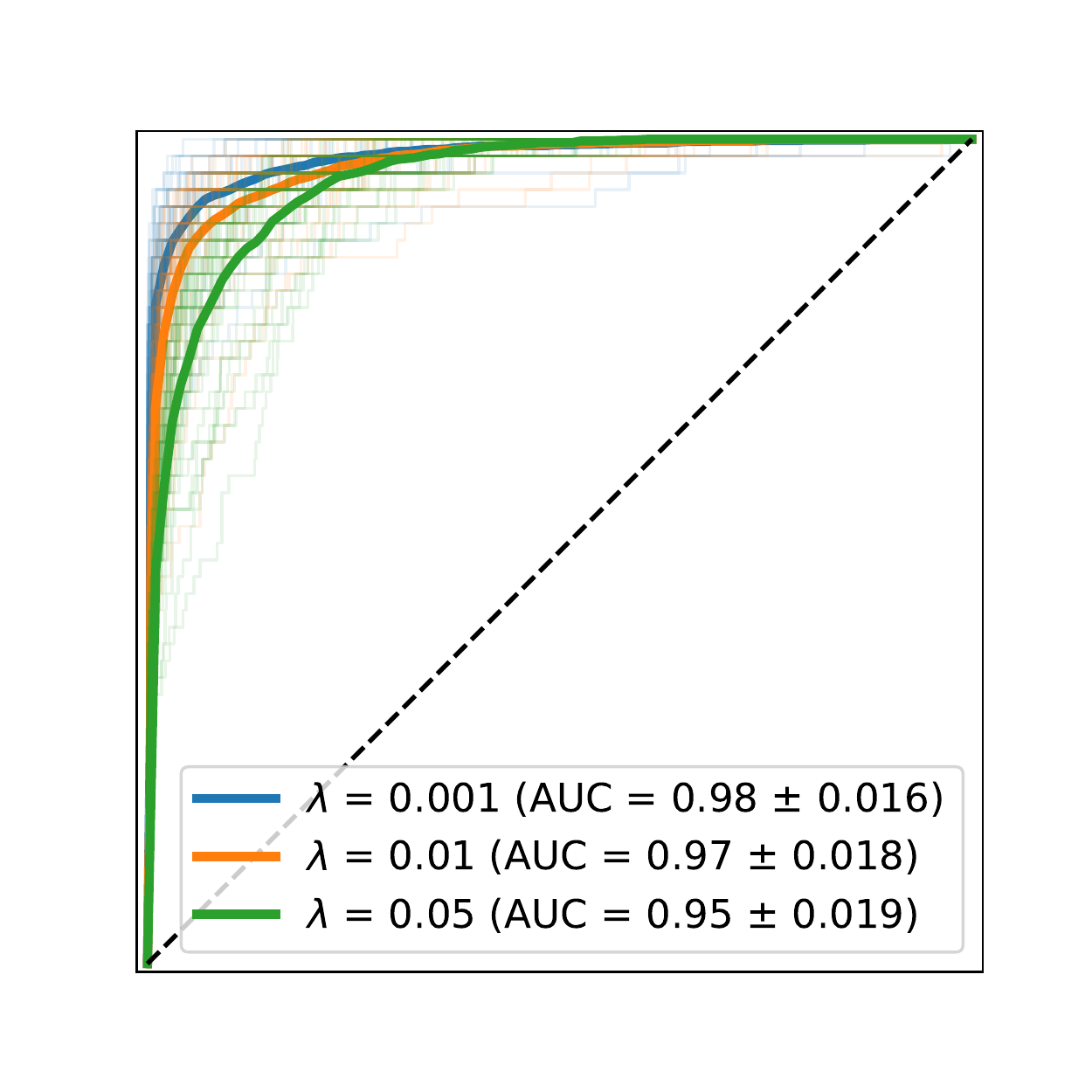}}\\[-2ex]
	
	\subfloat{\includegraphics[trim = 1.5mm 3mm 11mm 14mm, clip,width=0.27\textwidth]{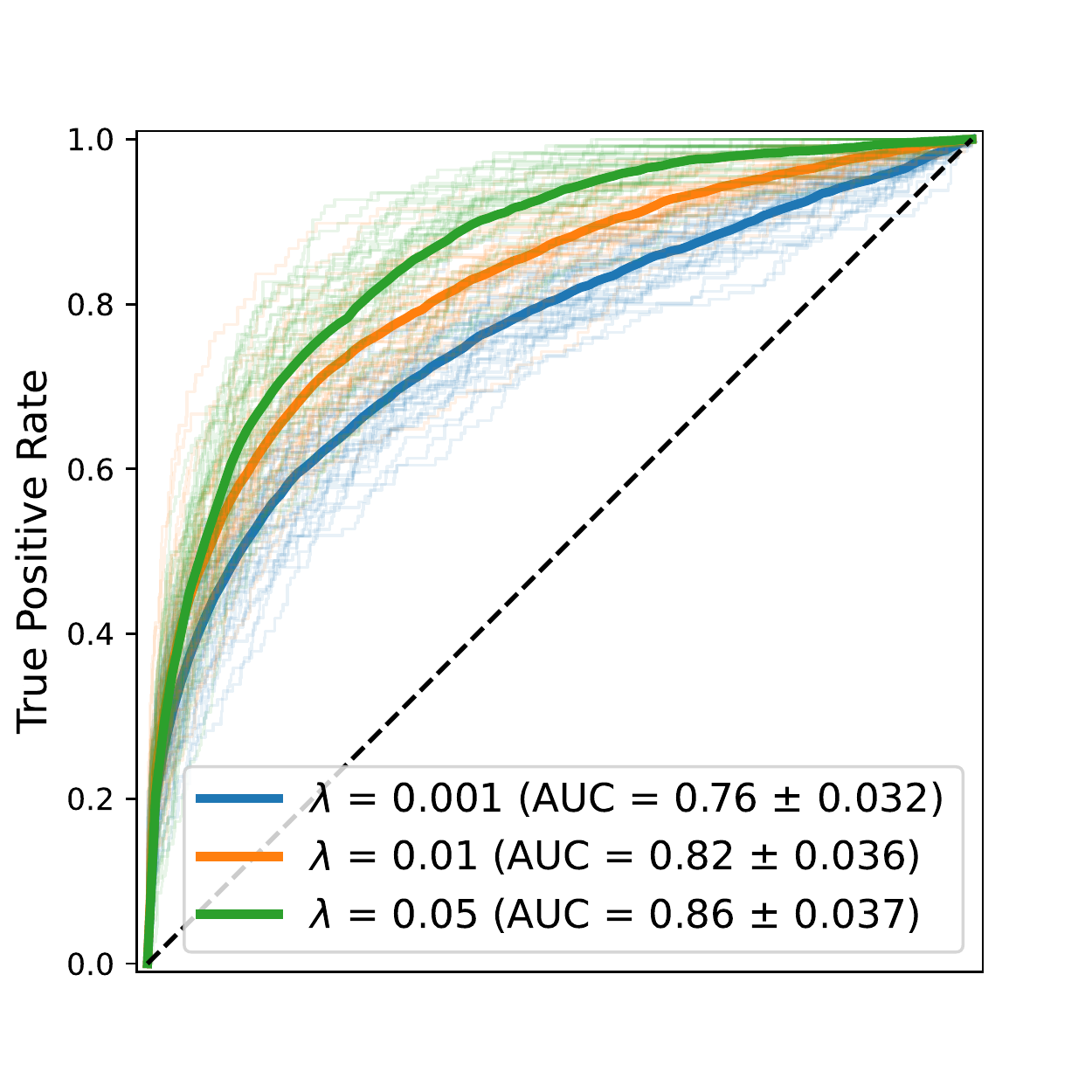}}\hfill
	\subfloat{\includegraphics[trim = 14mm 3mm 11mm 14mm, clip,width=0.24\textwidth]{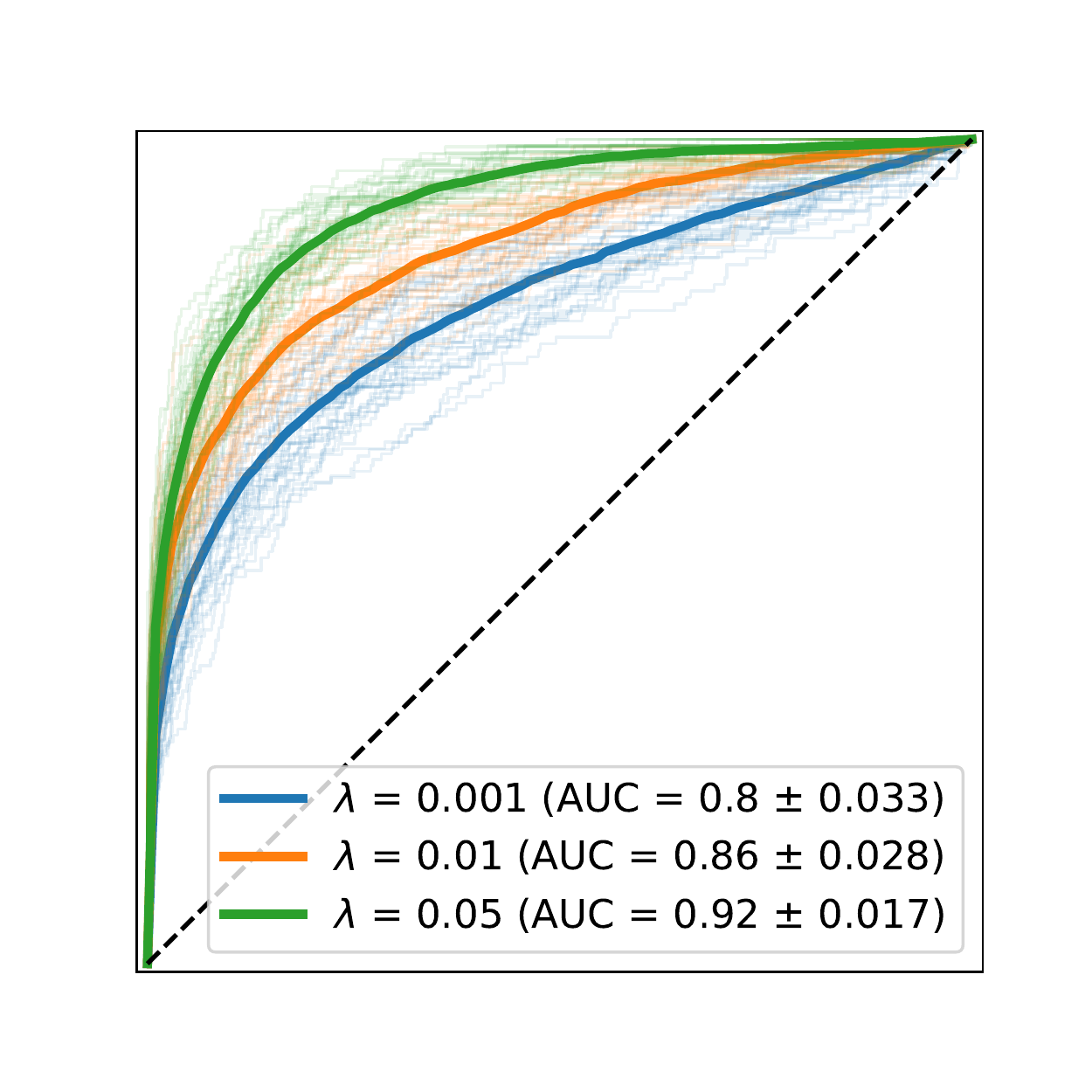}}\hfill  
	\subfloat{\includegraphics[trim = 14mm 3mm 11mm 14mm, clip,width=0.24\textwidth]{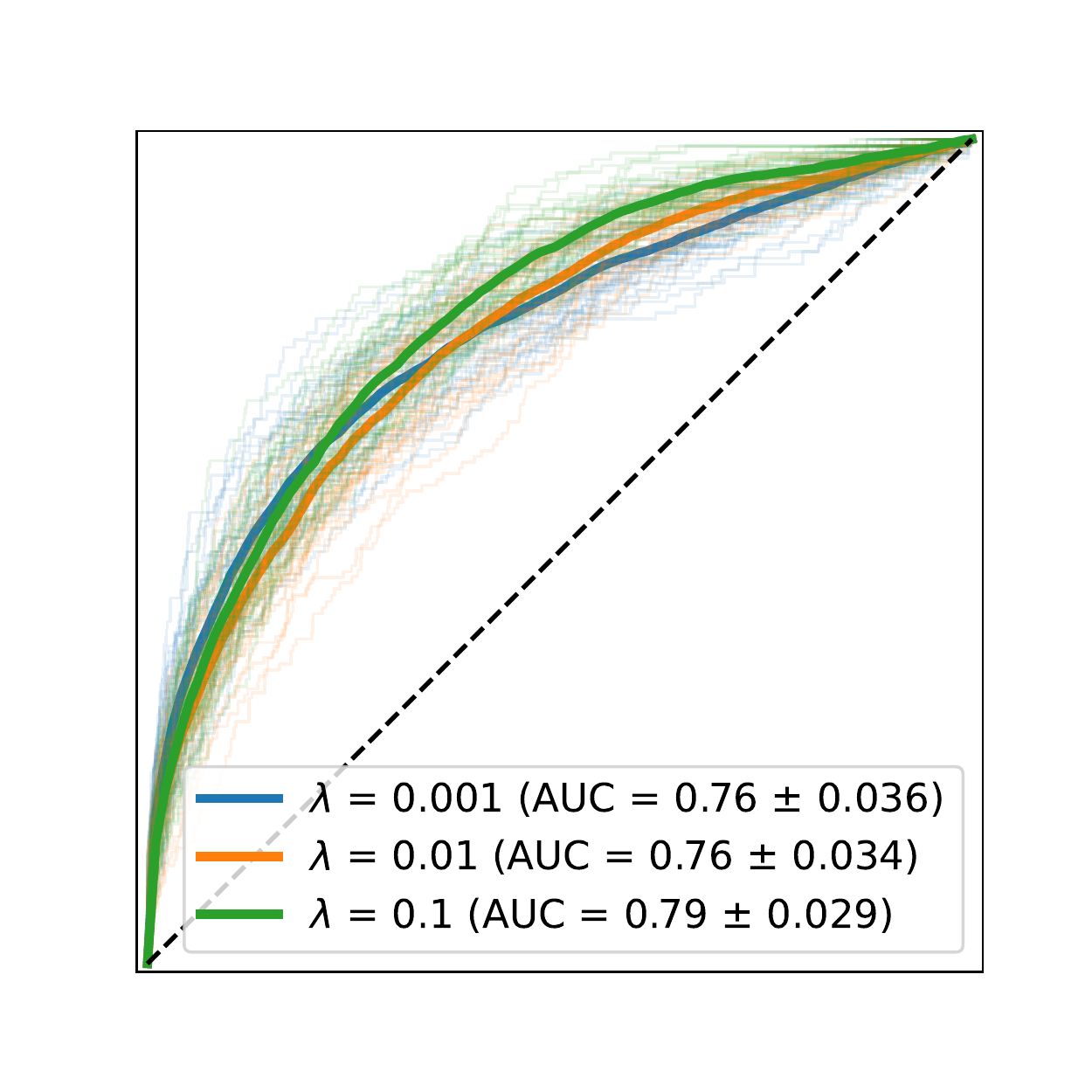}}\hfill
	\subfloat{\includegraphics[trim = 14mm 3mm 11mm 14mm, clip,width=0.24\textwidth]{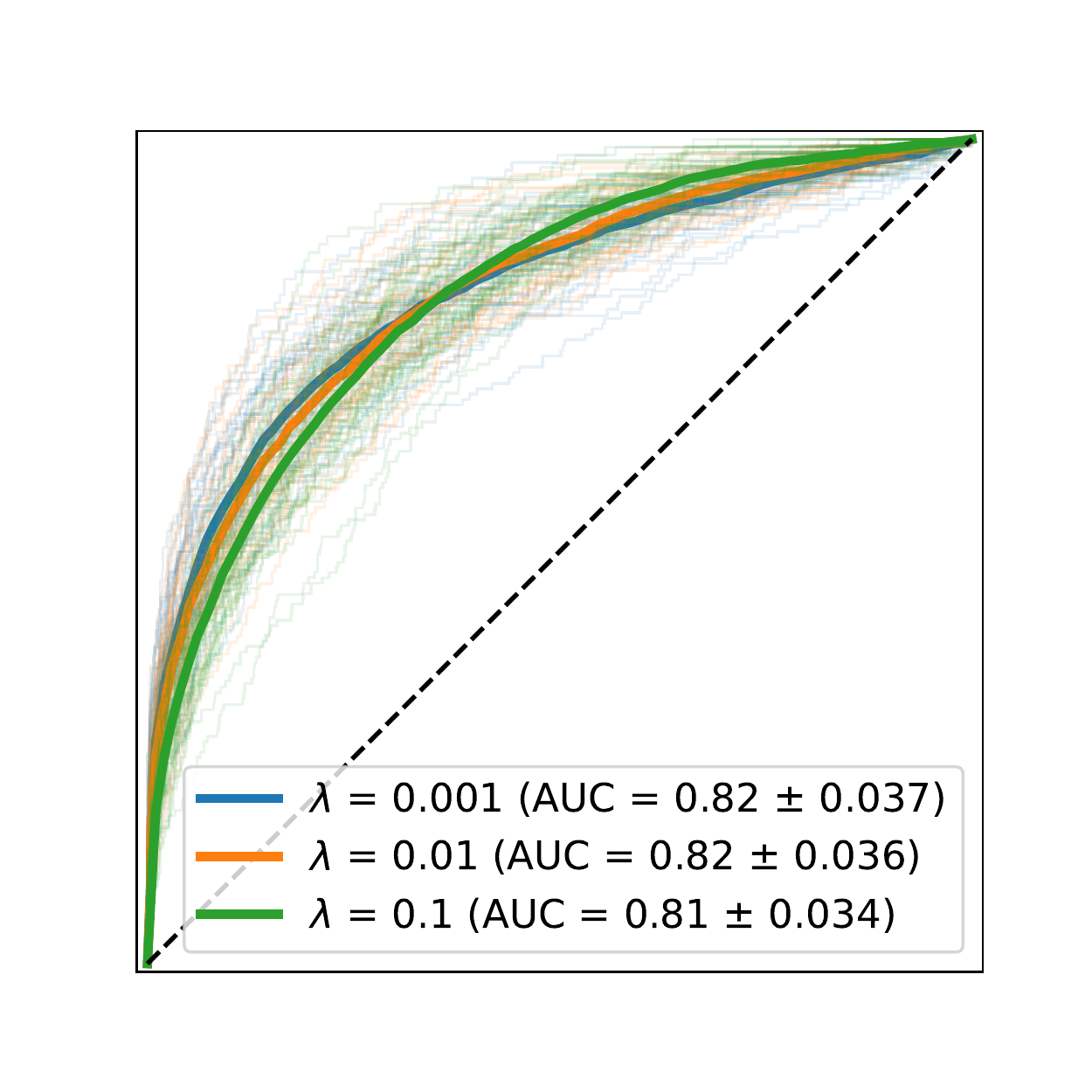}}\\[-2ex]
	
	\subfloat{\includegraphics[trim = 1.5mm 3mm 11mm 14mm, clip,width=0.27\textwidth]{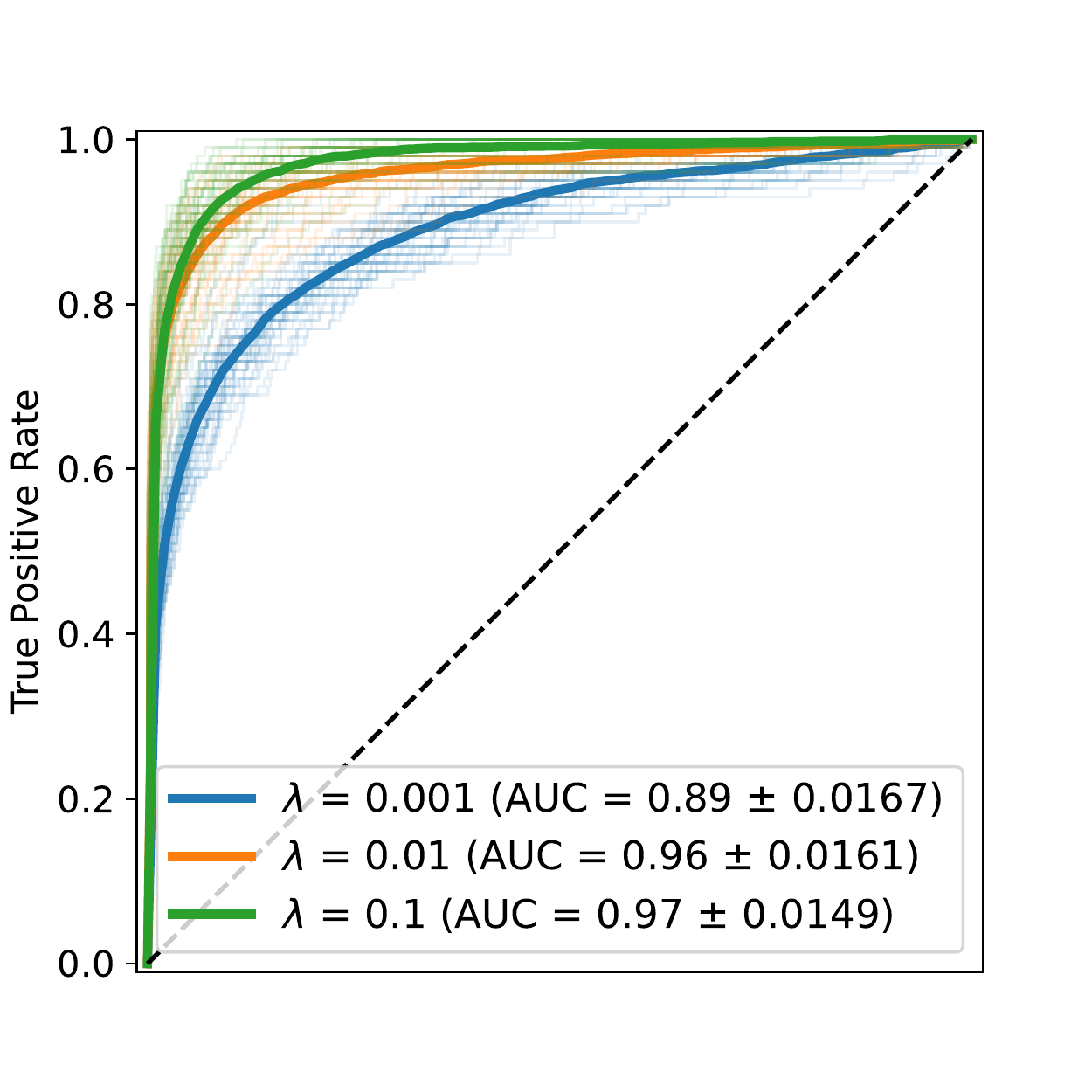}}\hfill
	\subfloat{\includegraphics[trim = 14mm 3mm 11mm 14mm, clip,width=0.24\textwidth]{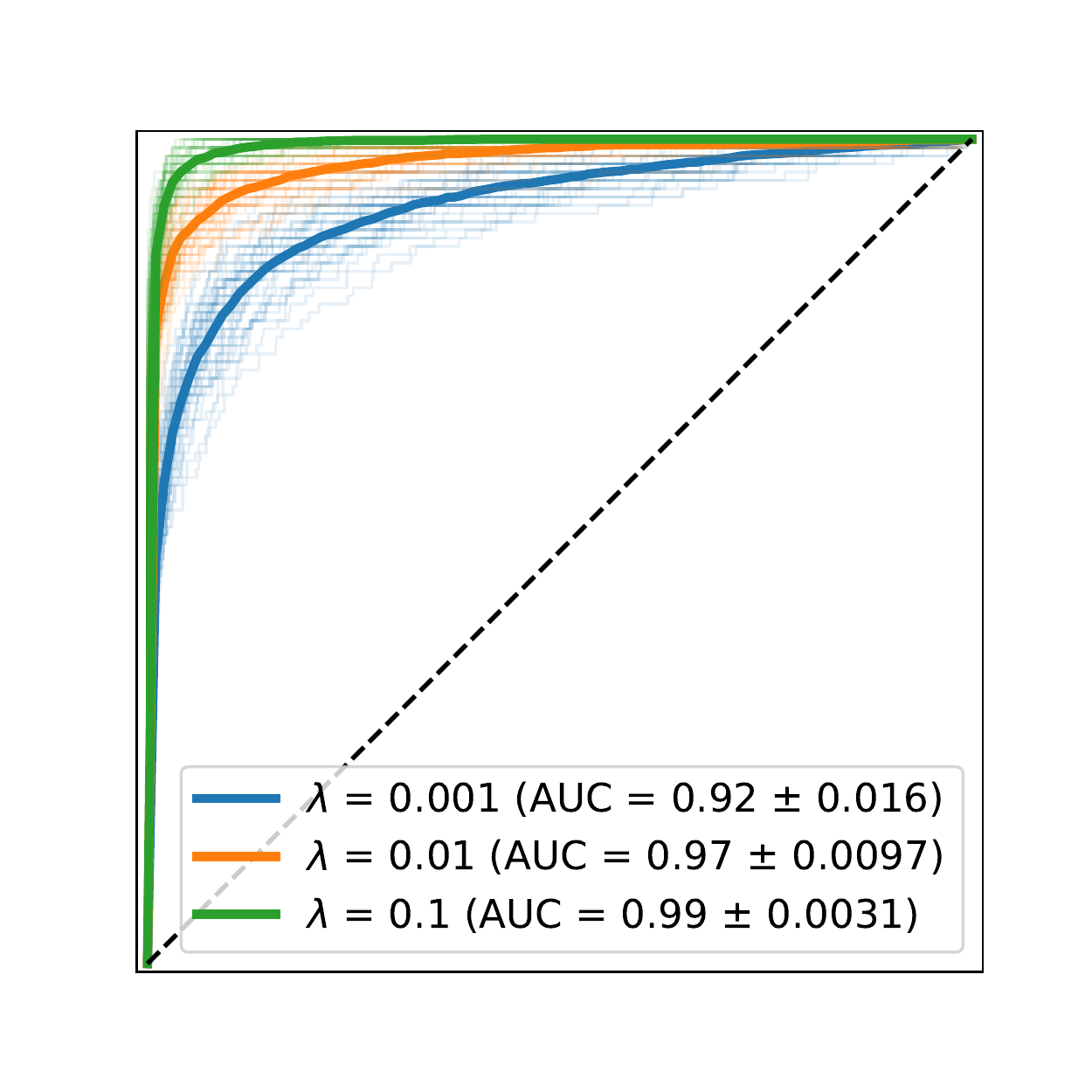}}\hfill  
	\subfloat{\includegraphics[trim = 14mm 3mm 11mm 14mm, clip,width=0.24\textwidth]{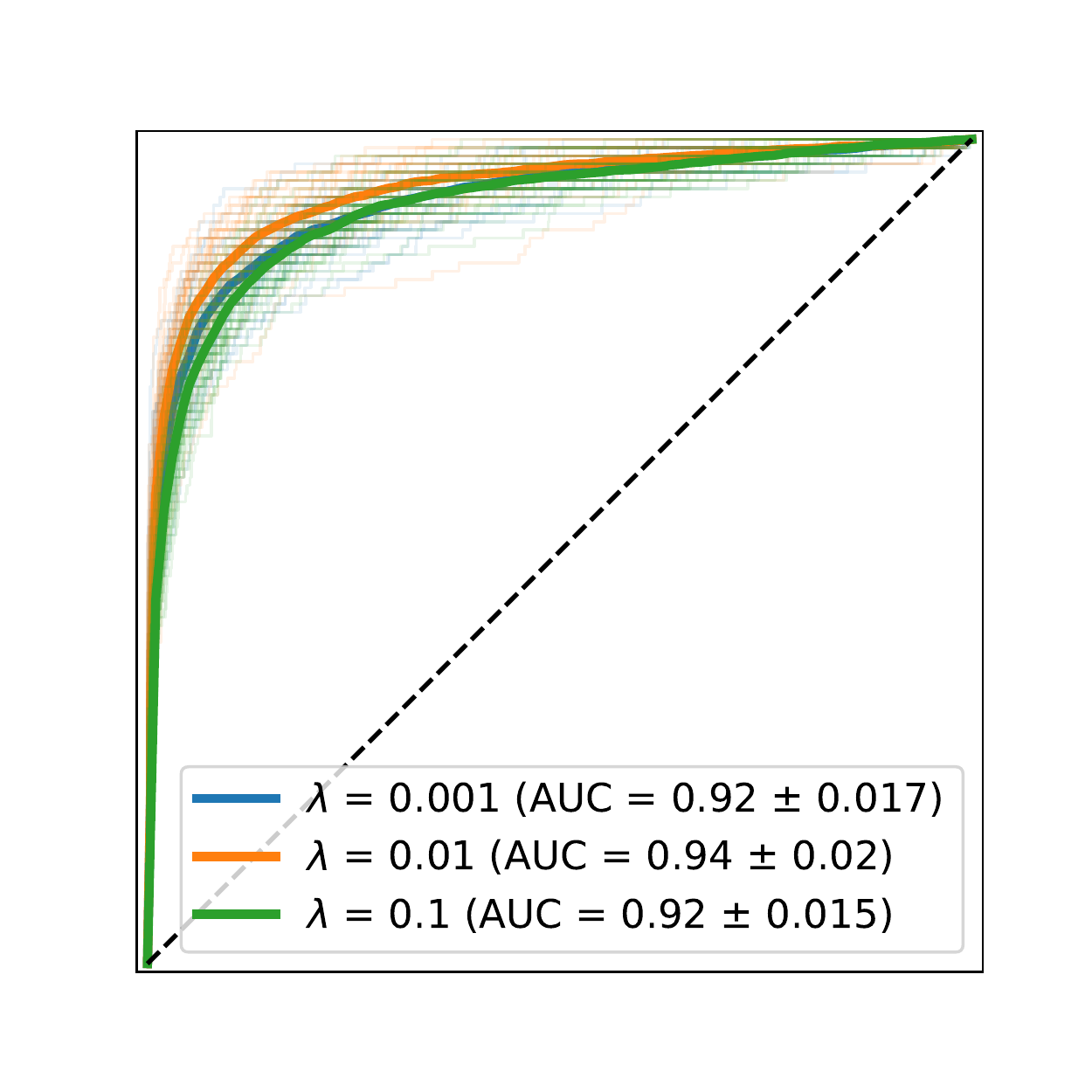}}\hfill
	\subfloat{\includegraphics[trim = 14mm 3mm 11mm 14mm, clip,width=0.24\textwidth]{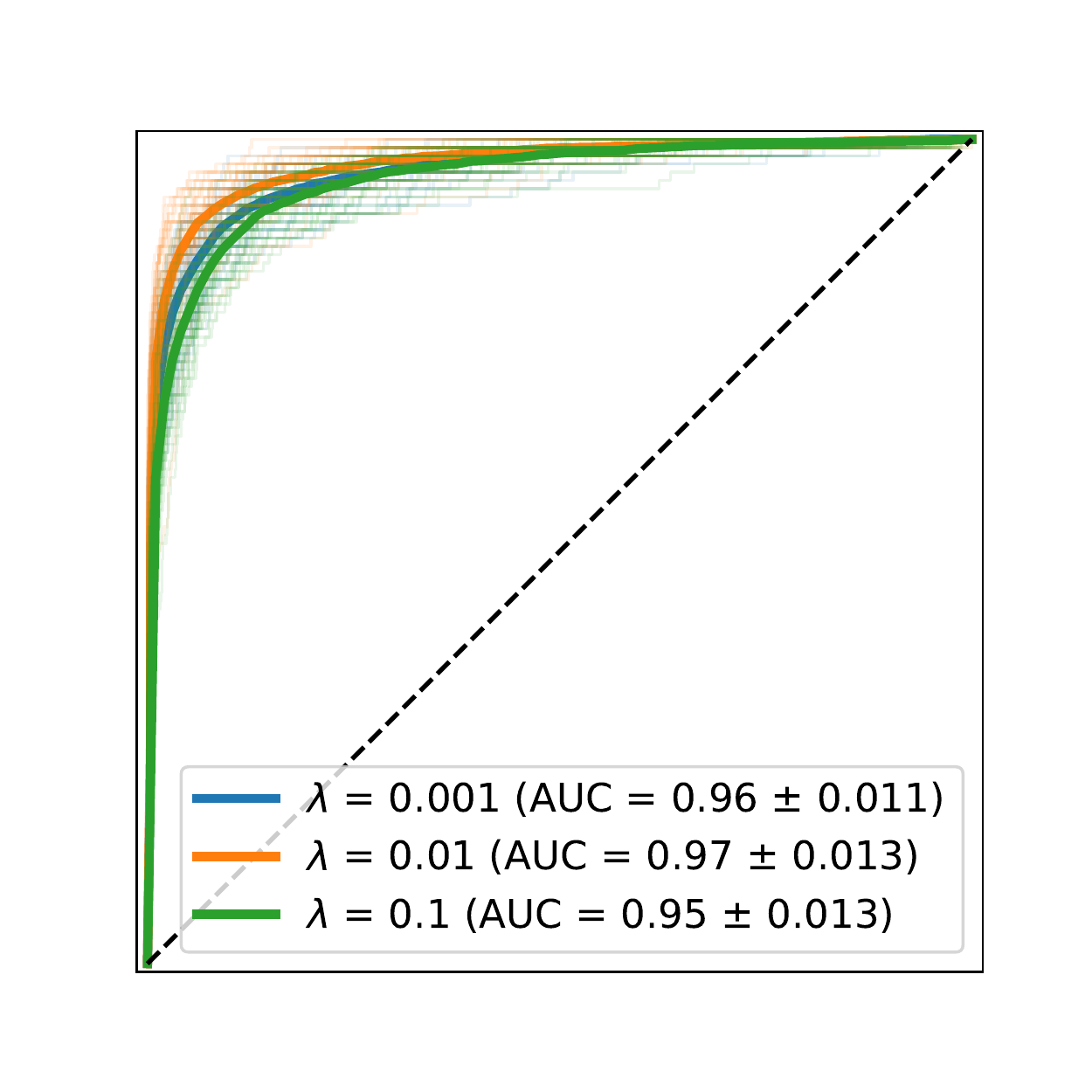}}\\[-2ex]
	\addtocounter{subfigure}{-12}
	\subfloat[\textsf{GGM}]{\includegraphics[trim = 1.5mm 2.5mm 11mm 14mm, clip,width=0.27\textwidth]{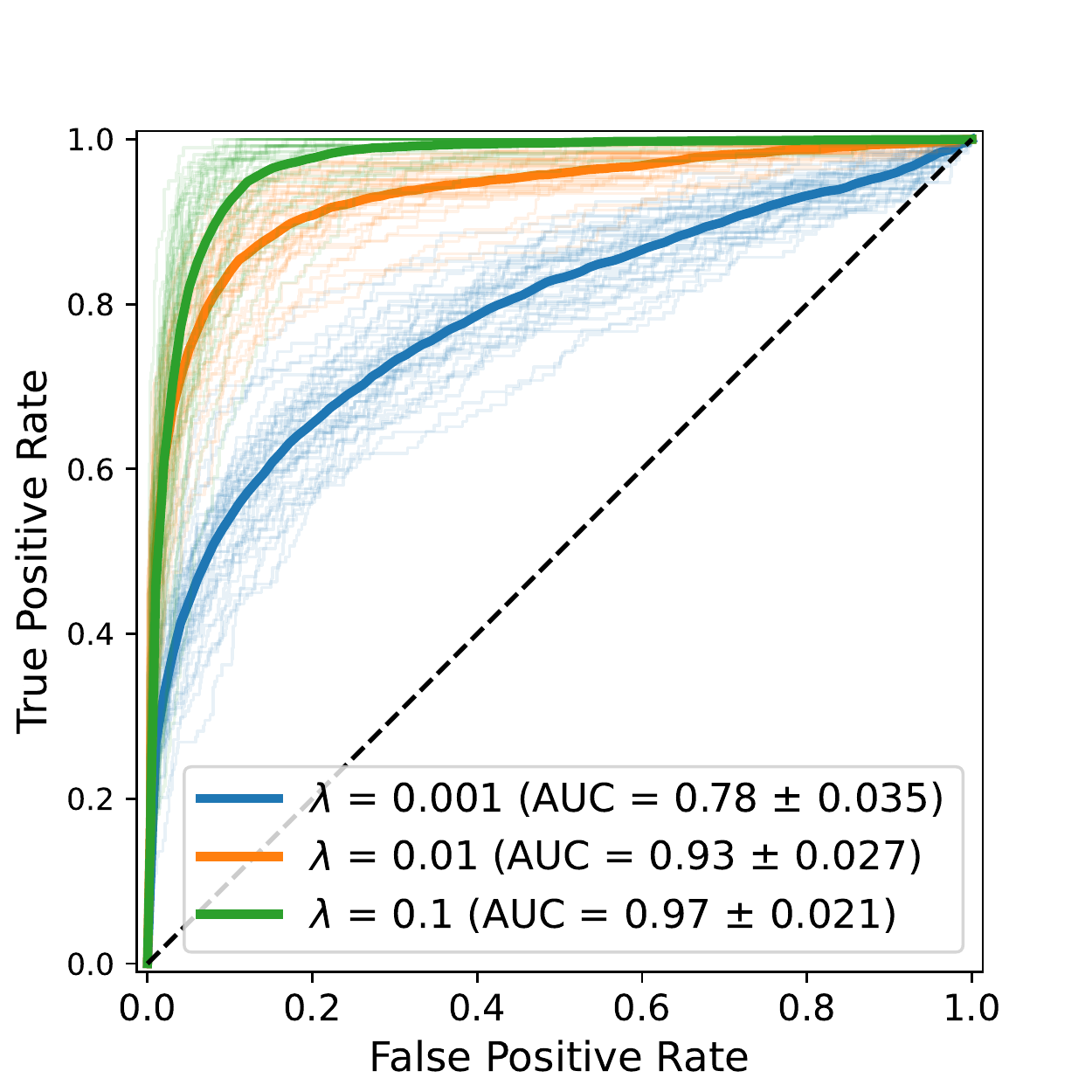}}\hfill
	\subfloat[\textsf{EGM}]{\includegraphics[trim = 14mm 2mm 11mm 14mm, clip, width=0.24\textwidth]{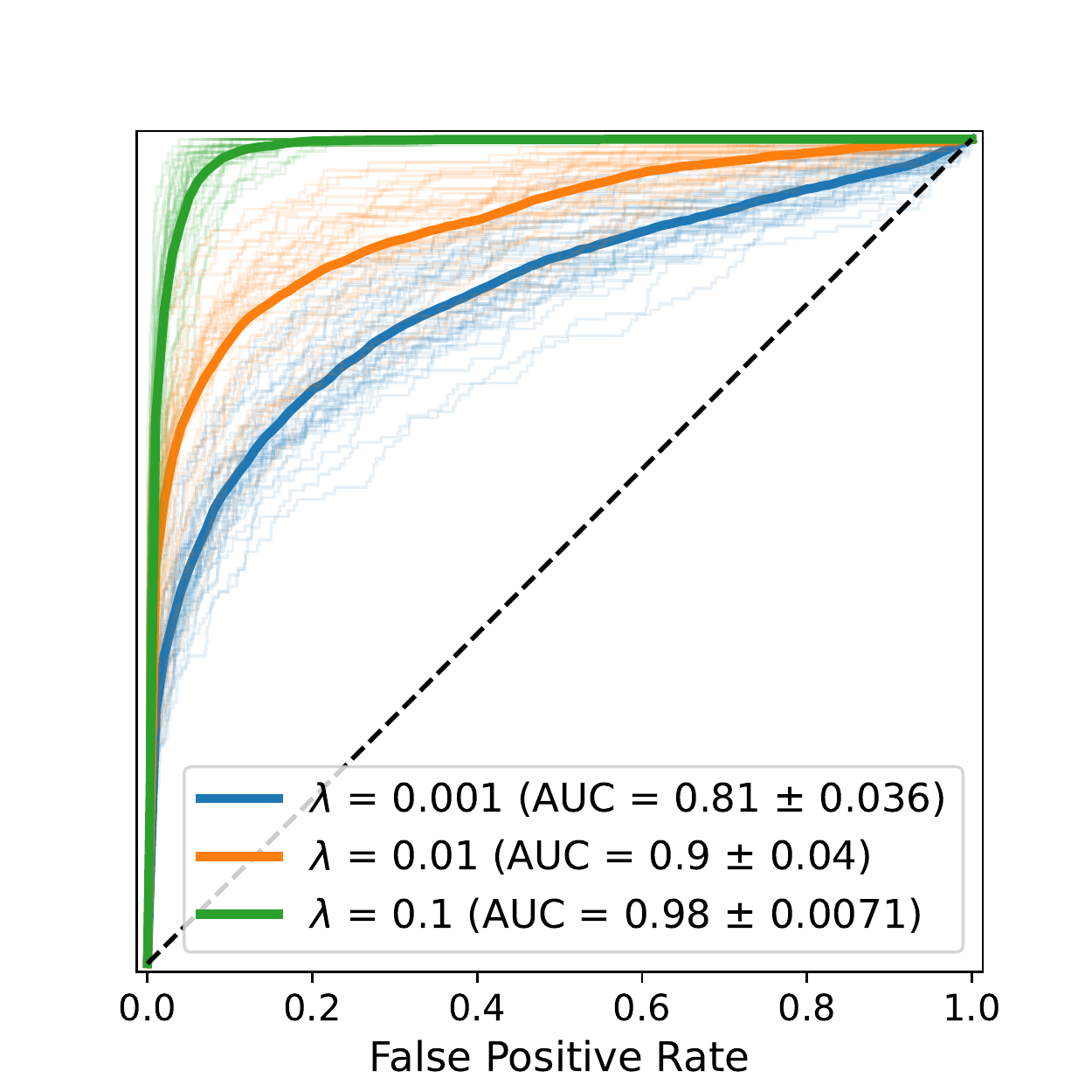}}\hfill  
	\subfloat[\textsf{GGFM} ($k=20$)]{\includegraphics[trim = 14mm 2mm 11mm 14mm, clip,width=0.24\textwidth]{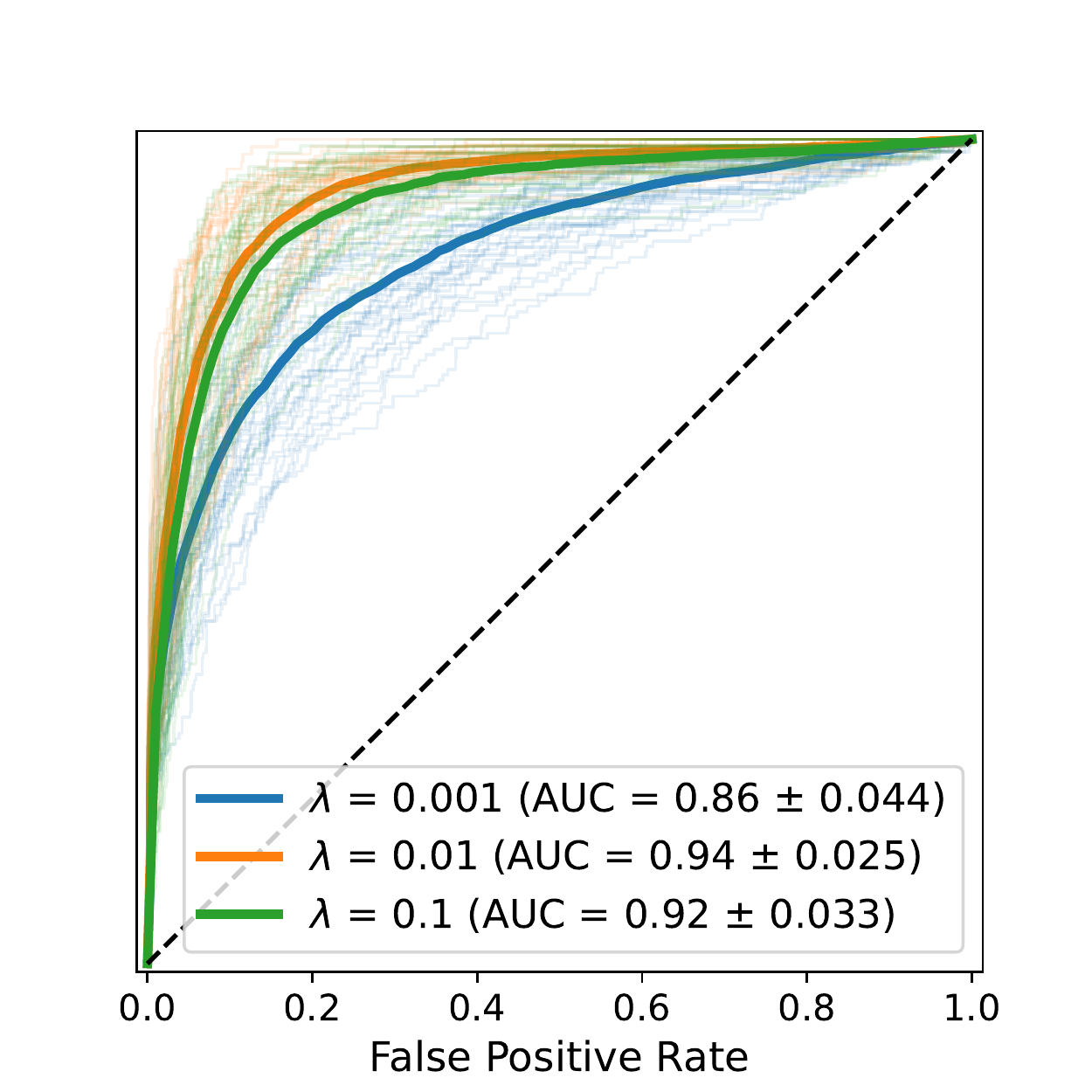}}\hfill
	\subfloat[\textsf{EGFM} ($k=20$)]{\includegraphics[trim = 14mm 2mm 11mm 14mm, clip,width=0.24\textwidth]{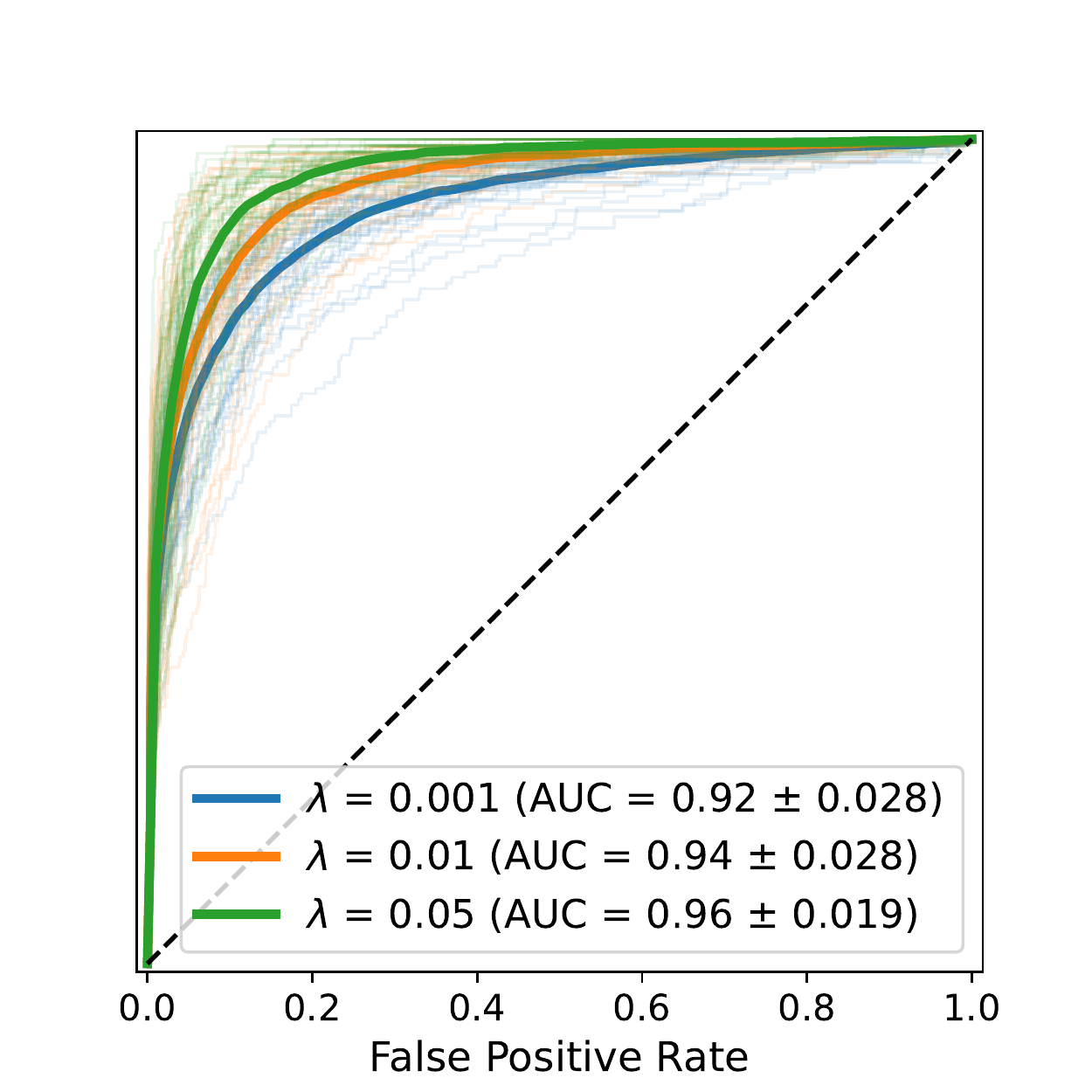}}
	\caption{\textbf{Sensitivity to $\lambda$}: mean ROC curves obtained from the estimated adjacency matrix $\hat{\mathbf{A}}$ of Barabási-Albert (first row), Erdős–Rényi (second row), Watts-Strogatz (third row) and Random geometric (fourth row) graphs with \textsf{GGM/EGM/GGFM/EGFM} algorithms with different values of the sparsity parameter $\lambda$. }
	\label{fig:sensitivity_lambda}
\end{figure}

\autoref{fig:sensitivity_lambda} shows the sensitivity of the proposed algorithms to the sparsity parameter $\lambda$ for each considered graph model, with $p=50$ and $\nu=3$. We observe that the performances are slightly different as $\lambda$ varies.
For certain graph models (especially random geometric) \textsf{GGM/EGM} appear to be sensitive to this parameter, so its order of magnitude should be chosen carefully.
We also observe that factor-model based approaches \textsf{GGFM/EGFM} have the interesting property of being less sensitive to a changing $\lambda$ for each graph model.

\begin{figure}
	\subfloat{\includegraphics[trim = 1.5mm 3mm 11mm 14mm, clip,width=0.27\textwidth]{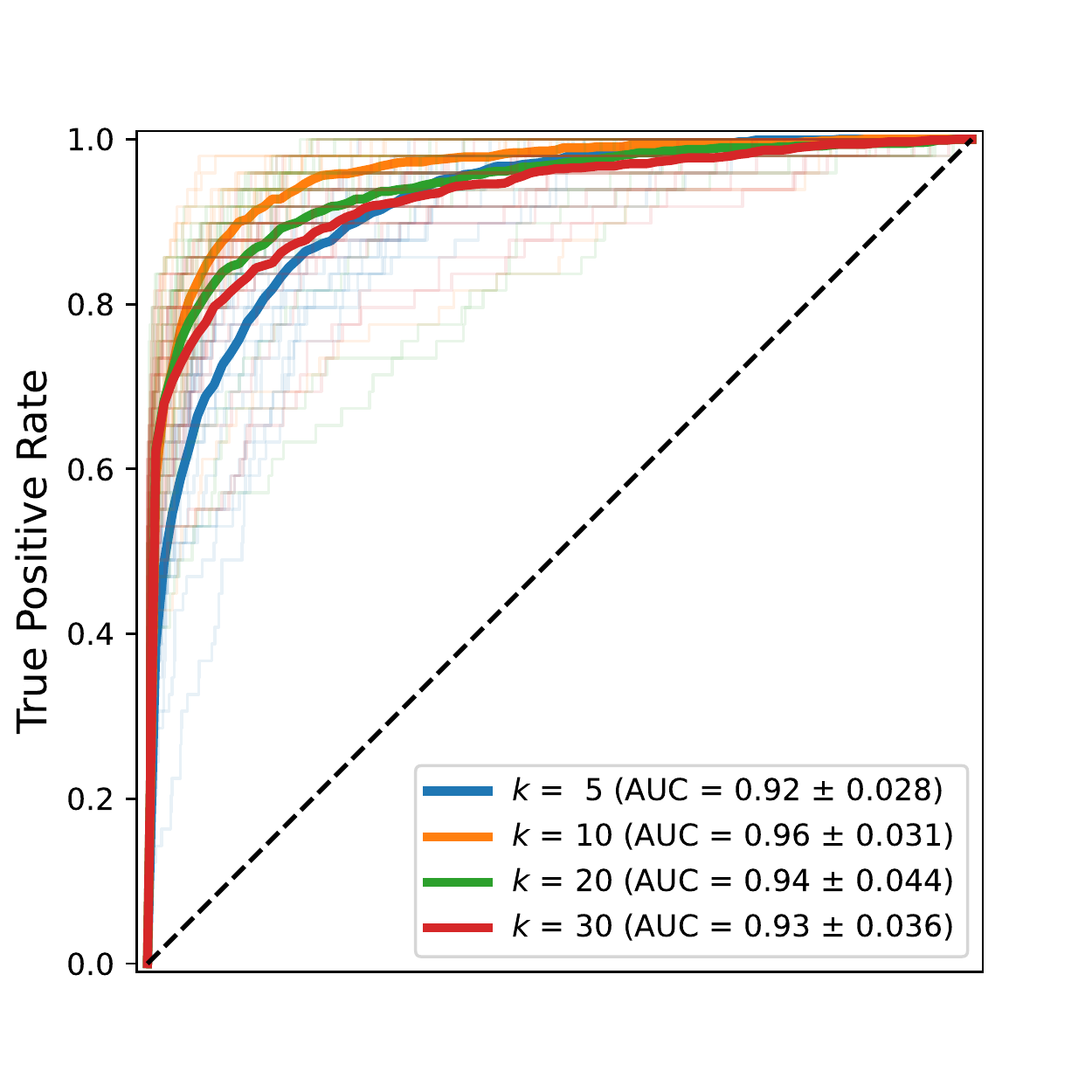}}\hfill
	\subfloat{\includegraphics[trim = 14mm 3mm 11mm 14mm, clip,width=0.24\textwidth]{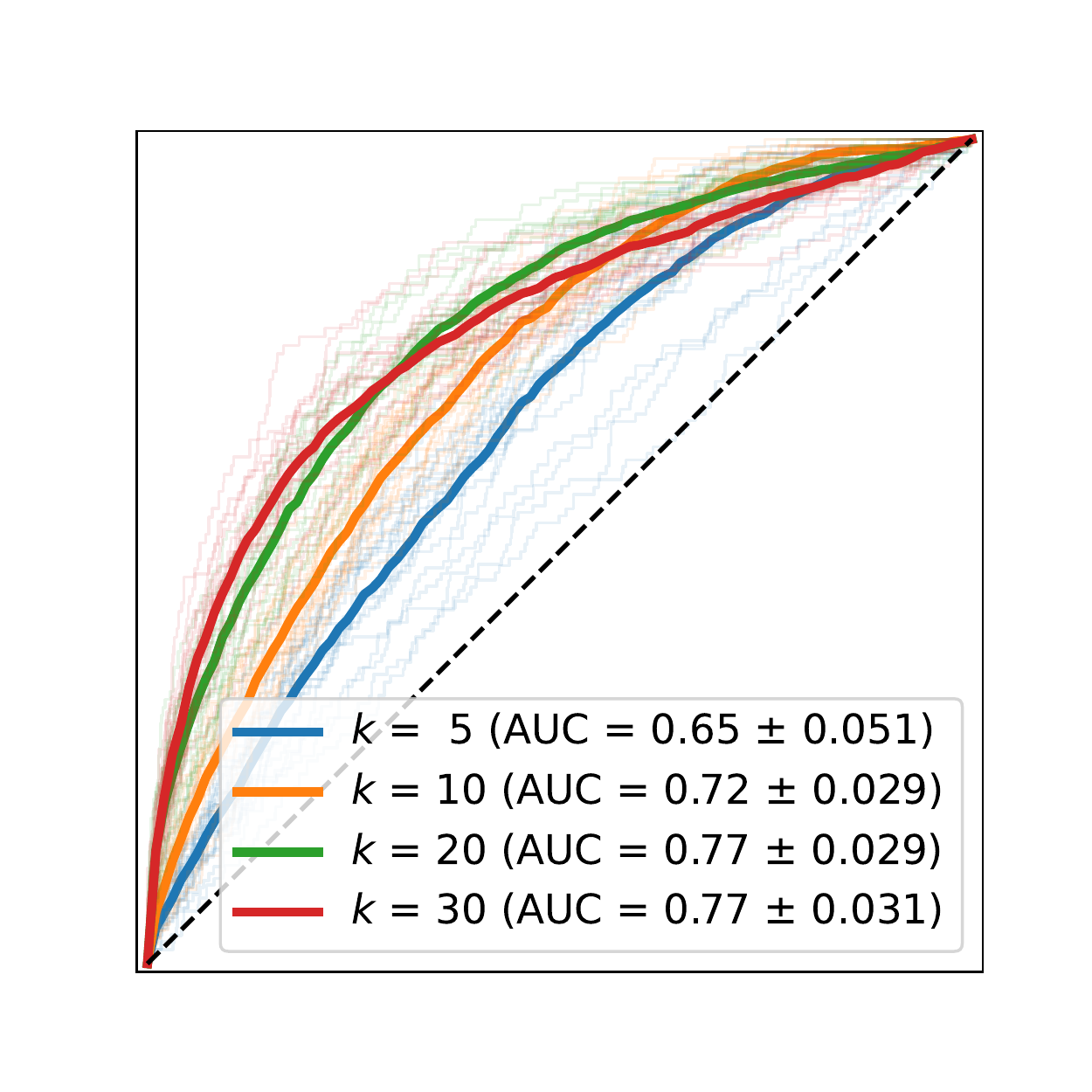}}\hfill  
	\subfloat{\includegraphics[trim = 14mm 3mm 11mm 14mm, clip,width=0.24\textwidth]{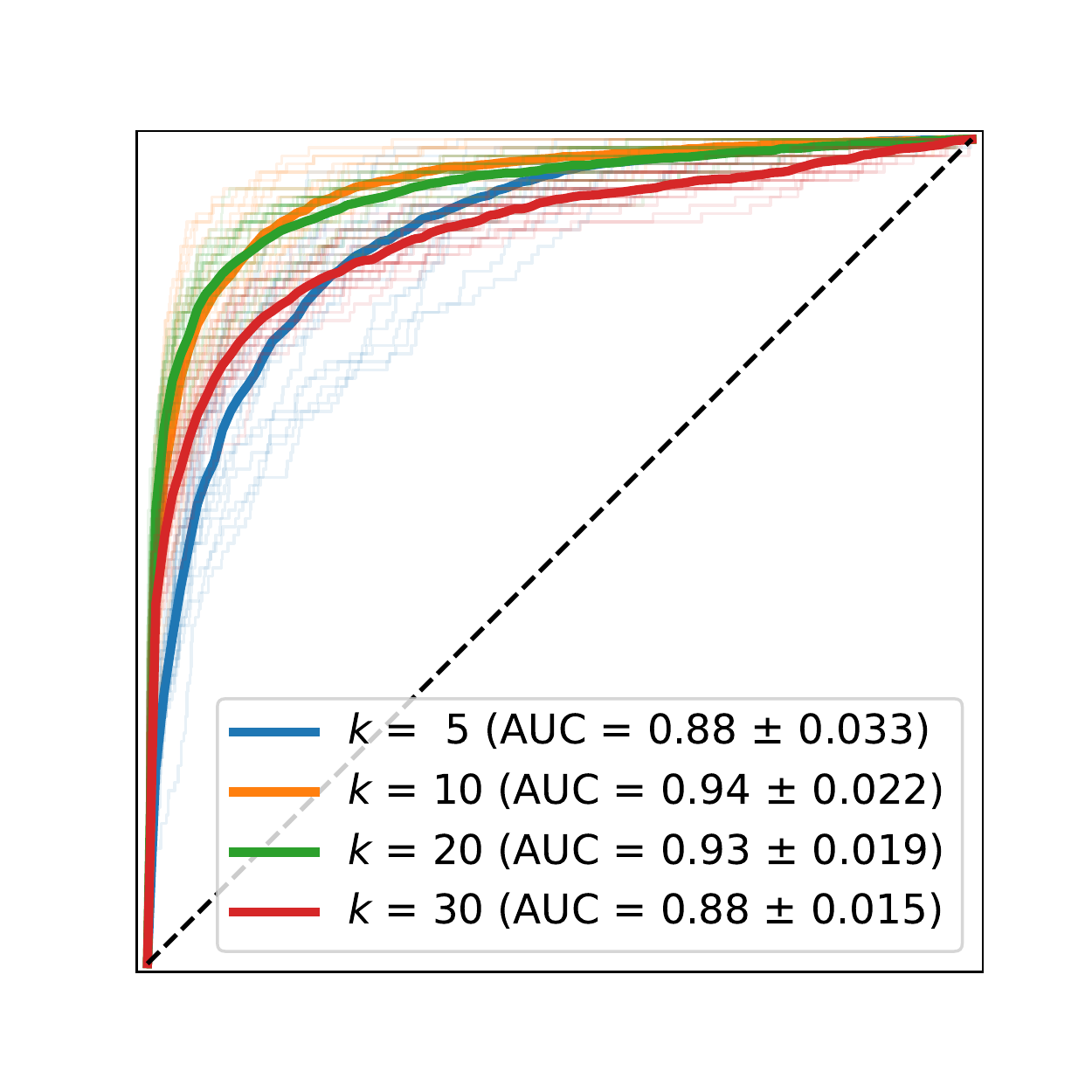}}\hfill
	\subfloat{\includegraphics[trim = 14mm 3mm 11mm 14mm, clip,width=0.24\textwidth]{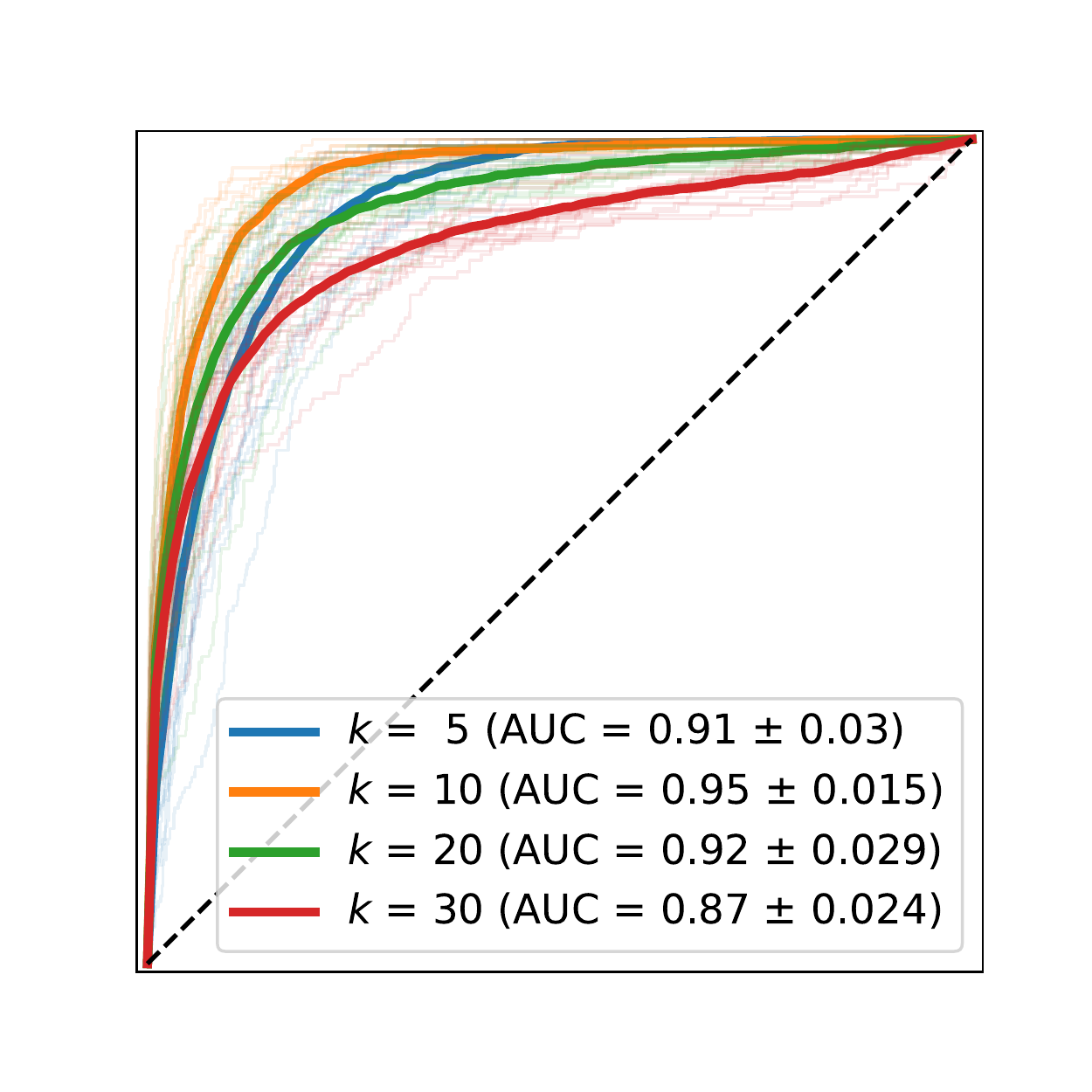}}\\[-2ex]
	
	\addtocounter{subfigure}{-4}
	\subfloat[Barab\'asi-Albert]{\includegraphics[trim = 1.5mm 3mm 11mm 14mm, clip,width=0.27\textwidth]{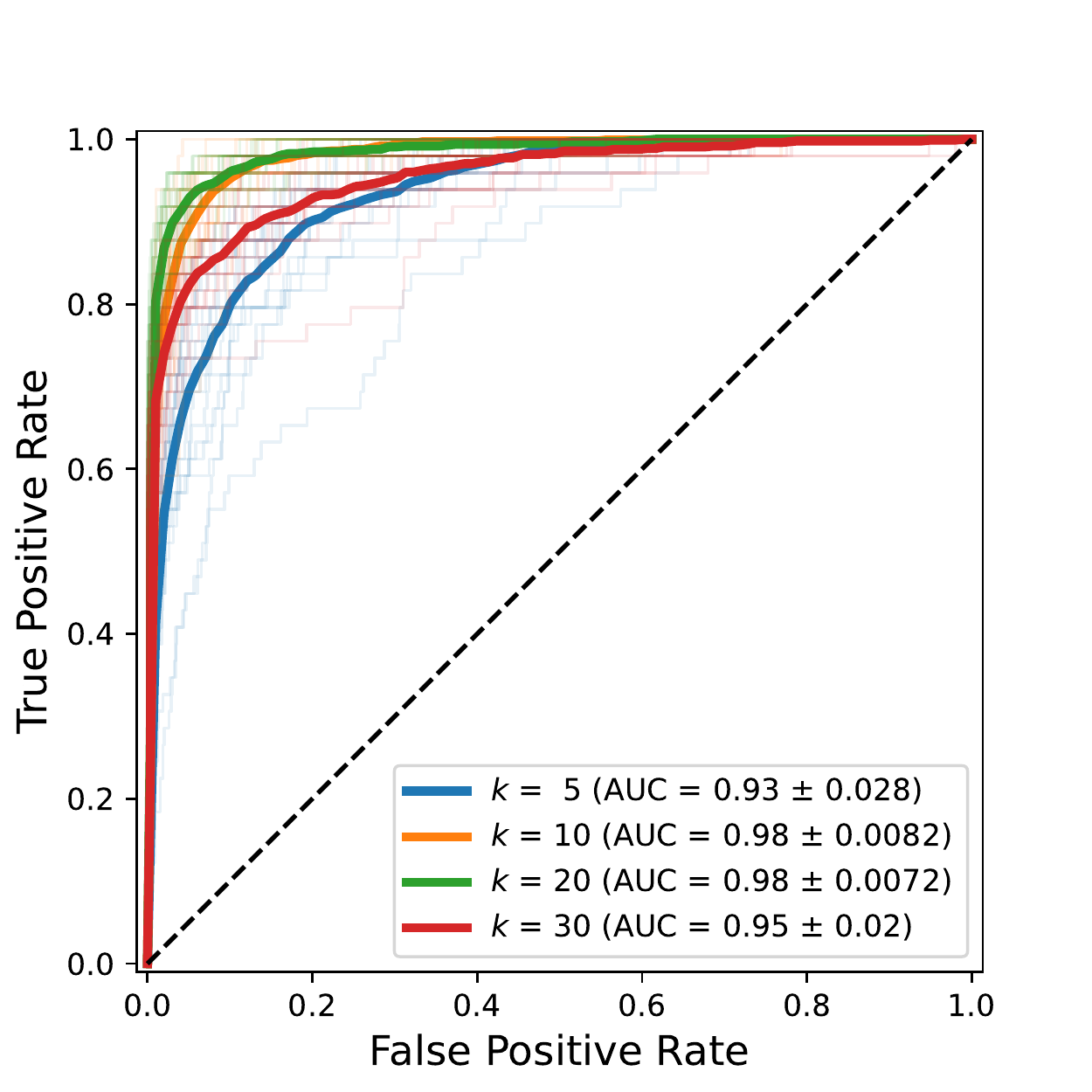}}\hfill
	\subfloat[Erdős–Rényi]{\includegraphics[trim = 14mm 2mm 11mm 14mm, clip, width=0.24\textwidth]{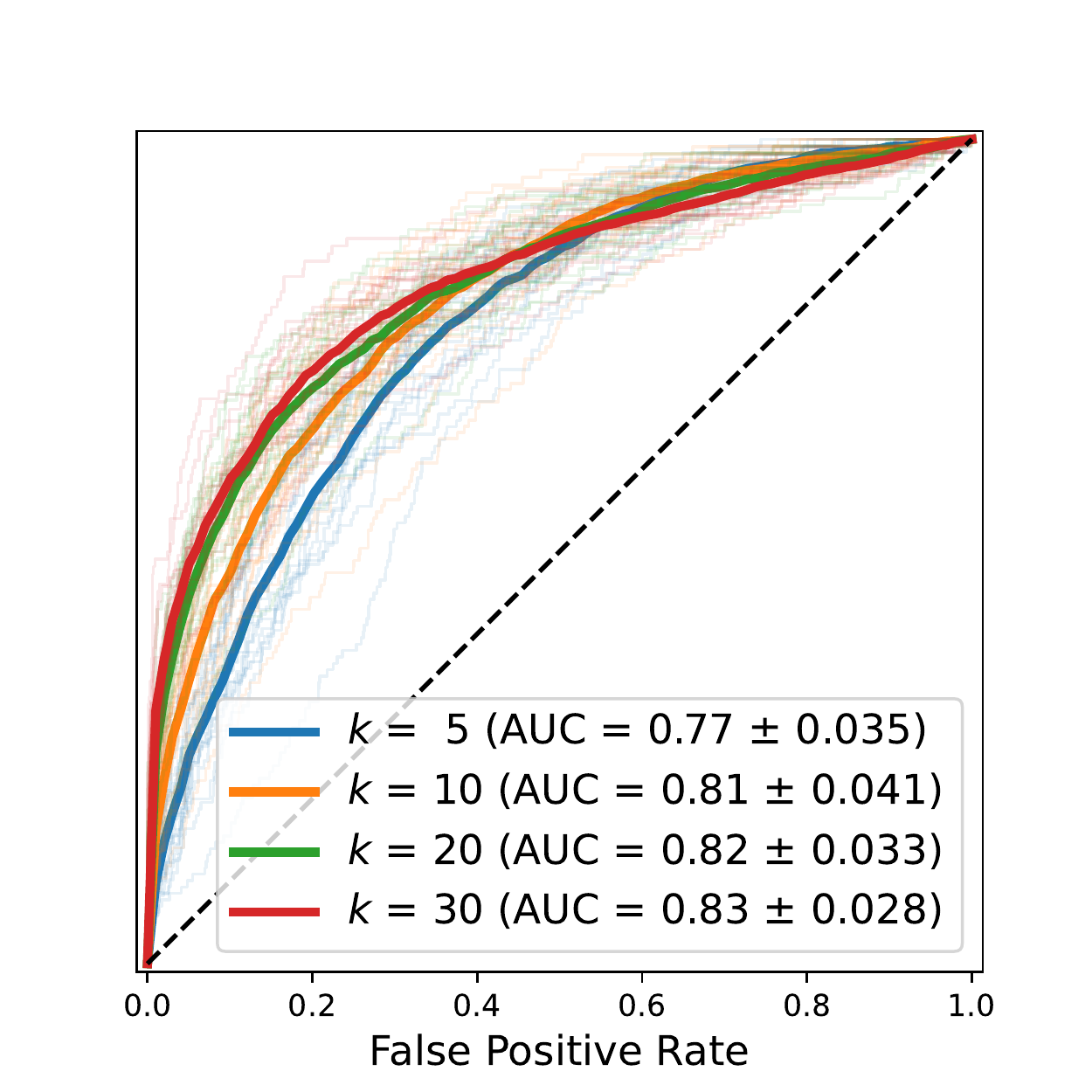}}\hfill  
	\subfloat[Watts-Strogatz]{\includegraphics[trim = 14mm 2mm 11mm 14mm, clip,width=0.24\textwidth]{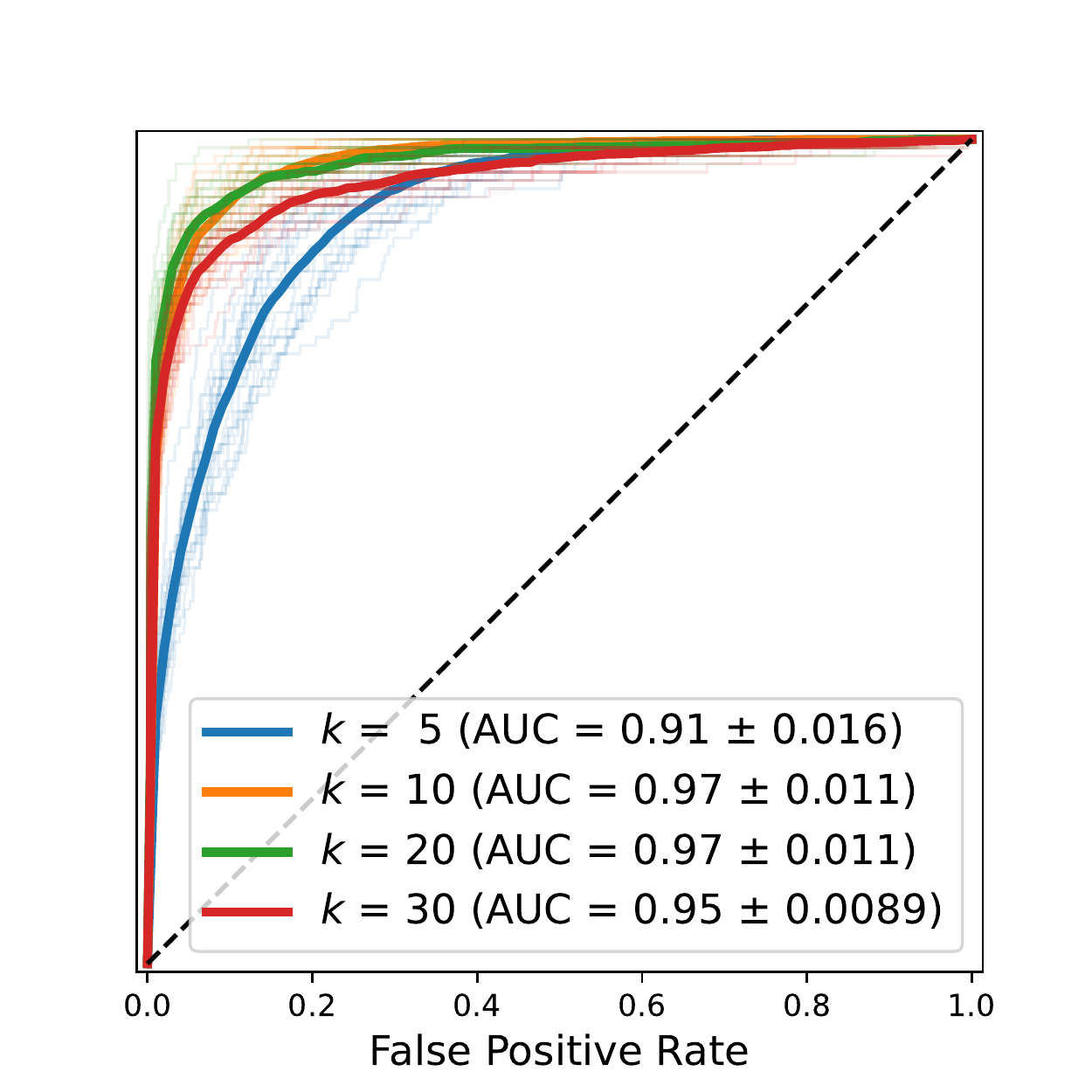}}\hfill
	\subfloat[Random geometric]{\includegraphics[trim = 14mm 2mm 11mm 14mm, clip,width=0.24\textwidth]{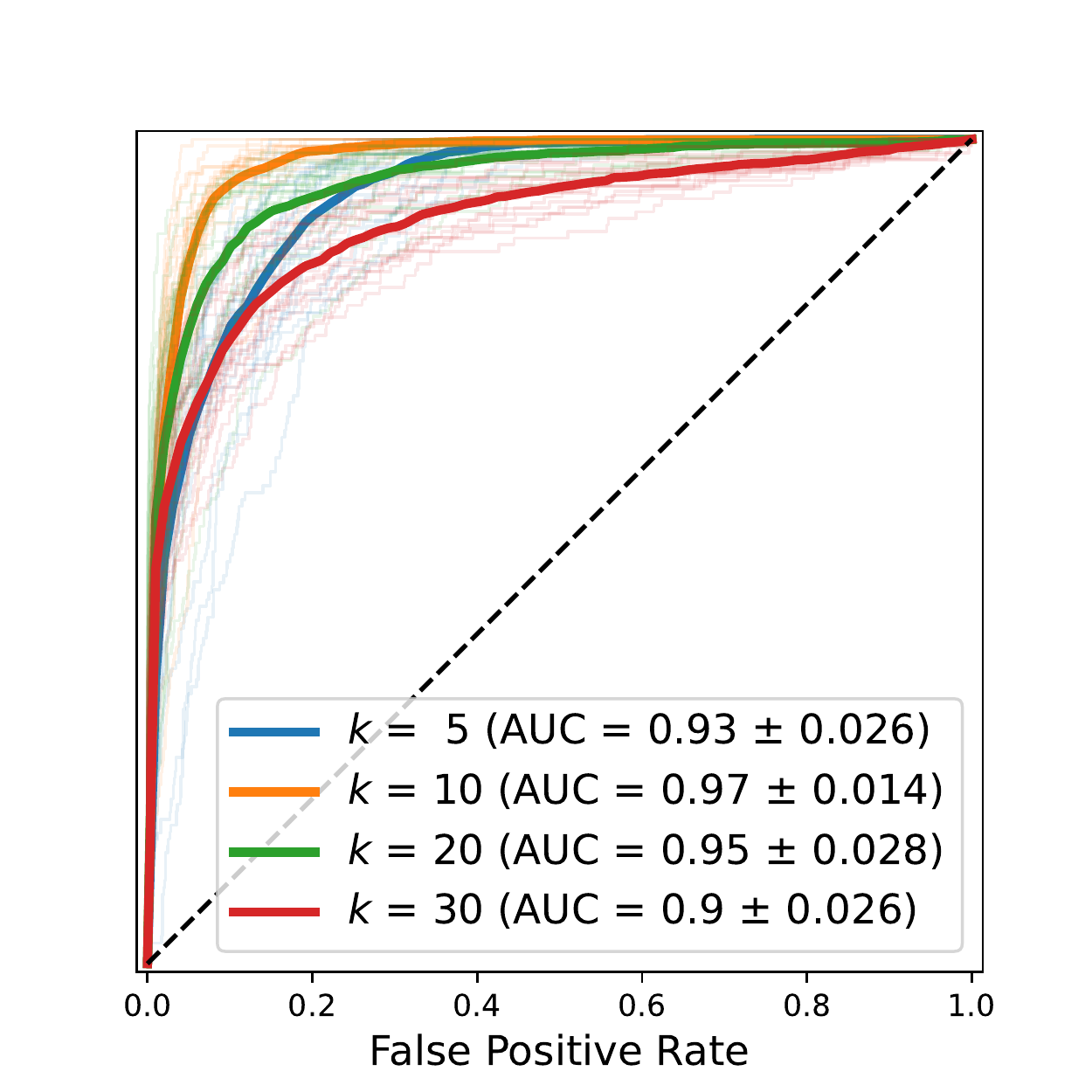}}
	\caption{\textbf{Sensitivity to $k$}: mean ROC curves obtained from the estimated adjacency matrix $\hat{\mathbf{A}}$ with \textsf{GGFM/EGFM} algorithms (first and second row, respectively) with different values of the rank $k$. For each case, parameters $\lambda$ are fixed according to Figure~\ref{fig:sensitivity_lambda}.}
	\label{fig:sensitivity_rank}
\end{figure}

\autoref{fig:sensitivity_rank} shows the sensitivity to the rank $k$, also with $p=50$ and $\nu=3$. We observe that acceptable performances are obtained when $k$ is decreased. As the data is not necessarily low-rank, similar results are obtained compared to \textsf{GGM/EGM}, which was already observed in \autoref{subsec:result}. 
However, unlike \textsf{GGM/EGM}, we show that factor-model based approaches are particularly useful for providing interpretable and computable graphs from real-world data, as illustrated in \autoref{subsec:real_data}.

%
%
%
%
\bibliography{biblio} 

\begin{thebibliography}{}

\bibitem[Absil et~al., 2008]{absil2008optimization}
Absil, P.-A., Mahony, R., and Sepulchre, R. (2008).
\newblock {\em Optimization Algorithms on Matrix Manifolds}.
\newblock Princeton University Press, Princeton, NJ, USA.

\bibitem[Anderson and Fang, 1990]{anderson1990theory}
Anderson, T.~W. and Fang, K.-T. (1990).
\newblock Theory and applications of elliptically contoured and related
  distributions.

\bibitem[Aytes, 2012]{aytes2012}
Aytes, A. (2012).
\newblock Return of the crowds: Mechanical turk and neoliberal states of
  exception.
\newblock In {\em Digital labor}, pages 87--105. Routledge.

\bibitem[Benfenati et~al., 2020]{benfenati2020proximal}
Benfenati, A., Chouzenoux, E., and Pesquet, J.-C. (2020).
\newblock Proximal approaches for matrix optimization problems: Application to
  robust precision matrix estimation.
\newblock {\em Signal Processing}, 169:107417.

\bibitem[Bhatia, 2009]{bhatia2009positive}
Bhatia, R. (2009).
\newblock {\em Positive definite matrices}.
\newblock Princeton University Press.

\bibitem[Bonnabel and Sepulchre, 2009]{bonabel2009riemannian}
Bonnabel, S. and Sepulchre, R. (2009).
\newblock {R}iemannian metric and geometric mean for positive semidefinite
  matrices of fixed rank.
\newblock {\em SIAM Journal on Matrix Analysis and Applications},
  31(3):1055--1070.

\bibitem[Bouchard et~al., 2021]{bouchard2021riemannian}
Bouchard, F., Breloy, A., Ginolhac, G., Renaux, A., and Pascal, F. (2021).
\newblock A riemannian framework for low-rank structured elliptical models.
\newblock {\em IEEE Transactions on Signal Processing}, 69:1185--1199.

\bibitem[Boumal, 2020]{boumal2020introduction}
Boumal, N. (2020).
\newblock An introduction to optimization on smooth manifolds.
\newblock {\em Available online, May}, 3.

\bibitem[Cai et~al., 2022]{cai2022mtp2}
Cai, J.-F., Cardoso, J. V. d.~M., Palomar, D.~P., and Ying, J. (2022).
\newblock Fast projected newton-like method for precision matrix estimation
  under total positivity.

\bibitem[Chandra et~al., 2021]{chandra2021bayesian}
Chandra, N.~K., Mueller, P., and Sarkar, A. (2021).
\newblock Bayesian scalable precision factor analysis for massive sparse
  gaussian graphical models.
\newblock {\em arXiv preprint arXiv:2107.11316}.

\bibitem[Chung, 1997]{chung1997spectral}
Chung, F.~R. (1997).
\newblock {\em Spectral graph theory}, volume~92.
\newblock American Mathematical Soc.

\bibitem[Cordasco and Gargano, 2010]{cordasco2010community}
Cordasco, G. and Gargano, L. (2010).
\newblock Community detection via semi-synchronous label propagation
  algorithms.
\newblock {\em 2010 IEEE International Workshop on: Business Applications of
  Social Network Analysis (BASNA)}, pages 1--8.

\bibitem[Crawford, 2021]{crawford2021}
Crawford, K. (2021).
\newblock {\em The atlas of AI: Power, politics, and the planetary costs of
  artificial intelligence}.
\newblock Yale University Press.

\bibitem[de~Miranda~Cardoso et~al., 2021]{de2021graphical}
de~Miranda~Cardoso, J.~V., Ying, J., and Palomar, D. (2021).
\newblock Graphical models in heavy-tailed markets.
\newblock {\em Advances in Neural Information Processing Systems},
  34:19989--20001.

\bibitem[Dempster, 1972]{dempster1972covariance}
Dempster, A.~P. (1972).
\newblock Covariance selection.
\newblock {\em Biometrics}, pages 157--175.

\bibitem[Dra{\v{s}}kovi{\'c} and Pascal, 2018]{dravskovic2018new}
Dra{\v{s}}kovi{\'c}, G. and Pascal, F. (2018).
\newblock New insights into the statistical properties of $ m $-estimators.
\newblock {\em IEEE Transactions on Signal Processing}, 66(16):4253--4263.

\bibitem[Edelman et~al., 1998]{edelman1998geometry}
Edelman, A., Arias, T.~A., and Smith, S.~T. (1998).
\newblock The geometry of algorithms with orthogonality constraints.
\newblock {\em SIAM journal on Matrix Analysis and Applications},
  20(2):303--353.

\bibitem[Egilmez et~al., 2017]{egilmez2017graph}
Egilmez, H.~E., Pavez, E., and Ortega, A. (2017).
\newblock Graph learning from data under laplacian and structural constraints.
\newblock {\em IEEE Journal of Selected Topics in Signal Processing},
  11(6):825--841.

\bibitem[Fallat et~al., 2017]{fallat2017total}
Fallat, S., Lauritzen, S., Sadeghi, K., Uhler, C., Wermuth, N., and Zwiernik,
  P. (2017).
\newblock Total positivity in markov structures.
\newblock {\em The Annals of Statistics}, pages 1152--1184.

\bibitem[Fattahi and Sojoudi, 2019]{fattahi2019graphical}
Fattahi, S. and Sojoudi, S. (2019).
\newblock Graphical lasso and thresholding: Equivalence and closed-form
  solutions.
\newblock {\em Journal of machine learning research}.

\bibitem[Finegold and Drton, 2014]{finegold2014robust}
Finegold, M.~A. and Drton, M. (2014).
\newblock Robust graphical modeling with t-distributions.
\newblock {\em arXiv preprint arXiv:1408.2033}.

\bibitem[Friedman et~al., 2008]{friedman2008sparse}
Friedman, J., Hastie, T., and Tibshirani, R. (2008).
\newblock Sparse inverse covariance estimation with the graphical lasso.
\newblock {\em Biostatistics}, 9(3):432--441.

\bibitem[Gray and Suri, 2019]{gray2019}
Gray, M.~L. and Suri, S. (2019).
\newblock {\em Ghost work: How to stop Silicon Valley from building a new
  global underclass}.
\newblock Eamon Dolan Books.

\bibitem[Hein{\"a}vaara et~al., 2016]{heinavaara2016inconsistency}
Hein{\"a}vaara, O., Lepp{\"a}-Aho, J., Corander, J., and Honkela, A. (2016).
\newblock On the inconsistency of $\ell_1$-penalised sparse precision matrix
  estimation.
\newblock {\em BMC bioinformatics}, 17(16):99--107.

\bibitem[Hestenes and Stiefel, 1952]{hestenes1952methods}
Hestenes, M.~R. and Stiefel, E. (1952).
\newblock Methods of conjugate gradients for solving linear equation.
\newblock {\em Journal of research of the National Bureau of Standards},
  49(6):409.

\bibitem[Jeuris et~al., 2012]{jeuris2012survey}
Jeuris, B., Vandebril, R., and Vandereycken, B. (2012).
\newblock A survey and comparison of contemporary algorithms for computing the
  matrix geometric mean.
\newblock {\em Electronic Transactions on Numerical Analysis}, 39:379--402.

\bibitem[Kai-Tai and Yao-Ting, 1990]{kai1990generalized}
Kai-Tai, F. and Yao-Ting, Z. (1990).
\newblock {\em Generalized multivariate analysis}, volume~19.
\newblock Science Press Beijing and Springer-Verlag, Berlin.

\bibitem[Kalofolias, 2016]{kalofolias2016learn}
Kalofolias, V. (2016).
\newblock How to learn a graph from smooth signals.
\newblock In {\em Artificial Intelligence and Statistics}, pages 920--929.
  PMLR.

\bibitem[Khamaru and Mazumder, 2019]{khamaru2019computation}
Khamaru, K. and Mazumder, R. (2019).
\newblock Computation of the maximum likelihood estimator in low-rank factor
  analysis.
\newblock {\em Mathematical Programming}, 176(1):279--310.

\bibitem[Kovnatsky et~al., 2016]{kovnatsky2016madmm}
Kovnatsky, A., Glashoff, K., and Bronstein, M.~M. (2016).
\newblock Madmm: a generic algorithm for non-smooth optimization on manifolds.
\newblock In {\em European Conference on Computer Vision}, pages 680--696.
  Springer.

\bibitem[Kumar et~al., 2020]{kumar2020unified}
Kumar, S., Ying, J., de~Miranda~Cardoso, J.~V., and Palomar, D.~P. (2020).
\newblock A unified framework for structured graph learning via spectral
  constraints.
\newblock {\em J. Mach. Learn. Res.}, 21(22):1--60.

\bibitem[Lake and Tenenbaum, 2010]{lake2010discovering}
Lake, B. and Tenenbaum, J. (2010).
\newblock Discovering structure by learning sparse graphs.

\bibitem[Lam and Fan, 2009]{lam2009sparsistency}
Lam, C. and Fan, J. (2009).
\newblock Sparsistency and rates of convergence in large covariance matrix
  estimation.
\newblock {\em The Annals of statistics}, 37(6B):4254--4278.

\bibitem[Lauritzen et~al., 2019]{lauritzen2019maximum}
Lauritzen, S., Uhler, C., and Zwiernik, P. (2019).
\newblock Maximum likelihood estimation in gaussian models under total
  positivity.
\newblock {\em The Annals of Statistics}, 47(4):1835--1863.

\bibitem[Lauritzen, 1996]{lauritzen1996graphical}
Lauritzen, S.~L. (1996).
\newblock {\em Graphical models}, volume~17.
\newblock Clarendon Press.

\bibitem[Ledoit and Wolf, 2004]{ledoit2004well}
Ledoit, O. and Wolf, M. (2004).
\newblock A well-conditioned estimator for large-dimensional covariance
  matrices.
\newblock {\em Journal of multivariate analysis}, 88(2):365--411.

\bibitem[Li and Gui, 2006]{li2006gradient}
Li, H. and Gui, J. (2006).
\newblock Gradient directed regularization for sparse gaussian concentration
  graphs, with applications to inference of genetic networks.
\newblock {\em Biostatistics}, 7(2):302--317.

\bibitem[Maronna, 1976]{maronna1976robust}
Maronna, R.~A. (1976).
\newblock Robust m-estimators of multivariate location and scatter.
\newblock {\em The annals of statistics}, pages 51--67.

\bibitem[Marti et~al., 2021]{marti2021review}
Marti, G., Nielsen, F., Bi{\'n}kowski, M., and Donnat, P. (2021).
\newblock A review of two decades of correlations, hierarchies, networks and
  clustering in financial markets.
\newblock {\em Progress in Information Geometry}, pages 245--274.

\bibitem[Massart and Absil, 2018]{massart2018quotient}
Massart, E. and Absil, P.-A. (2018).
\newblock Quotient geometry with simple geodesics for the manifold of
  fixed-rank positive-semidefinite matrices.
\newblock {\em Technical Report UCL-INMA-2018.06}.

\bibitem[Mazumder and Hastie, 2012]{mazumder2012graphical}
Mazumder, R. and Hastie, T. (2012).
\newblock The graphical lasso: New insights and alternatives.
\newblock {\em Electronic journal of statistics}, 6:2125.

\bibitem[Meng et~al., 2014]{meng2014learning}
Meng, Z., Eriksson, B., and Hero, A. (2014).
\newblock Learning latent variable gaussian graphical models.
\newblock In {\em International Conference on Machine Learning}, pages
  1269--1277. PMLR.

\bibitem[Meyer et~al., 2011]{meyer2011regression}
Meyer, G., Bonnabel, S., and Sepulchre, R. (2011).
\newblock Regression on fixed-rank positive semidefinite matrices: a
  {R}iemannian approach.
\newblock {\em Journal of Machine Learning Research}, 12:593--625.

\bibitem[Neuman et~al., 2021]{neuman2021restricted}
Neuman, A.~M., Xie, Y., and Sun, Q. (2021).
\newblock Restricted {R}iemannian geometry for positive semidefinite matrices.
\newblock {\em arXiv preprint arXiv:2105.14691}.

\bibitem[Newman, 2006]{newman2006modularity}
Newman, M. E.~J. (2006).
\newblock Modularity and community structure in networks.
\newblock {\em Proceedings of the National Academy of Sciences},
  103(23):8577--8582.

\bibitem[Osherson et~al., 1991]{osherson1991animal}
Osherson, D.~N., Stern, J., Wilkie, O., Stob, M., and Smith, E.~E. (1991).
\newblock Default probability.
\newblock {\em Cognitive Science}, 15(2):251--269.

\bibitem[Peltier et~al., 2017]{peltier2017assessing}
Peltier, A., Froger, J.-L., Villeneuve, N., and Catry, T. (2017).
\newblock Assessing the reliability and consistency of {I}n{SAR} and {GNSS}
  data for retrieving 3{D}-displacement rapid changes, the example of the 2015
  {P}iton de la {F}ournaise eruptions.
\newblock {\em Journal of Volcanology and Geothermal Research}, 344:106--120.

\bibitem[Robertson and Symons, 2007]{robertson2007maximum}
Robertson, D. and Symons, J. (2007).
\newblock Maximum likelihood factor analysis with rank-deficient sample
  covariance matrices.
\newblock {\em Journal of Multivariate Analysis}, 98(4):813--828.

\bibitem[Rubin and Thayer, 1982]{rubin1982algorithms}
Rubin, D.~B. and Thayer, D.~T. (1982).
\newblock Em algorithms for ml factor analysis.
\newblock {\em Psychometrika}, 47(1):69--76.

\bibitem[Shen et~al., 2012]{shen2012likelihood}
Shen, X., Pan, W., and Zhu, Y. (2012).
\newblock Likelihood-based selection and sharp parameter estimation.
\newblock {\em Journal of the American Statistical Association},
  107(497):223--232.

\bibitem[Shuman et~al., 2013]{shuman2013emerging}
Shuman, D.~I., Narang, S.~K., Frossard, P., Ortega, A., and Vandergheynst, P.
  (2013).
\newblock The emerging field of signal processing on graphs: Extending
  high-dimensional data analysis to networks and other irregular domains.
\newblock {\em IEEE signal processing magazine}, 30(3):83--98.

\bibitem[Skovgaard, 1984]{skovgaard1984riemannian}
Skovgaard, L.~T. (1984).
\newblock A riemannian geometry of the multivariate normal model.
\newblock {\em Scandinavian journal of statistics}, pages 211--223.

\bibitem[Smith et~al., 2011]{smith2011network}
Smith, S.~M., Miller, K.~L., Salimi-Khorshidi, G., Webster, M., Beckmann,
  C.~F., Nichols, T.~E., Ramsey, J.~D., and Woolrich, M.~W. (2011).
\newblock Network modelling methods for fmri.
\newblock {\em Neuroimage}, 54(2):875--891.

\bibitem[Smith, 2005]{smith2005covariance}
Smith, S.~T. (2005).
\newblock Covariance, subspace, and intrinsic crame/spl acute/r-rao bounds.
\newblock {\em IEEE Transactions on Signal Processing}, 53(5):1610--1630.

\bibitem[Smittarello et~al., 2019]{smittarello2019magma}
Smittarello, D., Cayol, V., Pinel, V., Peltier, A., Froger, J.-L., and
  Ferrazzini, V. (2019).
\newblock Magma propagation at {P}iton de la {F}ournaise from joint inversion
  of {I}n{SAR} and {GNSS}.
\newblock {\em Journal of Geophysical Research: Solid Earth},
  124(2):1361--1387.

\bibitem[Stegle et~al., 2015]{stegle2015computational}
Stegle, O., Teichmann, S.~A., and Marioni, J.~C. (2015).
\newblock Computational and analytical challenges in single-cell
  transcriptomics.
\newblock {\em Nature Reviews Genetics}, 16(3):133--145.

\bibitem[Tarzanagh and Michailidis, 2018]{tarzanagh2018estimation}
Tarzanagh, D.~A. and Michailidis, G. (2018).
\newblock Estimation of graphical models through structured norm minimization.
\newblock {\em Journal of machine learning research}, 18(1).

\bibitem[Tipping and Bishop, 1999]{tipping1999probabilistic}
Tipping, M.~E. and Bishop, C.~M. (1999).
\newblock Probabilistic principal component analysis.
\newblock {\em Journal of the Royal Statistical Society: Series B (Statistical
  Methodology)}, 61(3):611--622.

\bibitem[Tubaro and Casilli, 2019]{tubaro2019}
Tubaro, P. and Casilli, A.~A. (2019).
\newblock Micro-work, artificial intelligence and the automotive industry.
\newblock {\em Journal of Industrial and Business Economics}, 46:333--345.

\bibitem[Tyler, 1987]{tyler1987distribution}
Tyler, D.~E. (1987).
\newblock A distribution-free m-estimator of multivariate scatter.
\newblock {\em The annals of Statistics}, pages 234--251.

\bibitem[Vandereycken et~al., 2012]{vandereycken2012riemannian}
Vandereycken, B., Absil, P.-A., and Vandewalle, S. (2012).
\newblock A {R}iemannian geometry with complete geodesics for the set of
  positive semidefinite matrices of fixed rank.
\newblock {\em IMA Journal of Numerical Analysis}, 33(2):481--514.

\bibitem[Vershynin, 2012]{vershynin2012close}
Vershynin, R. (2012).
\newblock How close is the sample covariance matrix to the actual covariance
  matrix?
\newblock {\em Journal of Theoretical Probability}, 25(3):655--686.

\bibitem[Vogel and Fried, 2011]{vogel2011elliptical}
Vogel, D. and Fried, R. (2011).
\newblock Elliptical graphical modelling.
\newblock {\em Biometrika}, 98(4):935--951.

\bibitem[Wald et~al., 2019]{wald2019globally}
Wald, Y., Noy, N., Elidan, G., and Wiesel, A. (2019).
\newblock Globally optimal learning for structured elliptical losses.
\newblock {\em Advances in Neural Information Processing Systems}, 32.

\bibitem[Ying et~al., 2020]{ying2020nonconvex}
Ying, J., de~Miranda~Cardoso, J.~V., and Palomar, D. (2020).
\newblock Nonconvex sparse graph learning under laplacian constrained graphical
  model.
\newblock {\em Advances in Neural Information Processing Systems},
  33:7101--7113.

\bibitem[Yoshida and West, 2010]{yoshida2010bayesian}
Yoshida, R. and West, M. (2010).
\newblock Bayesian learning in sparse graphical factor models via variational
  mean-field annealing.
\newblock {\em J Mach Learn Res}, 11:1771--1798.

\bibitem[Zhang et~al., 2013]{zhang2013multivariate}
Zhang, T., Wiesel, A., and Greco, M.~S. (2013).
\newblock Multivariate generalized gaussian distribution: Convexity and
  graphical models.
\newblock {\em IEEE Transactions on Signal Processing}, 61(16):4141--4148.

\bibitem[Zhao and Jiang, 2006]{zhao2006probabilistic}
Zhao, J. and Jiang, Q. (2006).
\newblock Probabilistic pca for t distributions.
\newblock {\em Neurocomputing}, 69(16-18):2217--2226.

\bibitem[Zhao et~al., 2019]{zhao2019optimization}
Zhao, L., Wang, Y., Kumar, S., and Palomar, D.~P. (2019).
\newblock Optimization algorithms for graph laplacian estimation via {ADMM} and
  {MM}.
\newblock {\em IEEE Transactions on Signal Processing}, 67(16):4231--4244.

\bibitem[Zhou et~al., 2019]{zhou2019robust}
Zhou, R., Liu, J., Kumar, S., and Palomar, D.~P. (2019).
\newblock Robust factor analysis parameter estimation.
\newblock In {\em International Conference on Computer Aided Systems Theory},
  pages 3--11. Springer.

\end{thebibliography}

\end{document}